\documentclass{article}

\PassOptionsToPackage{numbers,sort&compress}{natbib}

 \usepackage[preprint]{neurips_2026}

\usepackage[utf8]{inputenc} % allow utf-8 input
\usepackage[T1]{fontenc}    % use 8-bit T1 fonts
\usepackage{hyperref}       % hyperlinks
\usepackage{url}            % simple URL typesetting
\usepackage{booktabs}       % professional-quality tables
\usepackage{amsfonts}       % blackboard math symbols
\usepackage{nicefrac}       % compact symbols for 1/2, etc.
\usepackage{microtype}      % microtypography
\usepackage{xcolor}         % colors

\usepackage{algorithm}
\usepackage{algorithmic}
\usepackage[most]{tcolorbox}
\usepackage{subcaption}
\usepackage{tabularray}
\usepackage{wrapfig}
\usepackage{multirow}
\usepackage{graphicx}
\usepackage{makecell}
\usepackage{fontawesome}
\usepackage{enumitem}

\usepackage{amsmath}
\usepackage{amssymb}
\usepackage{mathtools}
\usepackage{amsthm}
\usepackage{amsfonts}
\usepackage{bbm}

\title{Same Question, Different Answers:\\
Evaluating LLM Reliability Beyond Accuracy}

\author{%
Kazem Faghih\thanks{Equal contribution. Correspondence to: kazemf@umd.edu} \quad
Yize Cheng\footnotemark[1] \quad
Shoumik Saha \quad
Mobina Pournemat \\
\bf
Armin Gerami \quad
Soheil Feizi \\
University of Maryland, College Park \\
\texttt{\{kazemf, yzcheng, smksaha, mpournem, agerami, sfeizi\}@umd.edu}\\
\\
\faGithub~\textbf{Project}: \url{https://github.com/kazemf78/llm-consistency-study.git}
}

\begin{document}

\maketitle

\begin{abstract}
Large language models (LLMs) often achieve strong accuracy on benchmarks, yet it remains unclear how reliably they apply this knowledge when the same question is phrased in different but equivalent ways. In this work, we study how model answers change under meaning-preserving paraphrases across factual question answering and mathematical reasoning tasks. Across four benchmarks and 13 models, we find that model outputs frequently depend on the exact wording of the prompt. While overall accuracy typically changes only modestly across paraphrases, instance-level behavior is far less stable: for many questions, models alternate between correct and incorrect answers depending on phrasing, with mismatch rates reaching more than 23\%. Conditioning on questions that are answered correctly in their original form reveals even larger failures measured by answer flip rates, showing that single-prompt correctness is often a poor indicator of reliability. At the same time, we find that models often produce a correct answer for \emph{at least one} paraphrase of a question, suggesting that the underlying knowledge is present but inconsistently retrieved. Building on this observation, we show that a simple self-paraphrasing strategy can partially recover this latent knowledge and improve performance at inference time. Together, these findings suggest that standard accuracy metrics can mask substantial instability, and that evaluating consistency across equivalent inputs provides a clearer picture of LLM reliability.
\end{abstract}

% \begin{abstract}
% Large language models (LLMs) often achieve strong accuracy on standard benchmarks, yet it remains unclear how reliably they apply this knowledge when the same question is phrased in equivalent ways. In this work, we study how model answers change under meaning-preserving paraphrases across factual question answering and mathematical reasoning tasks. Across four benchmarks and 13 models, we find that correctness frequently depends on the exact wording of the prompt. While overall accuracy typically changes only modestly across paraphrases, instance-level behavior is far less stable: for many questions, models alternate between correct and incorrect answers depending on phrasing, with mismatch rates reaching up to 23\%. Conditioning on questions answered correctly in their original form reveals larger failures, showing that single-prompt correctness is often a poor indicator of reliability. At the same time, models often produce a correct answer for \emph{at least one} paraphrase of a question, suggesting that the underlying knowledge is present but inconsistently retrieved. Building on this observation, we show that a simple self-paraphrasing strategy can recover some latent knowledge and improve performance at inference time. Together, these findings suggest that standard accuracy metrics can mask instability, and that evaluating consistency across equivalent inputs provides a clearer picture of LLM reliability.
% \end{abstract}

\section{Introduction}

Large Language Models (LLMs) have demonstrated remarkable capabilities in natural language understanding~\cite{devlin2019bertpretrainingdeepbidirectional, brown2020languagemodelsfewshotlearners, chowdhery2022palmscalinglanguagemodeling}, question answering~\cite{raffel2023exploringlimitstransferlearning, khashabi2020unifiedqacrossingformatboundaries}, and reasoning~\cite{wei2023chainofthoughtpromptingelicitsreasoning, Guo_2025}. 
As these models are deployed in an expanding range of real-world applications, an important and often under-evaluated dimension of reliability is output consistency under semantically equivalent inputs. Although variability can be acceptable or even desirable in open-ended settings such as summarization and creative generation~\cite{jiang2025artificialhivemindopenendedhomogeneity}, many high-stakes or decision-critical workflows require invariant behavior: \emph{meaning-preserving rephrasings should yield meaning-preserving outputs.} Such variation naturally arises in real-world usage, where different users express the same intent through diverse phrasings. This highlights a fundamental requirement for reliable deployment: consistency under meaning-preserving linguistic variation.
%However, as LLMs are increasingly used in diverse real-world scenarios, the \textbf{consistency} of their outputs across semantically equivalent inputs becomes a significant aspect of reliability. While some tasks, such as summarization or creative open-ended generation, tolerate or even benefit from output variation~\cite{jiang2025artificialhivemindopenendedhomogeneity}, many settings require stable, invariant behavior where semantically identical inputs lead to semantically identical outputs. 

Factual question-answering and mathematical problem-solving are clear examples where inconsistency directly undermines reliability, and similar concerns arise in classification, structured prediction, etc. When a model provides conflicting answers to semantically equivalent versions of the same input, it calls into question whether it genuinely captures the underlying meaning and core semantics, or instead relies on superficial cues or surface-level patterns. Therefore, characterizing the consistency of model behavior under meaning-preserving transformations is essential for measuring the true reliability of LLMs for real-world deployment.

While several studies have examined LLM robustness to input variations~\cite{alzahrani2024benchmarks, mizrahi2024state, sclar2023quantifying}, most focus on relatively shallow perturbations, such as prompt formatting changes, minor lexical edits, or adversarial noise, and are typically evaluated in multiple-choice settings, which do not reflect the complexity of open-ended generation. More recent efforts~\cite{lunardi2025robustness, choi2025roparq} have begun to study paraphrase-level variations at scale, showing that rewording alone can substantially affect model predictions. However, these studies are largely restricted to multiple-choice tasks, where output variability is inherently constrained. Moreover, recent works~\cite{pezeshkpour2024large, li-etal-2024-multiple} have argued that multiple-choice evaluation can be a misleading proxy for model behavior in realistic settings.
More importantly, their evaluations primarily rely on aggregated accuracy metrics, which obscure instance-level variations and how individual predictions change across paraphrases, thereby masking substantial internal brittleness behind averaged results. 
In addition, the paraphrases used in these studies are often generated without rigorous semantic validation, making it unclear whether observed inconsistencies stem from the model itself or from “meaning drift” in the paraphrases. 
Together, these limitations yield an incomplete and coarse-grained understanding of how stable LLM behavior truly is under meaning-preserving rephrasings.
%As a result, current evaluations provide only a partial and coarse-grained view of how stable LLM behavior truly is under meaning-preserving rephrasings.

% To address these gaps, in this work, we perform a rigorous and fine-grained evaluation on LLM consistency through sets of semantically equivalent questions.
% To address these gaps, we show that standard evaluation protocols can systematically obscure substantial instance-level instability under meaning-preserving inputs. We introduce a fine-grained evaluation framework to characterize this hidden failure mode.
To address these gaps, we show that standard evaluation can systematically obscure a critical failure mode: substantial instance-level instability under semantically equivalent inputs. To characterize this phenomenon, we introduce a fine-grained evaluation framework based on sets of rigorously validated paraphrases.
We employ a constrained generation procedure coupled with strict semantic checks to ensure that every variation of a given question is semantically identical. 
We further evaluate models under deterministic decoding to attribute observed variability to the model itself rather than stochastic sampling.
%By evaluating models under deterministic decoding, we isolate inconsistencies that stem from the model's internal representations rather than from stochastic generation. 
Instead of relying solely on aggregated accuracy, we characterize the \textit{answer distribution} induced by paraphrasing: we measure how frequently outputs change across rephrasings, quantify disagreement within each paraphrase set, and track how correctness shifts across different formulations of the same question.
%we analyze the \textit{answer distribution} across paraphrases, examining how often answers change, how predictions disagree within a paraphrase set, and how correctness shifts across different formulations of the same question. 
This instance-level perspective enables a more precise and fine-grained assessment of model stability.

We conduct extensive experiments across 4 datasets and 13 models on factual and mathematical reasoning tasks, and find that even under deterministic decoding and strict semantic-equivalence controls, current LLMs exhibit substantial wording-induced inconsistency. Although changes in aggregated accuracy are relatively modest, typically between 0\% and 5\%, these averages mask significant instability at the instance level. Up to 23\% of questions receive different correctness labels after paraphrasing, and up to 47\% of paraphrase sets contain conflicting predictions. Models frequently answer an original question correctly but fail on semantically equivalent rephrasings, and in many cases succeed on paraphrases after failing on the original input. These results indicate that LLM behavior is far less stable than standard evaluations suggest and highlight the need for more fine-grained consistency analysis.

We also reveal a critical discrepancy between a model's latent knowledge and its observed performance. In many cases, models possess the internal capability to answer correctly, demonstrated by success on at least one paraphrase, yet they often fail to retrieve this knowledge consistently across semantically equivalent inputs. Conversely, limiting evaluation to only consistently correct answers exposes a \textbf{``reliability gap''}, where performance drops significantly below standard accuracy metrics. We formalize these distinctions by defining metrics for maximum (latent), expected, and reliable capability, quantifying the substantial divergences between them to provide a more rigorous view of model robustness and capability. Our core contributions can be summarized as follows:

% \begin{itemize}
% \item We propose a controlled evaluation framework for measuring LLM consistency. The framework relies on rigorously validated sets of semantically equivalent questions and is applicable to free-form generation settings beyond multiple-choice formats.
% \item We conduct a large-scale empirical study of open-weight and proprietary models across four datasets, covering both factual and mathematical question-answering tasks, to systematically examine paraphrase-induced variation in model predictions.
% \item We provide fine-grained analyses of answer distributions, revealing how often answers flip, how paraphrases disagree, and how correctness shifts across different rephrasings of the same input.
% \end{itemize}

\begin{itemize}%[leftmargin=*]
    % \item We propose a controlled evaluation framework for measuring LLM consistency using rigorously validated sets of semantically equivalent questions, enabling fine-grained analysis of prediction distributions beyond aggregate accuracy metrics.
    \item We propose a controlled framework for measuring LLM consistency using rigorously validated semantically equivalent questions, enabling fine-grained analysis of prediction distributions in free-form generation beyond multiple-choice settings and aggregate metrics.

    \item We conduct a large-scale empirical study of open-weight and proprietary models across factual and mathematical QA benchmarks, systematically characterizing paraphrase-induced disagreement, correctness flips, and instability under meaning-preserving inputs.

    \item We identify an under-explored failure mode in LLM evaluation: instance-level instability under meaning-preserving inputs, including correctness flips and a \textbf{reliability gap} between latent capability ($A_{\text{any}}$) and reliable capability ($A_{\text{strict}}$), even when accuracy stays stable.
\end{itemize}

% Our results highlight xxxx

Our results highlight that aggregated accuracy measures often mask and fail to reveal substantial instance-level inconsistencies, and can provide an incomplete and often overly optimistic view of LLM reliability. Models may answer correctly under some paraphrases yet fail on others, revealing a substantial gap between latent capability and reliable performance. These findings underscore the need for consistency-aware evaluation across paraphrased inputs for reliable evaluation of LLM reliability and capability.

% Our contributions can be summarized as follows:
% First, we introduce a controlled evaluation framework for measuring paraphrase consistency in LLMs, built around sets of rigorously validated, semantically equivalent questions and applicable to free-form generation settings, rather than being limited to multiple-choice formats.
% Second, we conduct a large-scale empirical study on 9 open-weight and proprietary models across 4 datasets—including factual and mathematical question-answering tasks—to examine how model predictions change under meaning-preserving rephrasings.
% Third, we provide fine-grained analyses of prediction trajectories, revealing how often answers flip, how paraphrases disagree, and how correctness shifts across different paraphrased versions of the same question.
% Finally, we show that even under greedy decoding, LLMs exhibit substantial inconsistencies that are not captured by standard accuracy metrics, highlighting important gaps in current evaluation practices.

\begin{figure*}[ht]
    \centering
    \includegraphics[width=1\linewidth]{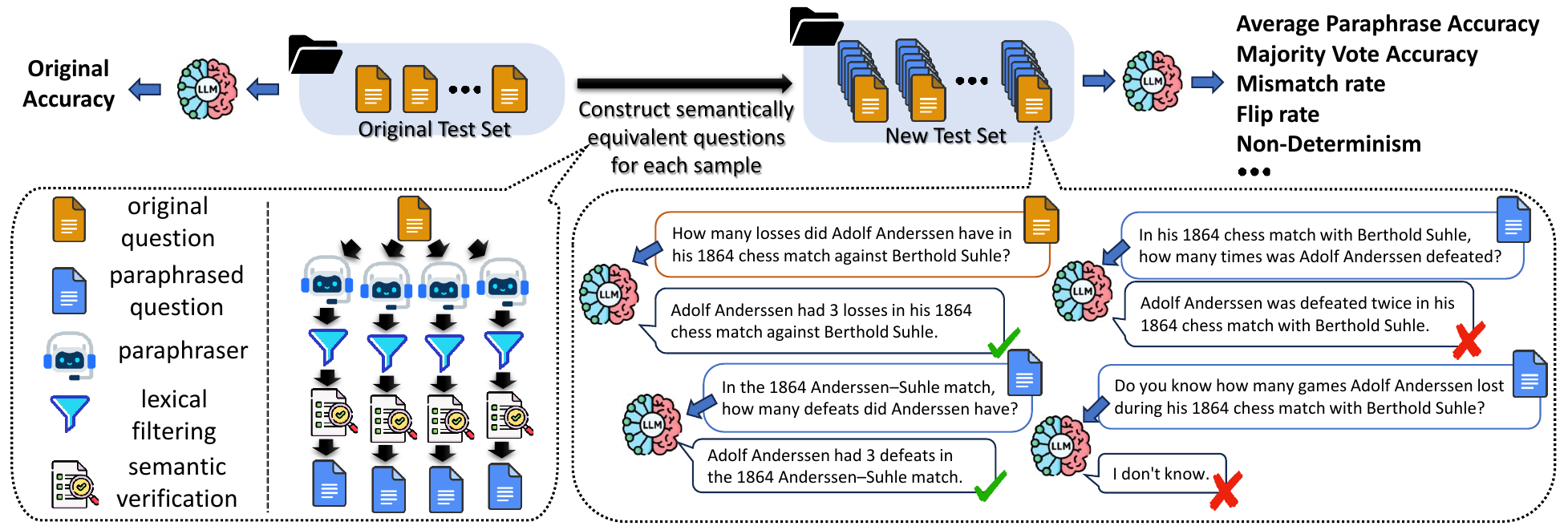}
    % \caption{An overview of of our evaluation framework and an illustration for model inconsistency. For a given benchmark test set, we construct semantically equivalent questions with strict verification for each sample, creating a new test set containing multiple versions of each question. This allows us to perform fine-grained consistency evaluation using our metric suite. The bubble box illustrates a partial example from SimpleQA where the model responds differently across multiple paraphrases, underscoring instance-level inconsistency.}
    \caption{
    Overview of the evaluation framework. For each input question, we generate multiple semantically equivalent paraphrases using controlled generation and verification, and evaluate model responses across them. The example illustrates how predictions vary across paraphrases, revealing instance-level instability despite semantically equivalent inputs.
    }
    \label{fig:teaser}
\end{figure*}
\section{Related Works}

Consistency under meaning-preserving transformations is a core requirement for reliable NLP systems. \citet{ribeiro2020beyond} formalized this principle through invariance tests, yet empirical work shows it is frequently violated: \citet{gan2019improving} observed substantial accuracy drops on paraphrased SQuAD questions, and \citet{elazar2021measuring} showed that pretrained language models often answer paraphrased factual queries inconsistently. 
More recent studies suggest that standard evaluations can further obscure this instability. 
For example, \citet{burnell2023rethink} argue that reporting only aggregate evaluation metrics makes it difficult to understand when and why models fail, advocating more granular instance-level analyses. Similarly, \citet{lunardi2025robustness} found that paraphrasing benchmark questions leads to significant performance drops while largely preserving leaderboard rankings, and \citet{alzahrani2024benchmarks} showed that even small benchmark perturbations can affect leaderboard outcomes. Related work also argues that single-prompt evaluation is unreliable \cite{polo2024efficient, mizrahi2024state}, and \citet{sclar2023quantifying} demonstrated that trivial, meaning-preserving prompt formatting changes can induce swings of up to 76 accuracy points. \citet{choi2025roparq} introduced RoParQ to measure cross-paraphrase consistency in multiple-choice QA. 
However, most of these studies either focus input variation on prompt formats and benchmark structures, or constrain their setting to multiple-choice or cloze tasks rather than free-form generation, without exposing the instance-level instability masked by aggregate metrics.

Parallel work has also explored robustness and surface-form sensitivity in reasoning and knowledge tasks. \citet{zhou2024paraphrase} and \citet{han2025analysis} showed that paraphrasing math problems can substantially alter LLM performance, while \citet{meier2025towards} found deeper syntactic paraphrases particularly challenging. GSM-Symbolic \cite{mirzadeh2024gsm} further showed that minor symbolic changes or logically irrelevant clauses can cause large performance variance. Other work proposes broader consistency and robustness frameworks \cite{nalbandyan2025score, raj2025improving, ghosh2024logical}. However, these studies typically measure aggregated accuracy changes or rely on task-specific perturbations. 
In contrast, we systematically analyze instance-level answer distributions across semantically equivalent inputs obtained via task-agnostic paraphrasing.
% In contrast to prior work that primarily focuses on aggregate robustness, we analyze instance-level prediction distributions across semantically equivalent inputs, revealing instability hidden by standard metrics.

% Parallel work on reasoning tasks reports similar surface-form sensitivity. \citet{zhou2024paraphrase} and \citet{han2025analysis} showed that paraphrasing math problems can substantially alter LLM performance, while \citet{meier2025towards} found deeper syntactic paraphrases particularly challenging. GSM-Symbolic \cite{mirzadeh2024gsm} further showed that minor symbolic changes or logically irrelevant clauses can cause large performance variance. Other work proposes broader consistency and robustness frameworks \cite{nalbandyan2025score, raj2025improving, ghosh2024logical}. In contrast, we analyze instance-level answer distributions over rigorously validated semantic paraphrase sets under deterministic decoding, directly probing paraphrastic fragility in free-form generation.
% \vspace{-5pt}
\section{A Framework for Fine-Grained Evaluation of Consistency and Reliability}
\label{sec:method}

We introduce a framework to evaluate the reliability of LLM by measuring their invariance to non-adversarial paraphrasing. Unlike adversarial perturbations, which are designed to induce failure, our transformations are strictly meaning-preserving and reflect natural linguistic variation. Our approach consists of two core components: a rigorous pipeline for generating semantically equivalent test sets, and a suite of metrics defined to quantify consistency beyond standard accuracy or aggregated accuracy. An overview of the framework is shown in Figure~\ref{fig:teaser}.

\subsection{Constructing Semantically Equivalent Test Sets}
\label{sec:paraphrase-protocal}
To isolate the effect of surface form on model predictions, we require sets of inputs that differ in wording but are semantically identical. For a given original query $Q^0$ from a standard benchmark, we generate a set of $k$ paraphrases $\mathcal{P} = \{Q^1, \dots, Q^k\}$ using a three-stage validation process designed to prevent meaning drift:

% \paragraph{1. Controlled Generation.} 
% We utilize a high-capability instruction-tuned model, specifically, \texttt{GPT-4.1-mini}, to generate diverse rephrasings of $Q^0$. The generation prompt explicitly constrains the model to preserve the original intent, key assumptions, and required information format while varying sentence structure and vocabulary. Some other details of this part, along with the paraphrase generation prompt is provided in Appendix~\ref{app:paraphrase-generation}.

\paragraph{(i). Controlled Generation.}

For each original question $Q^0$ (\raisebox{-0.2em}{\includegraphics[height=1em]{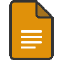}} in Figure.~\ref{fig:teaser}), we generate multiple semantically equivalent paraphrases (\raisebox{-0.2em}{\includegraphics[height=1em]{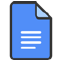}} in Figure.~\ref{fig:teaser}) using a high-capability instruction-tuned model, \texttt{GPT-4.1-mini}. To introduce controlled linguistic variability while preserving the original intent, assumptions, and required information, we condition generation on lightweight cues that encourage either neutral structural rephrasing or stylistic variation (e.g., tone). Full prompt templates and implementation details are provided in Appendix~\ref{app:paraphrase-generation}.

\paragraph{(ii). Lexical-Overlap Filtering.} 
To avoid paraphrases that differ only superficially from the original query (e.g., via punctuation changes or near-verbatim rewrites), we filter candidates based on lexical overlap. We define lexical overlap as the proportion of shared word tokens between a paraphrase and the original query, normalized by the length of the shorter text. Only paraphrases with lexical overlap below 0.8 are retained.

\paragraph{(iii). Dual-Stage Semantic Verification.} 
To ensure strict equivalence, we employ two filters in independent steps. First, we use an LLM judge, specifically \texttt{GPT-4.1-mini}, to verify that each $Q^i$ asks for the exact same information as $Q^0$, capturing subtle shifts in constraints or hallucinations. Second, an NLI-based filter (\texttt{facebook/bart-large-mnli}) checks for bidirectional entailment between $Q^0$ and each candidate $Q^i$ (threshold 0.75) to ensure all information and constraints remain intact. Candidates failing either check are discarded. The instruction to the LLM judge is shown in Appendix~\ref{app:judge-equivalence}.

Importantly, this controlled construction provides a conservative lower bound on real-world linguistic variability. If models exhibit instability even under strictly validated, meaning-preserving paraphrases, such effects are likely to be at least as pronounced under the broader and noisier distribution of real user queries.

% Separately, to empirically validate our semantic verification pipeline, we assess the reliability of our approach via human annotation on a subset of $\sim$1250 generated paraphrase pairs across datasets. Annotators were asked whether two questions “ask the exact same thing” (i.e., can be used interchangeably without changing the answer). Agreement between the automatic pipeline and human judgments ranges from 92\%–98\%, confirming that retained paraphrases preserve meaning. As a further robustness check, we evaluate a stricter multi-judge validation protocol in Appendix~\ref{app:validation}, which yields even higher agreement while leaving all conclusions unchanged. Additional details on the annotation protocol, per-dataset agreement, and multi-judge validation are provided in Appendix~\ref{app:validation}.

To validate our semantic verification pipeline, we conduct human annotation on $\sim$1250 paraphrase pairs, where annotators judge whether two questions “ask the exact same thing,” achieving 92–98\% agreement across datasets. Stricter multi-judge validation (Appendix~\ref{app:validation}) yields even higher agreement with unchanged conclusions.

\subsection{Metric Suite for Fine-grained Consistency Quantification}
\label{sec:metrics_def}
Let $\mathcal{Q}_i = \{Q_i^0, Q_i^1, \dots, Q_i^{k_i}\}$ be the set of validated queries for the $i$-th instance. We obtain the model's responses using deterministic decoding to ensure that observed variations stem from the model itself rather than sampling noise. 
% The raw outputs are mapped to a discrete label space $\mathcal{L}$ (e.g., \{Correct, Incorrect, Abstain\} or normalized numerical values). Consequently, for each question, the paraphrased variants induce an empirical distribution $P_i^l$ ($l \in \mathcal{L}$) over these labels. We define the following metrics to capture different granularities of consistency.
Each response is mapped to a discrete label space $\mathcal{L}$ (e.g., $\{\text{Correct}, \text{Incorrect}, \text{Abstain}\}$ or normalized numerical answers), inducing an empirical label distribution
$P_i = \{P_i^l\}_{l \in \mathcal{L}}$ over the paraphrases of question $i$.
We define the following metrics to capture different granularities of consistency:

\subsubsection{Accuracy Variants}
Standard evaluations report accuracy on a single prompt. We contrast this with metrics that account for semantic variation. Denote the number of query sets in the new test set as $N$, we report:
\begin{itemize}%[leftmargin=*]
    \item \textbf{Original Accuracy ($A_{\text{orig}}$):} The standard correctness on the original question $Q_i^0$, calculated as $A_{\text{orig}}=\frac{1}{N}\sum_i \mathbbm{1}[Q_i^0 \text{ is correct}]$
    \item \textbf{Average Paraphrase Accuracy ($A_{\text{dist}}$):} The expected accuracy over the distribution of paraphrases, calculated as $A_{\text{dist}}=\frac{1}{N}\sum_i\left(\frac{1}{|\mathcal{Q}_i|} \sum_{j} \mathbbm{1}[Q_i^j \text{ is correct}]\right)$.
    \item \textbf{Majority Vote Accuracy ($A_{\text{MV}}$):} The correctness of the label that appears most frequently across all variations in $\mathcal{Q}_i$, calculated as $A_{\text{MV}}=\frac{1}{N}\sum_i \mathbbm{1}[\text{MV}(Q_i)\text{ is correct}]$, where $\text{MV}(Q_i)$ is the majority vote prediction across all variations in $Q_i$.
\end{itemize}

\subsubsection{Instance-Level Stability}
\label{sec:instance-level-stability}
% To measure how often predictions shift for individual questions, we define:
% To quantify how predictions shift across paraphrases for individual questions, we define:
% \begin{itemize}
%     \item \textbf{Mismatch Rate:} The proportion of questions where the prediction on the original wording contradicts the majority vote prediction on the paraphrases.
To quantify how model predictions vary across paraphrases of the same question, we define:

\begin{itemize}%[leftmargin=*]
    \item \textbf{Mismatch Rate:} The proportion of questions where the prediction on the original wording differs from the majority vote prediction across paraphrases.
    % \item \textbf{Prediction Flip Rate:} We measure conditional probabilities of failure, specifically $P(\neg A_{\text{para}} \mid A_{\text{orig}})$, defined as the proportion of cases where the prediction on the original question was correct, but the majority vote prediction on the paraphrases were incorrect,  and $P(A_{\text{para}} \mid \neg A_{\text{orig}})$, defined as the proportion of cases where the majority vote prediction on paraphrases was correct, but the prediction on the original question was incorrect.
    \item \textbf{Prediction Flip Rate:} We measure directional instability via the conditional probabilities 
$P(\neg A_{\text{para}} \mid A_{\text{orig}})$ and 
$P(A_{\text{para}} \mid \neg A_{\text{orig}})$, where $A_{\text{para}}$ denotes correctness of the majority-vote prediction across paraphrases. These capture when correct original predictions become incorrect under paraphrasing, and when incorrect predictions are recovered, respectively.

    % \item \textbf{Normalized Entropy:} 
    % We quantify dispersion in the label distribution via a length-normalized entropy:
    % \[
    %     H_i \;=\; -\frac{1}{\log(k_i)} \sum_{l \in \mathcal{L}} P_i^l \log P_i^l .
    % \]
    % Higher entropy indicates greater variability in predictions across paraphrases, while values near zero correspond to near-deterministic behavior.

    % \item \textbf{IID-Mismatch Rate:} 
    % Defined as
    % \[
    %     1 - \sum_{l \in \mathcal{L}} (P_i^l)^2,
    % \]
    % this metric measures the probability that two independently sampled paraphrases of the same question receive different labels, capturing pairwise inconsistency in model behavior.

    % \item \textbf{Average Mode Share:} 
    % For each question, we define the mode share as
    % \[
    %     \max_{l \in \mathcal{L}} P_i^l,
    % \]
    % i.e., the proportion of paraphrases assigned to the most frequent label, reflecting the dominance of the model’s primary prediction.
    % We report the average mode share by averaging this quantity across all instances in the dataset.
    % \item \textbf{Non-Determinism (ND) Rate:} The fraction of sets $\mathcal{Q}_i$ where the model produces at least two distinct labels (i.e., absolute disagreement), regardless of correctness.
    \item \textbf{Non-Determinism (ND) Rate:} The fraction of paraphrase sets $\mathcal{Q}_i$ where the model produces at least two distinct labels, indicating disagreement across semantically equivalent inputs.

\end{itemize}
We report additional distributional stability metrics (e.g., entropy, IID mismatch, and mode share) in Appendix~\ref{app:stability-metrics}.

\subsubsection{Latent Knowledge Boundaries}
We bound the model's capabilities by distinguishing between reliable knowledge and potential knowledge:
\begin{itemize}%[leftmargin=*]
    \item \textbf{Maximum Capability ($A_{\text{any}}$):} The proportion of instances where the model generates the correct answer for \textit{at least one} variation in $\mathcal{Q}_i$. This quantity, which we also refer to as \emph{best-case accuracy}, approximates the upper bound of the model's latent knowledge. we provide a brief justification in Appendix~\ref{app:latent_knowledge_any}.
    \item \textbf{Reliable Capability ($A_{\text{strict}}$):} The proportion of instances where the model answers correctly for \textit{all} variations in $\mathcal{Q}_i$, representing robust and reliable knowledge. We also refer to this as \emph{reliable} or \emph{conservative} accuracy.
\end{itemize}
We refer to the discrepancy between latent capability ($A_{\text{any}}$) and reliable capability ($A_{\text{strict}}$) as the \textbf{reliability gap}.

\section{Experiments}
\label{sec:experiments}

\subsection{Experimental Setup}

\paragraph{Datasets} 
To ensure a comprehensive evaluation across distinct modalities of reasoning, we utilize four diverse benchmarks: \textsc{SimpleQA}~\cite{wei2024measuringshortformfactualitylarge} and \textsc{TruthfulQA}~\cite{lin2022truthfulqameasuringmodelsmimic} for factual knowledge, and \textsc{GSM8K}~\cite{cobbe2021trainingverifierssolvemath} and \textsc{Math-500}~\cite{hendrycks2021measuringmathematicalproblemsolving} for mathematical reasoning. For efficiency, we evaluate on random subsets of each dataset.\footnote{Sample sizes: \textsc{SimpleQA} ($n=1129$), \textsc{TruthfulQA} ($n=511$), \textsc{GSM8K} ($n=989$), and \textsc{Math-500} ($n=152$).} 

For the factual QA datasets, we adopt a three-way grading rubric: \textbf{Correct}, \textbf{Incorrect}, and \textbf{Not Attempted}. For mathematical tasks, we apply normalization to treat equivalent numerical forms (e.g., $0.25$ and $1/4$) as identical canonical labels. Evaluation is designed to be consistent across paraphrases, so differences reflect model behavior rather than evaluation artifacts (Appendix~\ref{app:evaluation}).

\paragraph{Models}
We evaluate a broad spectrum of 13 models spanning both open-weight and proprietary architectures to analyze the impact of scale and training methodology. Our open-weight set includes \texttt{Llama-3-8B} and \texttt{Llama-3.1-8B}~\cite{grattafiori2024llama3herdmodels}, \texttt{Qwen-2.5-7B}~\cite{qwen2025qwen25technicalreport}, and six variants from the \texttt{Qwen 3} family~\cite{yang2025qwen3technicalreport}, covering model sizes from 0.6B to 32B parameters. In addition, we evaluate four proprietary models from the GPT family: \texttt{GPT-3.5-Turbo}~\cite{brown2020languagemodelsfewshotlearners}, \texttt{GPT-4o}, \texttt{GPT-4.1}, and \texttt{GPT-4.1-Mini}~\cite{openai2024gpt4technicalreport}.

% \paragraph{Answer Evaluation.}
% For factual QA datasets, model responses are mapped to a three-way label space (Correct, Incorrect, Not Attempted). We employ an LLM-based evaluation protocol that follows official dataset criteria when available. To ensure robustness and reduce model-specific bias, we use a multi-judge evaluation framework with multiple independent judge models, where final labels are determined via agreement across judges. This setup ensures that evaluation is not dependent on a single judge model. Consistent prompting and normalization are applied across paraphrases to ensure that differences in labels reflect model behavior rather than evaluation artifacts. Additional details, prompts, and judge configurations are provided in Appendix~\ref{app:evaluation}.

\begin{table*}[t]
\centering

\caption{Accuracies and correctness rates across datasets in standard original accuracy (orig), average paraphrase accuracy (dist), and majority vote accuracy (MV). 
For Dist and MV, superscripts indicate change relative to the original wording; \textcolor[rgb]{0,0.502,0}{green}/\textcolor[rgb]{0.8,0,0}{red} denote improvements/degradations exceeding 1 percentage point.}
\label{tab:consistency_across_datasets}

\resizebox{0.95\linewidth}{!}{%
\begin{tabular}{lcccccccccccc}
\toprule

\multicolumn{1}{c}{\multirow{2}{*}{\textbf{Model}}}
& \multicolumn{3}{c}{\textbf{SimpleQA}}
& \multicolumn{3}{c}{\textbf{TruthfulQA}}
& \multicolumn{3}{c}{\textbf{GSM8K}}
& \multicolumn{3}{c}{\textbf{MATH-500}} \\

\cmidrule(lr){2-4}
\cmidrule(lr){5-7}
\cmidrule(lr){8-10}
\cmidrule(lr){11-13}

\multicolumn{1}{c}{}
% & \textbf{Original} & \textbf{Dist} & \textbf{MV}
% & \textbf{Original} & \textbf{Dist} & \textbf{MV}
% & \textbf{Original} & \textbf{Dist} & \textbf{MV}
% & \textbf{Original} & \textbf{Dist} & \textbf{MV} \\

% & \textbf{$Acc_{orig}$} & \textbf{$Acc_{dist}$} & \textbf{$Acc_{MV}$}
% & \textbf{$Acc_{orig}$} & \textbf{$Acc_{dist}$} & \textbf{$A_{MV}$}
% & \textbf{$Acc_{orig}$} & \textbf{$Acc_{dist}$} & \textbf{$A_{MV}$}
% & \textbf{$Acc_{orig}$} & \textbf{$Acc_{dist}$} & \textbf{$A_{MV}$} \\

& \textbf{$A_{\text{orig}}$} & \textbf{$A_{\text{dist}}$} & \textbf{$A_{\text{MV}}$}
& \textbf{$A_{\text{orig}}$} & \textbf{$A_{\text{dist}}$} & \textbf{$A_{\text{MV}}$}
& \textbf{$A_{\text{orig}}$} & \textbf{$A_{\text{dist}}$} & \textbf{$A_{\text{MV}}$}
& \textbf{$A_{\text{orig}}$} & \textbf{$A_{\text{dist}}$} & \textbf{$A_{\text{MV}}$} \\

\midrule

GPT-3.5 & 9.1 & 7.1\textsuperscript{\textcolor[rgb]{0.8,0,0}{-2.0}} & 6.8\textsuperscript{\textcolor[rgb]{0.8,0,0}{-2.3}} & 52.3 & 51.0\textsuperscript{\textcolor[rgb]{0.8,0,0}{-1.3}} & 52.7\textsuperscript{+0.5} & 80.6 & 77.3\textsuperscript{\textcolor[rgb]{0.8,0,0}{-3.3}} & 82.1\textsuperscript{\textcolor[rgb]{0,0.502,0}{+1.5}} & 44.8 & 43.3\textsuperscript{\textcolor[rgb]{0.8,0,0}{-1.6}} & 42.8\textsuperscript{\textcolor[rgb]{0.8,0,0}{-2.1}} \\
GPT-4.1 & 38.3 & 37.4\textsuperscript{-0.9} & 38.4\textsuperscript{+0.1} & 76.3 & 73.7\textsuperscript{\textcolor[rgb]{0.8,0,0}{-2.6}} & 75.1\textsuperscript{\textcolor[rgb]{0.8,0,0}{-1.2}} & 94.0 & 89.7\textsuperscript{\textcolor[rgb]{0.8,0,0}{-4.3}} & 93.6\textsuperscript{-0.4} & 94.4 & 90.5\textsuperscript{\textcolor[rgb]{0.8,0,0}{-3.9}} & 90.9\textsuperscript{\textcolor[rgb]{0.8,0,0}{-3.5}} \\
GPT-4.1-Mini & 15.7 & 15.2\textsuperscript{-0.5} & 15.1\textsuperscript{-0.6} & 70.6 & 68.2\textsuperscript{\textcolor[rgb]{0.8,0,0}{-2.4}} & 69.9\textsuperscript{-0.8} & 94.9 & 89.6\textsuperscript{\textcolor[rgb]{0.8,0,0}{-5.3}} & 93.5\textsuperscript{\textcolor[rgb]{0.8,0,0}{-1.4}} & 95.9 & 91.5\textsuperscript{\textcolor[rgb]{0.8,0,0}{-4.4}} & 93.2\textsuperscript{\textcolor[rgb]{0.8,0,0}{-2.7}} \\
GPT-4o & 19.8 & 17.4\textsuperscript{\textcolor[rgb]{0.8,0,0}{-2.4}} & 16.3\textsuperscript{\textcolor[rgb]{0.8,0,0}{-3.4}} & 56.6 & 55.8\textsuperscript{-0.8} & 56.3\textsuperscript{-0.3} & 94.0 & 89.2\textsuperscript{\textcolor[rgb]{0.8,0,0}{-4.9}} & 93.2\textsuperscript{-0.8} & 79.2 & 78.3\textsuperscript{-0.9} & 79.2\textsuperscript{+0.0} \\
LLaMA-3 8B & 3.5 & 3.1\textsuperscript{-0.4} & 2.9\textsuperscript{-0.7} & 28.8 & 29.6\textsuperscript{+0.9} & 28.8\textsuperscript{+0.0} & 77.8 & 71.4\textsuperscript{\textcolor[rgb]{0.8,0,0}{-6.4}} & 74.5\textsuperscript{\textcolor[rgb]{0.8,0,0}{-3.3}} & 30.0 & 26.6\textsuperscript{\textcolor[rgb]{0.8,0,0}{-3.4}} & 26.4\textsuperscript{\textcolor[rgb]{0.8,0,0}{-3.6}} \\
LLaMA-3.1 8B & 0.9 & 1.0\textsuperscript{+0.1} & 0.9\textsuperscript{+0.0} & 30.9 & 28.8\textsuperscript{\textcolor[rgb]{0.8,0,0}{-2.1}} & 26.8\textsuperscript{\textcolor[rgb]{0.8,0,0}{-4.2}} & 84.9 & 80.0\textsuperscript{\textcolor[rgb]{0.8,0,0}{-4.9}} & 83.8\textsuperscript{\textcolor[rgb]{0.8,0,0}{-1.1}} & 49.2 & 52.4\textsuperscript{\textcolor[rgb]{0,0.502,0}{+3.2}} & 53.0\textsuperscript{\textcolor[rgb]{0,0.502,0}{+3.8}} \\
Qwen-2.5 7B & 0.4 & 0.5\textsuperscript{+0.0} & 0.3\textsuperscript{-0.2} & 44.8 & 43.3\textsuperscript{\textcolor[rgb]{0.8,0,0}{-1.5}} & 43.5\textsuperscript{\textcolor[rgb]{0.8,0,0}{-1.3}} & 91.0 & 84.8\textsuperscript{\textcolor[rgb]{0.8,0,0}{-6.2}} & 88.8\textsuperscript{\textcolor[rgb]{0.8,0,0}{-2.2}} & 75.0 & 71.5\textsuperscript{\textcolor[rgb]{0.8,0,0}{-3.5}} & 74.3\textsuperscript{-0.7} \\
Qwen-3 0.6B & 0.0 & 0.0\textsuperscript{+0.0} & 0.0\textsuperscript{+0.0} & 20.4 & 21.3\textsuperscript{+0.9} & 19.0\textsuperscript{\textcolor[rgb]{0.8,0,0}{-1.5}} & 62.7 & 58.3\textsuperscript{\textcolor[rgb]{0.8,0,0}{-4.4}} & 60.3\textsuperscript{\textcolor[rgb]{0.8,0,0}{-2.3}} & 58.9 & 55.4\textsuperscript{\textcolor[rgb]{0.8,0,0}{-3.5}} & 54.3\textsuperscript{\textcolor[rgb]{0.8,0,0}{-4.7}} \\
Qwen-3 1.7B & 1.1 & 0.9\textsuperscript{-0.2} & 0.6\textsuperscript{-0.5} & 16.9 & 18.3\textsuperscript{\textcolor[rgb]{0,0.502,0}{+1.4}} & 16.9\textsuperscript{+0.0} & 83.1 & 77.4\textsuperscript{\textcolor[rgb]{0.8,0,0}{-5.7}} & 81.4\textsuperscript{\textcolor[rgb]{0.8,0,0}{-1.7}} & 80.7 & 73.5\textsuperscript{\textcolor[rgb]{0.8,0,0}{-7.3}} & 76.3\textsuperscript{\textcolor[rgb]{0.8,0,0}{-4.4}} \\
Qwen-3 4B & 1.7 & 1.6\textsuperscript{-0.1} & 1.3\textsuperscript{-0.4} & 35.7 & 38.6\textsuperscript{\textcolor[rgb]{0,0.502,0}{+2.9}} & 37.8\textsuperscript{\textcolor[rgb]{0,0.502,0}{+2.1}} & 90.8 & 85.6\textsuperscript{\textcolor[rgb]{0.8,0,0}{-5.2}} & 90.7\textsuperscript{-0.1} & 90.5 & 87.0\textsuperscript{\textcolor[rgb]{0.8,0,0}{-3.5}} & 89.1\textsuperscript{\textcolor[rgb]{0.8,0,0}{-1.5}} \\
Qwen-3 8B & 3.0 & 2.0\textsuperscript{-0.9} & 1.5\textsuperscript{\textcolor[rgb]{0.8,0,0}{-1.4}} & 47.4 & 49.9\textsuperscript{\textcolor[rgb]{0,0.502,0}{+2.5}} & 50.8\textsuperscript{\textcolor[rgb]{0,0.502,0}{+3.4}} & 93.2 & 87.0\textsuperscript{\textcolor[rgb]{0.8,0,0}{-6.3}} & 91.6\textsuperscript{\textcolor[rgb]{0.8,0,0}{-1.7}} & 81.1 & 82.3\textsuperscript{\textcolor[rgb]{0,0.502,0}{+1.1}} & 82.5\textsuperscript{\textcolor[rgb]{0,0.502,0}{+1.4}} \\
Qwen-3 14B & 5.0 & 3.7\textsuperscript{\textcolor[rgb]{0.8,0,0}{-1.2}} & 3.6\textsuperscript{\textcolor[rgb]{0.8,0,0}{-1.4}} & 62.6 & 60.8\textsuperscript{\textcolor[rgb]{0.8,0,0}{-1.8}} & 63.1\textsuperscript{+0.5} & 94.6 & 89.0\textsuperscript{\textcolor[rgb]{0.8,0,0}{-5.6}} & 93.1\textsuperscript{\textcolor[rgb]{0.8,0,0}{-1.4}} & 94.4 & 89.3\textsuperscript{\textcolor[rgb]{0.8,0,0}{-5.1}} & 90.2\textsuperscript{\textcolor[rgb]{0.8,0,0}{-4.2}} \\
Qwen-3 32B & 4.2 & 4.0\textsuperscript{-0.2} & 3.5\textsuperscript{-0.7} & 65.2 & 61.3\textsuperscript{\textcolor[rgb]{0.8,0,0}{-3.9}} & 62.4\textsuperscript{\textcolor[rgb]{0.8,0,0}{-2.8}} & 94.2 & 88.9\textsuperscript{\textcolor[rgb]{0.8,0,0}{-5.4}} & 92.4\textsuperscript{\textcolor[rgb]{0.8,0,0}{-1.9}} & 92.5 & 87.7\textsuperscript{\textcolor[rgb]{0.8,0,0}{-4.8}} & 90.4\textsuperscript{\textcolor[rgb]{0.8,0,0}{-2.1}} \\

\bottomrule
\end{tabular}
}

\end{table*}
\begin{table*}[t]
\centering

% \caption{Hidden instability beyond accuracy. For each dataset, we report (i) the change in majority-vote (MV) accuracy relative to the original accuracy ($\Delta\text{Acc}_\text{MV}=A_{\text{MV}}-A_{\text{orig}}$), and (ii) the mismatch rate: the fraction of instances where the original prediction differs from the MV prediction. Large mismatch with small $\Delta$MV Acc indicates large instance-level instability masked by aggregated accuracies.}
% \caption{Hidden instability beyond accuracy. For each dataset, we report 
% (i) the change in majority-vote (MV) accuracy relative to the original accuracy 
% ($\Delta A_{\text{MV}} = A_{\text{MV}} - A_{\text{orig}}$), 
% (ii) the mismatch rate: the fraction of instances where the original prediction 
% differs from the MV prediction, and 
% (iii) $P(\text{ND})$, the fraction of input sets where the model produces at least 
% two distinct labels across paraphrases (non-deterministic predictions). 
% Large mismatch rate or $P(\text{ND})$ with small $\Delta A_{\text{MV}}$ indicates substantial instance-level 
% instability masked by aggregated accuracy.}
\caption{Hidden instability beyond accuracy. We report 
$\Delta A_{\text{MV}}$, mismatch rate (original vs. MV disagreement), and 
$P(\text{ND})$ (paraphrase sets with multiple labels). 
High mismatch or $P(\text{ND})$ with small $\Delta A_{\text{MV}}$ reveals substantial hidden instability.}
\label{tab:hidden_instability}

\resizebox{0.95\linewidth}{!}{%
% \begin{tabular}{lcccccccc}
% \toprule

% \multicolumn{1}{c}{\multirow{2}{*}{\textbf{Model}}}
% & \multicolumn{2}{c}{\textbf{SimpleQA}}
% & \multicolumn{2}{c}{\textbf{TruthfulQA}}
% & \multicolumn{2}{c}{\textbf{GSM8K}}
% & \multicolumn{2}{c}{\textbf{MATH-500}} \\

% \cmidrule(lr){2-3}
% \cmidrule(lr){4-5}
% \cmidrule(lr){6-7}
% \cmidrule(lr){8-9}

% \multicolumn{1}{c}{}
% & \textbf{$\Delta\text{Acc}_\text{MV}$} & \textbf{Mismatch Rate}
% & \textbf{$\Delta\text{Acc}_\text{MV}$} & \textbf{Mismatch Rate}
% & \textbf{$\Delta\text{Acc}_\text{MV}$} & \textbf{Mismatch Rate}
% & \textbf{$\Delta\text{Acc}_\text{MV}$} & \textbf{Mismatch Rate} \\
\begin{tabular}{l|ccc|ccc|ccc|ccc}
\toprule

\multicolumn{1}{c}{\multirow{2}{*}{\textbf{Model}}}
& \multicolumn{3}{c}{\textbf{SimpleQA}}
& \multicolumn{3}{c}{\textbf{TruthfulQA}}
& \multicolumn{3}{c}{\textbf{GSM8K}}
& \multicolumn{3}{c}{\textbf{MATH-500}} \\

\cmidrule(lr){2-4}
\cmidrule(lr){5-7}
\cmidrule(lr){8-10}
\cmidrule(lr){11-13}

% \multicolumn{1}{c}{}
% & $\Delta\text{Acc}_{\text{MV}}$ & Mismatch Rate & $P(\text{ND})$
% & $\Delta\text{Acc}_{\text{MV}}$ & Mismatch Rate & $P(\text{ND})$
% & $\Delta\text{Acc}_{\text{MV}}$ & Mismatch Rate & $P(\text{ND})$
% & $\Delta\text{Acc}_{\text{MV}}$ & Mismatch Rate & $P(\text{ND})$ \\
\multicolumn{1}{c}{}
& $\Delta A_{\text{MV}}$ & \makecell[c]{Mismatch\\Rate} & $P(\text{ND})$
& $\Delta A_{\text{MV}}$ & \makecell[c]{Mismatch\\Rate} & $P(\text{ND})$
& $\Delta A_{\text{MV}}$ & \makecell[c]{Mismatch\\Rate} & $P(\text{ND})$
& $\Delta A_{\text{MV}}$ & \makecell[c]{Mismatch\\Rate} & $P(\text{ND})$ \\

% \midrule

% % ==== paste model rows here ====
% % ModelName & sqa_dmv & sqa_mismatch & tqa_dmv & tqa_mismatch & gsm_dmv & gsm_mismatch & math_dmv & math_mismatch \\

% GPT-3.5 & -2.33 & 21.80 & 0.48 & 16.47 & 1.51 & 12.70 & -2.07 & 17.24 \\
% GPT-4.1 & 0.10 & 13.94 & -1.23 & 9.14 & -0.41 & 3.09 & -3.50 & 4.90 \\
% GPT-4.1-Mini & -0.58 & 6.93 & -0.76 & 11.14 & -1.43 & 2.66 & -2.74 & 4.11 \\
% GPT-4o & -3.44 & 19.12 & -0.25 & 16.41 & -0.82 & 2.88 & 0.00 & 9.72 \\
% LLaMA-3 8B & -0.67 & 15.68 & 0.00 & 17.17 & -3.33 & 13.68 & -3.57 & 12.14 \\
% LLaMA-3.1 8B & 0.00 & 2.60 & -4.16 & 18.96 & -1.09 & 10.64 & 3.79 & 21.97 \\
% Qwen-2.5 7B & -0.18 & 1.78 & -1.31 & 17.54 & -2.18 & 6.13 & -0.69 & 7.64 \\
% Qwen-3 0.6B & 0.00 & 9.13 & -1.46 & 21.87 & -2.35 & 17.70 & -4.65 & 23.26 \\
% Qwen-3 1.7B & -0.49 & 21.83 & 0.00 & 23.12 & -1.69 & 11.40 & -4.44 & 11.85 \\
% Qwen-3 4B & -0.40 & 20.30 & 2.08 & 19.01 & -0.10 & 5.67 & -1.46 & 7.30 \\
% Qwen-3 8B & -1.44 & 16.31 & 3.40 & 15.71 & -1.66 & 4.78 & 1.40 & 8.39 \\
% Qwen-3 14B & -1.39 & 16.27 & 0.54 & 11.65 & -1.44 & 3.29 & -4.20 & 5.59 \\
% Qwen-3 32B & -0.68 & 15.04 & -2.84 & 13.66 & -1.86 & 4.12 & -2.05 & 6.16 \\

% \bottomrule

\midrule
GPT-3.5 & -2.33 & 21.80 & 39.83 & 0.48 & 16.47 & 30.79 & 1.51 & 12.70 & 31.65 & -2.07 & 17.24 & 28.28 \\
GPT-4.1 & 0.10 & 13.94 & 29.71 & -1.23 & 9.14 & 20.99 & -0.41 & 3.09 & 15.05 & -3.50 & 4.90 & 7.69 \\
GPT-4.1-Mini & -0.58 & 6.93 & 18.00 & -0.76 & 11.14 & 25.32 & -1.43 & 2.66 & 15.93 & -2.74 & 4.11 & 7.53 \\
GPT-4o & -3.44 & 19.12 & 33.84 & -0.25 & 16.41 & 28.28 & -0.82 & 2.88 & 16.44 & 0.00 & 9.72 & 11.81 \\
LLaMA-3 8B & -0.67 & 15.68 & 32.89 & 0.00 & 17.17 & 36.36 & -3.33 & 13.68 & 34.02 & -3.57 & 12.14 & 24.29 \\
LLaMA-3.1 8B & 0.00 & 2.60 & 4.22 & -4.16 & 18.96 & 29.87 & -1.09 & 10.64 & 28.45 & 3.79 & 21.97 & 31.82 \\
Qwen-2.5 7B & -0.18 & 1.78 & 4.98 & -1.31 & 17.54 & 33.25 & -2.18 & 6.13 & 22.12 & -0.69 & 7.64 & 18.06 \\
Qwen-3 0.6B & 0.00 & 9.13 & 8.23 & -1.46 & 21.87 & 41.11 & -2.35 & 17.70 & 47.55 & -4.65 & 23.26 & 27.13 \\
Qwen-3 1.7B & -0.49 & 21.83 & 47.17 & 0.00 & 23.12 & 39.22 & -1.69 & 11.40 & 31.68 & -4.44 & 11.85 & 21.48 \\
Qwen-3 4B & -0.40 & 20.30 & 43.08 & 2.08 & 19.01 & 47.40 & -0.10 & 5.67 & 23.09 & -1.46 & 7.30 & 14.60 \\
Qwen-3 8B & -1.44 & 16.31 & 35.60 & 3.40 & 15.71 & 31.41 & -1.66 & 4.78 & 20.79 & 1.40 & 8.39 & 9.09 \\
Qwen-3 14B & -1.39 & 16.27 & 38.49 & 0.54 & 11.65 & 26.02 & -1.44 & 3.29 & 17.16 & -4.20 & 5.59 & 6.29 \\
Qwen-3 32B & -0.68 & 15.04 & 35.45 & -2.84 & 13.66 & 25.77 & -1.86 & 4.12 & 16.49 & -2.05 & 6.16 & 12.33 \\
\bottomrule

\end{tabular}
}

\end{table*}

\subsection{The Illusion of Stability: Aggregated vs. Instance-Level Results}

Table~\ref{tab:consistency_across_datasets} presents the original accuracy ($A_{\text{orig}}$), average paraphrase accuracy ($A_{\text{dist}}$), and majority vote accuracy ($A_{\text{MV}}$) for each model across the four evaluation datasets.\footnote{For the two QA datasets, we employed a three-way rubric. Hence, beyond accuracy ($A$), we also report the proportion of incorrect responses ($B$) and ``not attempted'' responses ($C$). These metrics are calculated for the original questions ($B_{\text{orig}}, C_{\text{orig}}$), the mean across paraphrases ($B_{\text{dist}}, C_{\text{dist}}$), and the majority vote ($B_{\text{MV}}, C_{\text{MV}}$). Detailed results are provided in Tables~\ref{tab:simpleqa-dist-aggregate}--\ref{tab:truthfulqa-dist-majority} in Appendix~\ref{append:more_result_tables}.} We observe that $A_{\text{dist}}$ and $A_{\text{MV}}$ exhibit only modest fluctuations from $A_{\text{orig}}$, typically ranging between 0--5\%, with a maximum deviation of 7.3\%. These observations align with prior findings~\cite{lunardi2025robustness, mirzadeh2024gsm}. Notably, model performance on reasoning-intensive tasks (GSM8K and MATH-500) demonstrates higher variance, characterized by slightly larger accuracy fluctuations compared to the QA benchmarks.

While these aggregated accuracy shifts appear negligible, Table~\ref{tab:hidden_instability} reveals a significant ``hidden instability" at the instance level. As defined in Section~\ref{sec:instance-level-stability}, the mismatch rate quantifies samples where the majority vote across paraphrases diverges from the model’s response to the original question, and $P(\text{ND})$ measures the proportion of sets $Q_i$ where the model produce at least two distinct labels. Intuitively, one might expect $P(\text{ND})$ to be fairly low for a somewhat stable system with small accuracy fluctuations, and the mismatch rate to be roughly equivalent to the absolute gap between $A_{\text{MV}}$ and $A_{\text{orig}}$, assuming the gap is driven solely by samples where the original answer deviates from the majority consensus. 

However, the data in Table~\ref{tab:hidden_instability} shows that $P(\text{ND})$ can reach as high as 47.55\%, and the mismatch rate can be disproportionately high—often several times larger than the accuracy gap. Most strikingly, even in cases where the accuracy gap ($|A_{\text{MV}} - A_{\text{orig}}|$) is zero, we observe mismatch rates as high as 23.12\%. This confirms that aggregated accuracy scores obscure significant instance-level volatility. In essence, the apparent stability of the final score is often a statistical byproduct of offsetting errors, where correct original answers are lost on paraphrases while incorrect original answers are ``corrected'' by the majority vote, rather than a reflection of genuine model reliability.

To quantify this error-canceling phenomenon, we report prediction flip rates  (defined in Section~\ref{sec:instance-level-stability}) in Tables~\ref{tab:simpleqa-flip-nd}--\ref{tab:math-500-flip-nd} of Appendix~\ref{append:more_result_tables}. The results indicate a substantial number of instances where a correct original response is overturned by an incorrect majority vote, as well as instances where an initially incorrect response is rectified through the majority consensus across paraphrases.

Finally, to characterize the extent of response inconsistency, we report the IID-Mismatch rate, Entropy, and Average Mode Share in Tables~\ref{tab:simpleqa-acc-mismatch-combined}--\ref{tab:math-500-anslevel-variability} (Appendix~\ref{append:more_result_tables}). For the mathematical reasoning datasets, these metrics are calculated using two distinct granularities: (1) a binary correct/incorrect classification (the setting used in Tables~\ref{tab:gsm8k-acc-mismatch} and~\ref{tab:math-500-acc-mismatch}), and (2) a value-sensitive approach that treats distinct incorrect numerical outputs as unique categories (the setting used in Tables~\ref{tab:gsm8k-anslevel-variability} and~\ref{tab:math-500-anslevel-variability}). This gives two distinct constructions of the label space $\mathcal{L}$. 
The results reveal that paraphrase-induced instability is widespread but takes different forms across tasks. On \textsc{SimpleQA} and \textsc{TruthfulQA}, mismatch, entropy, and IID-mismatch remain substantial even when $\Delta$Acc is near zero, showing that seemingly small shifts in the aggregated accuracy can mask large answer-level variation. On \textsc{GSM8K} and \textsc{MATH-500}, stronger models and reasoning-enabled variants are more stable under binary correctness, but value-level metrics still expose non-trivial variability in the generated answers. These findings further show that aggregated accuracy alone is not sufficiently explanatory of model reliability under paraphrasing.

% \subsection{Latent Knowledge vs. Consistent Capability}
\subsection{The Reliability Gap: Latent Knowledge vs. Consistent Capability}

\label{sec:latent-knowledge}

The observed sensitivity to paraphrasing raises a fundamental question: \textit{Is a model's reported accuracy a reliable estimate of its knowledge, or does it represent a fragile, stochastic snapshot of performance?} To investigate this, we decouple a model's theoretical ``knowledge ceiling" from its ``reliability floor". We define Latent\textit{ Maximum Capability} ($A_{\text{any}}$) as the proportion of questions where the model provides a correct answer for \textit{at least one} paraphrase, and \textit{Reliable Capability} ($A_{\text{strict}}$) as the proportion where it remains correct across \textit{all} paraphrases. 

\vspace{-5pt}
\begin{figure}[ht]
\centering
% \includegraphics[width=0.48\linewidth]{images/accuracy_paraphrase_range_truthfulqa_v7.pdf}
% \hfill
% \includegraphics[width=0.48\linewidth]{images/orig_vs_conservative_average_bestcase_TruthfulQA_p2.pdf}

\includegraphics[width=\linewidth]{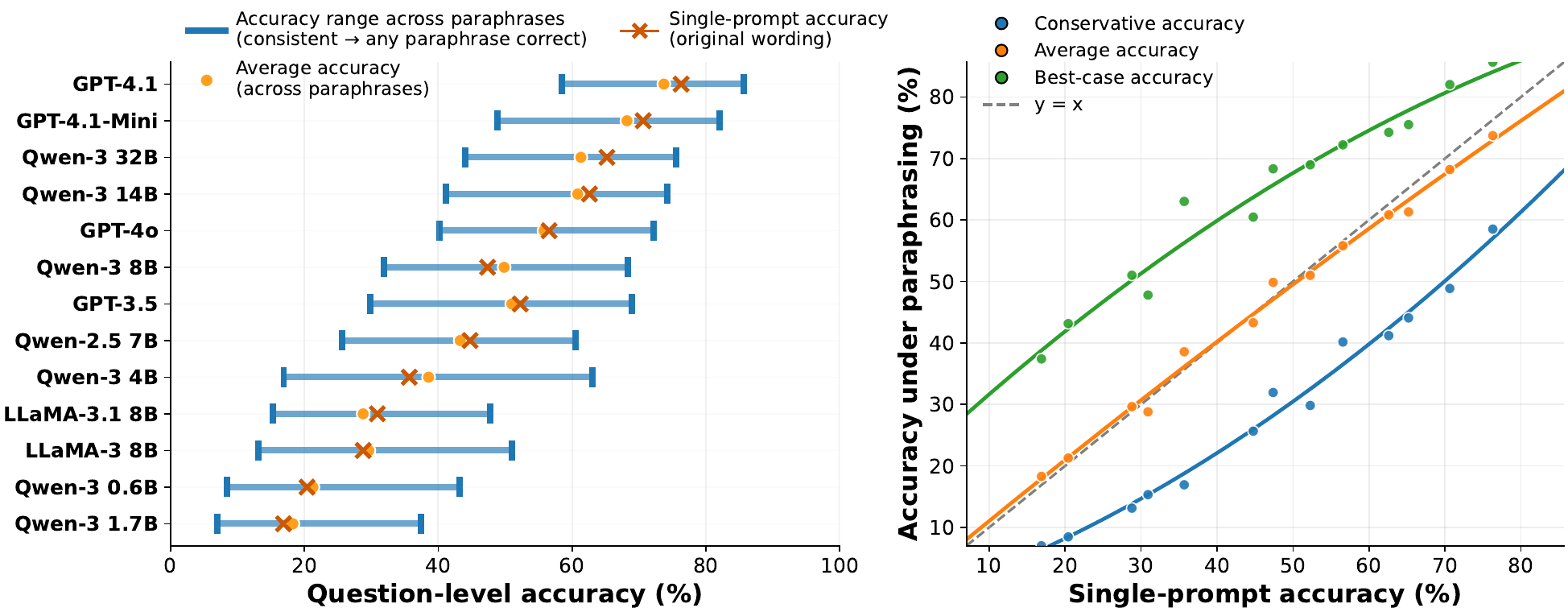}

\caption{
\textbf{Left:} Accuracy varies across paraphrases on \textsc{TruthfulQA}. Bars show range; orange = $A_{\text{dist}}$, red = $A_{\text{orig}}$. 
\textbf{Right:} Relationship between $A_{\text{orig}}$, $A_{\text{dist}}$, $A_{\text{any}}$, and $A_{\text{strict}}$, revealing the reliability gap between latent and reliable capability.
}
\label{fig:truthfulqa_combined}
\end{figure}
% \vspace{-5pt}

% Detailed results are provided in Tables~\ref{tab:simpleqa-anycorrect}--\ref{tab:math-500-anycorrect}, with a visual representation of the performance ranges for TruthfulQA in Figure~\ref{fig:truthfulqa_paraphrase_accuracy_range}. 
Detailed results are provided in Tables~\ref{tab:simpleqa-anycorrect}--\ref{tab:math-500-anycorrect}. Figure~\ref{fig:truthfulqa_combined} visualizes the performance ranges for \textsc{TruthfulQA}, while the corresponding plots for other datasets are provided in Figure~\ref{fig:paraphrase_ranges}.
We observe a substantial gap between maximum capability ($A_{\text{any}}$) and original accuracy ($A_{\text{orig}}$). For example, in some cases, the model possesses the internal knowledge to answer an additional 30\% of questions correctly but fails to retrieve it due to surface-level wording sensitivities. Conversely, $A_{\text{strict}}$ trails significantly behind $A_{\text{orig}}$, indicating that a large portion of a model's perceived success is contingent on specific prompt formulations rather than a robust grasp of the underlying concept. This suggests that ``correct" answers on standard benchmarks often overestimate a model’s reliably accessible knowledge or actual mastery.

% This perspective is further supported by the conditional statistic
% $P(A_{\text{any}} \mid \neg A_{\text{orig}})$, reported in
% Tables~\ref{tab:simpleqa-anycorrect}--\ref{tab:math-500-anycorrect},
% which measures cases where the original wording is answered incorrectly
% but at least one paraphrase elicits the correct answer. Across datasets,
% this probability is often non-trivial, suggesting a practically important
% failure mode: even when a model answers a question incorrectly, there may
% still be a significant chance that it actually has the relevant knowledge
% but fails to express it under the original phrasing.

This latent-capability gap is further reflected in the conditional statistic
$P(A_{\text{any}} \mid \neg A_{\text{orig}})$, reported in
Tables~\ref{tab:simpleqa-anycorrect}--\ref{tab:math-500-anycorrect},
which measures cases where the original wording is answered incorrectly
but at least one paraphrase elicits the correct answer. Across datasets,
this quantity is often substantial, indicating that even when a model
fails on the original prompt, there can still be a significant chance
that the underlying knowledge is present but not reliably retrieved.

To determine whether these performance gaps fundamentally change how we rank and compare different architectures, we analyze the scaling behavior of the three capability measures.
% Figure~\ref{fig:truthfulqa_metric_correlations} consolidates these relationships—Reliable ($A_{\text{strict}}$), Average ($A_{\text{dist}}$), and Latent ($A_{\text{any}}$)—into a single view for the \textsc{TruthfulQA} benchmark.
In Figure~\ref{fig:truthfulqa_combined}, we illustrate the relationships between Reliable ($A_{\text{strict}}$), Average ($A_{\text{dist}}$), and Latent ($A_{\text{any}}$) capabilities on the \textsc{TruthfulQA} benchmark by fitting a 2nd order polynomial function to the data points. We choose 2nd order functions as the hypothesis set to strike a balance between smoothness and expressiveness. While we illustrate the trends on \textsc{TruthfulQA} here for clarity, the same qualitative patterns hold across all remaining evaluated datasets (as shown in Figure~\ref{fig:metric_correlations}).

First, observing the lower bound, conservative accuracy ($A_{\text{strict}}$) generally scales with $A_{\text{orig}}$, confirming that stronger models tend to be more robust. However, $A_{\text{strict}}$ consistently lies below the identity line, indicating a persistent gap relative to single-prompt accuracy. The trend also exhibits a convex upward shape, with the gap widening for intermediate-performing models. Although stronger models tend to reduce this gap, it remains visibly present even at high accuracy levels, suggesting that perfect robustness to paraphrasing is rarely achieved.

In contrast, the average paraphrase accuracy ($A_{\text{dist}}$) remains tightly correlated with $A_{\text{orig}}$ across the spectrum. The data points cluster closely around the identity line with minimal deviation—typically slightly below it—indicating that $A_{\text{orig}}$ remains a statistically sound proxy for comparing relative aggregate capabilities, despite occasional localized ranking instabilities.

% Finally, the upper bound ($A_{\text{any}}$) reveals latent knowledge. The gap between $A_{\text{orig}}$ and $A_{\text{any}}$ corresponds to questions for which at least one paraphrase elicits the correct answer. 
% This gap suggests that models may possess knowledge that is not consistently retrieved across different phrasings of the same question, with stronger models in some cases appearing to retrieve this knowledge more reliably.
Finally, the upper bound ($A_{\text{any}}$) reveals latent knowledge. The gap between $A_{\text{orig}}$ and $A_{\text{any}}$ corresponds to questions for which at least one paraphrase elicits the correct answer. Across datasets, this gap tends to widen for intermediate-performing models and then narrow again for stronger ones, suggesting that models often possess knowledge that is not consistently retrieved across different phrasings of the same question, with stronger models appearing to access this knowledge more reliably.

Taken together, these results reveal a consistent ordering across models,
$A_{\text{strict}} < A_{\text{orig}} \approx A_{\text{dist}} < A_{\text{any}}$,
indicating that single-prompt accuracy lies within a broader capability band bounded by reliable consistency and latent knowledge.

\subsection{The Impact of Model Scale and Reasoning on Consistency}
% Having established the prevalence of inconsistency, we now isolate specific factors that modulate this behavior, focusing on model architecture and the linguistic properties of the paraphrases themselves.

% \subsubsection{The Impact of Model Scale and Reasoning}

% \begin{figure}[h]
%     \centering
%     \includegraphics[width=0.9\linewidth]{images/fig_qwen3_reasoning_vs_scale_dual_axis_v3.pdf}
%     \caption{The effect of model scale and reasoning mode on performance, averaged across 4 datasets. Accuracy (solid lines) generally improves with scale, while the mismatch rate (dashed lines) declines, indicating better consistency. CoT reasoning further enhances stability across all scales.}
%     \label{fig:factor-scale-cot}
%     \vspace{-2pt}
% \end{figure}

\begin{wrapfigure}{r}{0.52\columnwidth}
\vspace{-10pt}
\centering
\includegraphics[width=\linewidth]{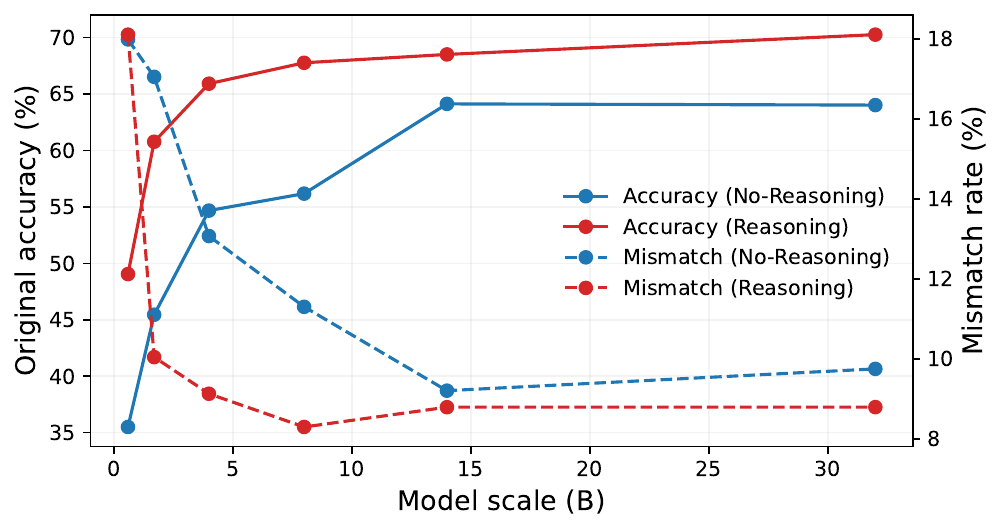}
\vspace{-6pt}
\caption{\small Effect of scale and CoT on accuracy (solid) and mismatch (dashed).}
    % \caption{The effect of model scale and reasoning mode on performance, averaged across 4 datasets. 
    % % Accuracy (solid lines) generally improves with scale, while the mismatch rate (dashed lines) declines, indicating better consistency. CoT reasoning further enhances stability across all scales.
    % }
\vspace{-10pt}
% \vspace{-5pt}
\label{fig:factor-scale-cot}
\end{wrapfigure}

To decouple the effects of parameter count and whether or not the model operates in reasoning mode, we conduct controlled experiments using six variants of the \texttt{Qwen 3} model family~\cite{yang2025qwen3technicalreport} (spanning 0.6B, 1.7B, 4B, 8B, 14B, and 32B). Each model is evaluated both with and without long Chain-of-Thought (CoT) reasoning. 

Figure~\ref{fig:factor-scale-cot} illustrates the average accuracy and mismatch rates across all four datasets (per-dataset breakdowns are available in Figure~\ref{fig:per-model-qwen3} of Appendix~\ref{append:more_result_tables}). Two clear trends emerge. First, consistency is positively correlated with model scale: larger models not only achieve higher original accuracy ($A_{\text{orig}}$) but also demonstrate greater stability across paraphrases. Second, enabling long CoT reasoning yields improvements in both accuracy and consistency compared to standard prompting.

% \begin{figure*}
%     \centering
%     \includegraphics[width=\linewidth]{images/correct_vs_paratone.pdf}
%     \caption{Comparison of Correctness Rate (\%) across original, plain, and toned paraphrase styles. While stylized inputs have a negligible effect on factual QA, they noticeably degrade performance on mathematical reasoning tasks.}
%     \label{fig:para_type_vs_correct}
% \end{figure*}

% \begin{table*}[ht]
% \centering
% \caption{Impact of Self-Paraphrasing on answer distribution. Conditioning on self-generated paraphrases generally shifts the distribution from ``Not Attempted'' or ``Incorrect'' toward ``Correct,'' bridging the gap between displayed and latent capability.}
% \label{tab:self_paraphrasing_results}
% \small
% \renewcommand{\arraystretch}{1.2}
% \begin{tabular*}{0.85\linewidth}{@{\extracolsep{\fill}}l ccc ccc}
% \toprule
% \multirow{2}{*}{\textbf{Model}} & \multicolumn{3}{c}{\textbf{Original Prompting} (\%)} & \multicolumn{3}{c}{\textbf{Self-Paraphrasing} (\%)} \\
% \cmidrule(lr){2-4} \cmidrule(lr){5-7}
% & \textbf{Corr.} & \textbf{Inc.} & \textbf{N/A} & \textbf{Corr.} & \textbf{Inc.} & \textbf{N/A} \\
% \midrule
% gpt-4.1 & 45.1 & 37.0 & 17.9 & 40.2 & 44.6 & 15.1 \\
% gpt-4o & 23.6 & 25.4 & 51.0 & 27.7 & 19.2 & 53.1 \\
% gpt-4.1-mini & 24.1 & 60.6 & 15.3 & \textbf{32.6} & 27.2 & 40.2 \\
% gpt-3.5-turbo & 9.3 & 15.7 & 75.0 & \textbf{13.6} & 41.4 & 45.0 \\
% Qwen2.5-7B-Instruct & 6.8 & 2.6 & 90.6 & \textbf{16.3} & 5.1 & 78.6 \\
% Llama-3.1-8B-Instruct & 16.1 & 14.5 & 69.4 & \textbf{17.8} & 5.3 & 76.9 \\
% \bottomrule
% \end{tabular*}
% % \vspace{10pt}
% \end{table*}

However, reasoning is not a panacea. As detailed in Tables~\ref{tab:simpleqa-flip-nd}--\ref{tab:math-500-flip-nd} and Tables~\ref{tab:simpleqa-acc-mismatch-combined}--\ref{tab:math-500-anslevel-variability}, even models operating in reasoning mode exhibit considerable instability. High prediction flip rates and mismatch rates persist, suggesting that while CoT helps ground the model, it does not fully immunize it against surface-level perturbations.

\subsection{Bridging the Capability Gap via Self-Paraphrasing}

The significant divergence between $A_{\text{orig}}$ and $A_{\text{any}}$ observed in Section~\ref{sec:latent-knowledge} implies that models often possess the requisite knowledge to answer a query but fail to retrieve it due to surface-level phrasing mismatches. Motivated by this ``reliability gap'', we explore a \textbf{Self-Paraphrasing} strategy. We hypothesize that by generating multiple semantic variations of a query, the model can construct a richer context that effectively acts as a test-time alignment step, mapping the user's intent to a latent space where knowledge retrieval is more robust.

\paragraph{Methodology}
We implement a self-paraphrasing mechanism in which the model first generates multiple semantically equivalent paraphrases of the input question, then produces a final answer conditioned on these variations, which provide auxiliary context for disambiguation.

\begin{wraptable}{r}{0.55\columnwidth}
\vspace{-10pt}
\centering
\scriptsize
\setlength{\tabcolsep}{3pt}
\renewcommand{\arraystretch}{1.05}

\begin{tabular}{l ccc ccc}
\toprule
\multirow{2}{*}{\textbf{Model}} & \multicolumn{3}{c}{\textbf{Original Prompting} (\%)} & \multicolumn{3}{c}{\textbf{Self-Paraphrasing} (\%)} \\
\cmidrule(lr){2-4} \cmidrule(lr){5-7}
& \textbf{Corr.} & \textbf{Inc.} & \textbf{N/A} & \textbf{Corr.} & \textbf{Inc.} & \textbf{N/A} \\
\midrule
GPT-3.5-turbo & 9.2 & 39.5 & 51.3 & \textbf{12.3} & 68.2 & 19.4 \\
GPT-4.1 & 39.4 & 50.0 & 10.6 & \textbf{41.8} & 50.2 & 8.0 \\
GPT-4o & \textbf{20.9} & 25.5 & 53.6 & 18.4 & 18.9 & 62.7 \\
LLaMA-3 8B & 4.0 & 39.2 & 56.7 & \textbf{5.8} & 54.0 & 40.2 \\
LLaMA-3.1 8B & 1.0 & 3.7 & 95.3 & \textbf{2.0} & 10.8 & 87.2 \\
Qwen2.5 7B & 0.5 & 2.4 & 97.2 & \textbf{1.1} & 21.8 & 77.1 \\
Qwen3 8B & 2.7 & 48.3 & 49.0 & \textbf{4.0} & 76.3 & 19.7 \\
\bottomrule
\end{tabular}

% \vspace{-6pt}
\caption{\small Impact of self-paraphrasing on answer distribution.}
\vspace{-10pt}
\label{tab:self_paraphrasing_results}
\end{wraptable}

We evaluate two implementations.
\textbf{Two-Stage Prompting:} Paraphrases and the final answer are generated in separate inference steps.
\textbf{One-Stage Prompting:} The model generates paraphrases and the final answer within a single context window, conditioning the answer on the preceding self-generated paraphrases.
Preliminary experiments show no meaningful performance difference, so we report results using the one-stage variant (Appendix~\ref{append:self-paraphrase-prompt}).

\paragraph{Evaluation \& Results}
We evaluate Self-Paraphrasing on a subset of \textsc{SimpleQA}, using a three-way labeling (Correct, Incorrect, Not Attempted).

% \input{tables/self_paraphrasing_results-tab}

% Table~\ref{tab:self_paraphrasing_results} presents the comparative results. Despite the limited scale, conditioning on self-generated paraphrases is associated with improved performance for many models. This shows self-paraphrasing could act as a lightweight, inference-time ``denoising'' mechanism that helps bridge the capability gaps of models.
% Table~\ref{tab:self_paraphrasing_results} shows that conditioning on self-generated paraphrases improves performance for many models. This suggests self-paraphrasing acts as a lightweight, inference-time ``denoising'' mechanism that helps bridge capability gaps.
Table~\ref{tab:self_paraphrasing_results} shows that conditioning on self-generated paraphrases improves performance for most models (6/7). Despite the limited scale, this indicates that self-paraphrasing can act as a lightweight, inference-time ``denoising'' mechanism that helps bridge capability gaps.

\section{Discussion}

% We discuss implications for LLM evaluation and robustness, with additional discussion on dataset contamination and deployment deferred to Appendix~\ref{app:discussion}.
In this section, we discuss implications of our findings on LLM evaluation and robustness. Additional discussions on the interpretation of benchmark contamination, possible mechanisms and deployment implications are provided in Appendix~\ref{app:discussion}. We discuss limitations, including finite paraphrase coverage, in Appendix~\ref{app:limitations}.

\subsection{Implications for Benchmark Design and Evaluation}

Our findings suggest that evaluating language models with a single prompt per example provides an incomplete view of reliability. Standard benchmarks implicitly treat accuracy on a fixed wording as a reliable measure of performance. However, our results show that predictions and correctness are better understood as distributions over semantically equivalent formulations, and that aggregate accuracy can mask instability. This motivates evaluation protocols that account for semantic variation, for example by reporting metrics such as reliable capability ($A_\text{strict}$), expected paraphrase accuracy ($A_\text{dist}$), and latent capability ($A_\text{any}$), or by evaluating models across paraphrased inputs.

\subsection{Consistency as a Robustness Property}

Paraphrase sensitivity can be viewed through the lens of robustness. In the view of adversarial robustness, models are expected to produce invariant outputs under transformations that preserve the semantics of the input~\cite{szegedy2013intriguing,goodfellow2014explaining,carlini2017towards}. In language models, paraphrasing constitutes such a transformation: for a meaning-preserving mapping $T$, a reliable system should ideally satisfy $f(x)=f(T(x))$. Our results show that this invariance often fails even under \textbf{natural}, \textbf{non-adversarial} rephrasings. 
% Importantly, all evaluations use deterministic decoding, isolating instability induced by semantic rephrasings rather than stochastic sampling effects. 
Evaluating models across paraphrased inputs therefore provides a practical test of semantic robustness under realistic language variation.

% \subsection{Paraphrase Sensitivity and Benchmark Contamination}

% Our findings also affect how benchmark contamination is interpreted. It's often argued that existing popular benchmarks suffer from test set contamination, and that current models have effectively ``seen'' almost all public test samples during training~\cite{golchin2023time, golchin2024datacontaminationquiztool, dekoninck2024constat, singh2024evaluationdatacontaminationllms,cheng-etal-2025-dyepack}. Therefore, models will often only perform well when tested with the given wordings in a benchmark test set, and will suffer a performance drop when the wordings are changed. However, we observed that while models may sometimes answer the original wording correctly while failing on semantically equivalent paraphrases, in many other cases models may succeed on paraphrases after failing on the original question. This \textbf{bidirectional} behavior suggests that performance on benchmark items cannot be explained solely by memorization or contamination. Instead, sensitivity to surface form and inconsistent knowledge retrieval can produce outcomes that resemble contamination effects. As a result, apparent contamination signals may partly reflect paraphrase sensitivity rather than purely memorized benchmark data.
\section{Conclusion}
This work exposes the ``illusion of stability" in LLM evaluation: while aggregated accuracy remains modestly stable, instance-level predictions fluctuate significantly under paraphrasing. We identify a pervasive ``reliability gap," where models possess latent knowledge ($A_\text{any}$) that they fail to access reliably ($A_\text{strict}$). However, we demonstrate that simple test-time strategies like self-paraphrasing can bridge this gap by triangulating internal knowledge. These findings underscore the need to move beyond single-prompt accuracy toward fine-grained, consistency-aware benchmarks that better reflect real-world reliability.

\begin{ack}
This work was supported in part by NSF CAREER Award 1942230, the ONR PECASE Award N00014-25-1-2378, ARO Early Career Program Award 310902-00001, Army Grant W911NF-21-2-0076, NSF Award CCF-2212458, NSF Award 2229885 (NSF Institute for Trustworthy AI in Law and Society, TRAILS), MURI Grant 14262683, DARPA AIQ Grant HR00112590066, and a Meta Research Award 314593-00001. We also thank Hossein Faghih for valuable discussions and significant assistance throughout the development of this project.
\end{ack}

\bibliographystyle{abbrvnat}
\bibliography{_ref}

@article{goodfellow2014explaining,
  title={Explaining and harnessing adversarial examples},
  author={Goodfellow, Ian J and Shlens, Jonathon and Szegedy, Christian},
  journal={arXiv preprint arXiv:1412.6572},
  year={2014}
}

@inproceedings{carlini2017towards,
  title={Towards evaluating the robustness of neural networks},
  author={Carlini, Nicholas and Wagner, David},
  booktitle={2017 ieee symposium on security and privacy (sp)},
  pages={39--57},
  year={2017},
  organization={Ieee}
}

@article{dekoninck2024constat,
  title={Constat: Performance-based contamination detection in large language models},
  author={Dekoninck, Jasper and M{\"u}ller, Mark and Vechev, Martin},
  journal={Advances in Neural Information Processing Systems},
  volume={37},
  pages={92420--92464},
  year={2024}
}

@inproceedings{cheng-etal-2025-dyepack,
    title = "{D}ye{P}ack: Provably Flagging Test Set Contamination in {LLM}s Using Backdoors",
    author = "Cheng, Yize  and
      Wang, Wenxiao  and
      Moayeri, Mazda  and
      Feizi, Soheil",
    editor = "Christodoulopoulos, Christos  and
      Chakraborty, Tanmoy  and
      Rose, Carolyn  and
      Peng, Violet",
    booktitle = "Proceedings of the 2025 Conference on Empirical Methods in Natural Language Processing",
    month = nov,
    year = "2025",
    address = "Suzhou, China",
    publisher = "Association for Computational Linguistics",
    url = "https://aclanthology.org/2025.emnlp-main.776/",
    doi = "10.18653/v1/2025.emnlp-main.776",
    pages = "15356--15373",
    ISBN = "979-8-89176-332-6"
}

@misc{singh2024evaluationdatacontaminationllms,
      title={Evaluation data contamination in LLMs: how do we measure it and (when) does it matter?}, 
      author={Aaditya K. Singh and Muhammed Yusuf Kocyigit and Andrew Poulton and David Esiobu and Maria Lomeli and Gergely Szilvasy and Dieuwke Hupkes},
      year={2024},
      eprint={2411.03923},
      archivePrefix={arXiv},
      primaryClass={cs.CL},
      url={https://arxiv.org/abs/2411.03923}, 
}

@article{golchin2023time,
  title={Time travel in llms: Tracing data contamination in large language models},
  author={Golchin, Shahriar and Surdeanu, Mihai},
  journal={arXiv preprint arXiv:2308.08493},
  year={2023}
}

@misc{golchin2024datacontaminationquiztool,
      title={Data Contamination Quiz: A Tool to Detect and Estimate Contamination in Large Language Models}, 
      author={Shahriar Golchin and Mihai Surdeanu},
      year={2024},
      eprint={2311.06233},
      archivePrefix={arXiv},
      primaryClass={cs.CL},
      url={https://arxiv.org/abs/2311.06233}, 
}

@article{szegedy2013intriguing,
  title={Intriguing properties of neural networks},
  author={Szegedy, Christian and Zaremba, Wojciech and Sutskever, Ilya and Bruna, Joan and Erhan, Dumitru and Goodfellow, Ian and Fergus, Rob},
  journal={arXiv preprint arXiv:1312.6199},
  year={2013}
}

@misc{jiang2025artificialhivemindopenendedhomogeneity,
      title={Artificial Hivemind: The Open-Ended Homogeneity of Language Models (and Beyond)}, 
      author={Liwei Jiang and Yuanjun Chai and Margaret Li and Mickel Liu and Raymond Fok and Nouha Dziri and Yulia Tsvetkov and Maarten Sap and Alon Albalak and Yejin Choi},
      year={2025},
      eprint={2510.22954},
      archivePrefix={arXiv},
      primaryClass={cs.CL},
      url={https://arxiv.org/abs/2510.22954}, 
}

@misc{wei2024measuringshortformfactualitylarge,
      title={Measuring short-form factuality in large language models}, 
      author={Jason Wei and Nguyen Karina and Hyung Won Chung and Yunxin Joy Jiao and Spencer Papay and Amelia Glaese and John Schulman and William Fedus},
      year={2024},
      eprint={2411.04368},
      archivePrefix={arXiv},
      primaryClass={cs.CL},
      url={https://arxiv.org/abs/2411.04368}, 
}

@article{grattafiori2024llama3herdmodels,
  title={The llama 3 herd of models},
  author={Grattafiori, Aaron and Dubey, Abhimanyu and Jauhri, Abhinav and Pandey, Abhinav and Kadian, Abhishek and Al-Dahle, Ahmad and Letman, Aiesha and Mathur, Akhil and Schelten, Alan and Vaughan, Alex and others},
  journal={arXiv preprint arXiv:2407.21783},
  year={2024}
}

@misc{hendrycks2021measuringmathematicalproblemsolving,
      title={Measuring Mathematical Problem Solving With the MATH Dataset}, 
      author={Dan Hendrycks and Collin Burns and Saurav Kadavath and Akul Arora and Steven Basart and Eric Tang and Dawn Song and Jacob Steinhardt},
      year={2021},
      eprint={2103.03874},
      archivePrefix={arXiv},
      primaryClass={cs.LG},
      url={https://arxiv.org/abs/2103.03874}, 
}

@misc{cobbe2021trainingverifierssolvemath,
      title={Training Verifiers to Solve Math Word Problems}, 
      author={Karl Cobbe and Vineet Kosaraju and Mohammad Bavarian and Mark Chen and Heewoo Jun and Lukasz Kaiser and Matthias Plappert and Jerry Tworek and Jacob Hilton and Reiichiro Nakano and Christopher Hesse and John Schulman},
      year={2021},
      eprint={2110.14168},
      archivePrefix={arXiv},
      primaryClass={cs.LG},
      url={https://arxiv.org/abs/2110.14168}, 
}

@misc{lin2022truthfulqameasuringmodelsmimic,
      title={TruthfulQA: Measuring How Models Mimic Human Falsehoods}, 
      author={Stephanie Lin and Jacob Hilton and Owain Evans},
      year={2022},
      eprint={2109.07958},
      archivePrefix={arXiv},
      primaryClass={cs.CL},
      url={https://arxiv.org/abs/2109.07958}, 
}

@misc{qwen2025qwen25technicalreport,
      title={Qwen2.5 Technical Report}, 
      author={Qwen and : and An Yang and Baosong Yang and Beichen Zhang and Binyuan Hui and Bo Zheng and Bowen Yu and Chengyuan Li and Dayiheng Liu and Fei Huang and Haoran Wei and Huan Lin and Jian Yang and Jianhong Tu and Jianwei Zhang and Jianxin Yang and Jiaxi Yang and Jingren Zhou and Junyang Lin and Kai Dang and Keming Lu and Keqin Bao and Kexin Yang and Le Yu and Mei Li and Mingfeng Xue and Pei Zhang and Qin Zhu and Rui Men and Runji Lin and Tianhao Li and Tianyi Tang and Tingyu Xia and Xingzhang Ren and Xuancheng Ren and Yang Fan and Yang Su and Yichang Zhang and Yu Wan and Yuqiong Liu and Zeyu Cui and Zhenru Zhang and Zihan Qiu},
      year={2025},
      eprint={2412.15115},
      archivePrefix={arXiv},
      primaryClass={cs.CL},
      url={https://arxiv.org/abs/2412.15115}, 
}

@misc{yang2025qwen3technicalreport,
      title={Qwen3 Technical Report}, 
      author={An Yang and Anfeng Li and Baosong Yang and Beichen Zhang and Binyuan Hui and Bo Zheng and Bowen Yu and Chang Gao and Chengen Huang and Chenxu Lv and Chujie Zheng and Dayiheng Liu and Fan Zhou and Fei Huang and Feng Hu and Hao Ge and Haoran Wei and Huan Lin and Jialong Tang and Jian Yang and Jianhong Tu and Jianwei Zhang and Jianxin Yang and Jiaxi Yang and Jing Zhou and Jingren Zhou and Junyang Lin and Kai Dang and Keqin Bao and Kexin Yang and Le Yu and Lianghao Deng and Mei Li and Mingfeng Xue and Mingze Li and Pei Zhang and Peng Wang and Qin Zhu and Rui Men and Ruize Gao and Shixuan Liu and Shuang Luo and Tianhao Li and Tianyi Tang and Wenbiao Yin and Xingzhang Ren and Xinyu Wang and Xinyu Zhang and Xuancheng Ren and Yang Fan and Yang Su and Yichang Zhang and Yinger Zhang and Yu Wan and Yuqiong Liu and Zekun Wang and Zeyu Cui and Zhenru Zhang and Zhipeng Zhou and Zihan Qiu},
      year={2025},
      eprint={2505.09388},
      archivePrefix={arXiv},
      primaryClass={cs.CL},
      url={https://arxiv.org/abs/2505.09388}, 
}

@article{openai2024gpt4technicalreport,
  title={Gpt-4 technical report},
  author={Achiam, Josh and Adler, Steven and Agarwal, Sandhini and Ahmad, Lama and Akkaya, Ilge and Aleman, Florencia Leoni and Almeida, Diogo and Altenschmidt, Janko and Altman, Sam and Anadkat, Shyamal and others},
  journal={arXiv preprint arXiv:2303.08774},
  year={2023}
}

@misc{brown2020languagemodelsfewshotlearners,
      title={Language Models are Few-Shot Learners}, 
      author={Tom B. Brown and Benjamin Mann and Nick Ryder and Melanie Subbiah and Jared Kaplan and Prafulla Dhariwal and Arvind Neelakantan and Pranav Shyam and Girish Sastry and Amanda Askell and Sandhini Agarwal and Ariel Herbert-Voss and Gretchen Krueger and Tom Henighan and Rewon Child and Aditya Ramesh and Daniel M. Ziegler and Jeffrey Wu and Clemens Winter and Christopher Hesse and Mark Chen and Eric Sigler and Mateusz Litwin and Scott Gray and Benjamin Chess and Jack Clark and Christopher Berner and Sam McCandlish and Alec Radford and Ilya Sutskever and Dario Amodei},
      year={2020},
      eprint={2005.14165},
      archivePrefix={arXiv},
      primaryClass={cs.CL},
      url={https://arxiv.org/abs/2005.14165}, 
}

@misc{chowdhery2022palmscalinglanguagemodeling,
      title={PaLM: Scaling Language Modeling with Pathways}, 
      author={Aakanksha Chowdhery and Sharan Narang and Jacob Devlin and Maarten Bosma and Gaurav Mishra and Adam Roberts and Paul Barham and Hyung Won Chung and Charles Sutton and Sebastian Gehrmann and Parker Schuh and Kensen Shi and Sasha Tsvyashchenko and Joshua Maynez and Abhishek Rao and Parker Barnes and Yi Tay and Noam Shazeer and Vinodkumar Prabhakaran and Emily Reif and Nan Du and Ben Hutchinson and Reiner Pope and James Bradbury and Jacob Austin and Michael Isard and Guy Gur-Ari and Pengcheng Yin and Toju Duke and Anselm Levskaya and Sanjay Ghemawat and Sunipa Dev and Henryk Michalewski and Xavier Garcia and Vedant Misra and Kevin Robinson and Liam Fedus and Denny Zhou and Daphne Ippolito and David Luan and Hyeontaek Lim and Barret Zoph and Alexander Spiridonov and Ryan Sepassi and David Dohan and Shivani Agrawal and Mark Omernick and Andrew M. Dai and Thanumalayan Sankaranarayana Pillai and Marie Pellat and Aitor Lewkowycz and Erica Moreira and Rewon Child and Oleksandr Polozov and Katherine Lee and Zongwei Zhou and Xuezhi Wang and Brennan Saeta and Mark Diaz and Orhan Firat and Michele Catasta and Jason Wei and Kathy Meier-Hellstern and Douglas Eck and Jeff Dean and Slav Petrov and Noah Fiedel},
      year={2022},
      eprint={2204.02311},
      archivePrefix={arXiv},
      primaryClass={cs.CL},
      url={https://arxiv.org/abs/2204.02311}, 
}

@misc{devlin2019bertpretrainingdeepbidirectional,
      title={BERT: Pre-training of Deep Bidirectional Transformers for Language Understanding}, 
      author={Jacob Devlin and Ming-Wei Chang and Kenton Lee and Kristina Toutanova},
      year={2019},
      eprint={1810.04805},
      archivePrefix={arXiv},
      primaryClass={cs.CL},
      url={https://arxiv.org/abs/1810.04805}, 
}

@article{Guo_2025,
  title={Deepseek-r1: Incentivizing reasoning capability in llms via reinforcement learning},
  author={Guo, Daya and Yang, Dejian and Zhang, Haowei and Song, Junxiao and Wang, Peiyi and Zhu, Qihao and Xu, Runxin and Zhang, Ruoyu and Ma, Shirong and Bi, Xiao and others},
  journal={arXiv preprint arXiv:2501.12948},
  year={2025}
}

@misc{wei2023chainofthoughtpromptingelicitsreasoning,
      title={Chain-of-Thought Prompting Elicits Reasoning in Large Language Models}, 
      author={Jason Wei and Xuezhi Wang and Dale Schuurmans and Maarten Bosma and Brian Ichter and Fei Xia and Ed Chi and Quoc Le and Denny Zhou},
      year={2023},
      eprint={2201.11903},
      archivePrefix={arXiv},
      primaryClass={cs.CL},
      url={https://arxiv.org/abs/2201.11903}, 
}

@misc{khashabi2020unifiedqacrossingformatboundaries,
      title={UnifiedQA: Crossing Format Boundaries With a Single QA System}, 
      author={Daniel Khashabi and Sewon Min and Tushar Khot and Ashish Sabharwal and Oyvind Tafjord and Peter Clark and Hannaneh Hajishirzi},
      year={2020},
      eprint={2005.00700},
      archivePrefix={arXiv},
      primaryClass={cs.CL},
      url={https://arxiv.org/abs/2005.00700}, 
}

@misc{raffel2023exploringlimitstransferlearning,
      title={Exploring the Limits of Transfer Learning with a Unified Text-to-Text Transformer}, 
      author={Colin Raffel and Noam Shazeer and Adam Roberts and Katherine Lee and Sharan Narang and Michael Matena and Yanqi Zhou and Wei Li and Peter J. Liu},
      year={2023},
      eprint={1910.10683},
      archivePrefix={arXiv},
      primaryClass={cs.LG},
      url={https://arxiv.org/abs/1910.10683}, 
}

@inproceedings{pezeshkpour2024large,
  title={Large language models sensitivity to the order of options in multiple-choice questions},
  author={Pezeshkpour, Pouya and Hruschka, Estevam},
  booktitle={Findings of the Association for Computational Linguistics: NAACL 2024},
  pages={2006--2017},
  year={2024}
}

@inproceedings{li-etal-2024-multiple,
    title = "Can Multiple-choice Questions Really Be Useful in Detecting the Abilities of {LLM}s?",
    author = "Li, Wangyue  and
      Li, Liangzhi  and
      Xiang, Tong  and
      Liu, Xiao  and
      Deng, Wei  and
      Garcia, Noa",
    editor = "Calzolari, Nicoletta  and
      Kan, Min-Yen  and
      Hoste, Veronique  and
      Lenci, Alessandro  and
      Sakti, Sakriani  and
      Xue, Nianwen",
    booktitle = "Proceedings of the 2024 Joint International Conference on Computational Linguistics, Language Resources and Evaluation (LREC-COLING 2024)",
    month = may,
    year = "2024",
    address = "Torino, Italia",
    publisher = "ELRA and ICCL",
    url = "https://aclanthology.org/2024.lrec-main.251/",
    pages = "2819--2834"
}

@article{ribeiro2020beyond,
  title={Beyond accuracy: Behavioral testing of NLP models with CheckList},
  author={Ribeiro, Marco Tulio and Wu, Tongshuang and Guestrin, Carlos and Singh, Sameer},
  journal={arXiv preprint arXiv:2005.04118},
  year={2020}
}

@inproceedings{gan2019improving,
  title={Improving the robustness of question answering systems to question paraphrasing},
  author={Gan, Wee Chung and Ng, Hwee Tou},
  booktitle={Proceedings of the 57th annual meeting of the association for computational linguistics},
  pages={6065--6075},
  year={2019}
}

@article{elazar2021measuring,
  title={Measuring and improving consistency in pretrained language models},
  author={Elazar, Yanai and Kassner, Nora and Ravfogel, Shauli and Ravichander, Abhilasha and Hovy, Eduard and Sch{\"u}tze, Hinrich and Goldberg, Yoav},
  journal={Transactions of the Association for Computational Linguistics},
  volume={9},
  pages={1012--1031},
  year={2021},
  publisher={MIT Press One Rogers Street, Cambridge, MA 02142-1209, USA journals-info~…}
}

@article{lunardi2025robustness,
  title={On robustness and reliability of benchmark-based evaluation of llms},
  author={Lunardi, Riccardo and Della Mea, Vincenzo and Mizzaro, Stefano and Roitero, Kevin},
  journal={arXiv preprint arXiv:2509.04013},
  year={2025}
}

@inproceedings{alzahrani2024benchmarks,
  title={When benchmarks are targets: Revealing the sensitivity of large language model leaderboards},
  author={Alzahrani, Norah and Alyahya, Hisham and Alnumay, Yazeed and Alrashed, Sultan and Alsubaie, Shaykhah and Almushayqih, Yousef and Mirza, Faisal and Alotaibi, Nouf and Al-Twairesh, Nora and Alowisheq, Areeb and others},
  booktitle={Proceedings of the 62nd Annual Meeting of the Association for Computational Linguistics (Volume 1: Long Papers)},
  pages={13787--13805},
  year={2024}
}

@article{polo2024efficient,
  title={Efficient multi-prompt evaluation of LLMs},
  author={Polo, Felipe Maia and Xu, Ronald and Weber, Lucas and Silva, M{\'\i}rian and Bhardwaj, Onkar and Choshen, Leshem and de Oliveira, Allysson Flavio Melo and Sun, Yuekai and Yurochkin, Mikhail},
  journal={arXiv preprint arXiv:2405.17202},
  year={2024}
}

@article{mizrahi2024state,
  title={State of what art? a call for multi-prompt llm evaluation},
  author={Mizrahi, Moran and Kaplan, Guy and Malkin, Dan and Dror, Rotem and Shahaf, Dafna and Stanovsky, Gabriel},
  journal={Transactions of the Association for Computational Linguistics},
  volume={12},
  pages={933--949},
  year={2024},
  publisher={MIT Press 255 Main Street, 9th Floor, Cambridge, Massachusetts 02142, USA~…}
}

@article{sclar2023quantifying,
  title={Quantifying Language Models' Sensitivity to Spurious Features in Prompt Design or: How I learned to start worrying about prompt formatting},
  author={Sclar, Melanie and Choi, Yejin and Tsvetkov, Yulia and Suhr, Alane},
  journal={arXiv preprint arXiv:2310.11324},
  year={2023}
}

@article{choi2025roparq,
  title={RoParQ: Paraphrase-Aware Alignment of Large Language Models Towards Robustness to Paraphrased Questions},
  author={Choi, Minjoon},
  journal={arXiv preprint arXiv:2511.21568},
  year={2025}
}

@article{zhou2024paraphrase,
  title={Paraphrase and solve: Exploring and exploiting the impact of surface form on mathematical reasoning in large language models},
  author={Zhou, Yue and Zhu, Yada and Antognini, Diego and Kim, Yoon and Zhang, Yang},
  journal={arXiv preprint arXiv:2404.11500},
  year={2024}
}

@inproceedings{han2025analysis,
  title={An Analysis of the Impact of Problem Paraphrasing on LLM-Based Mathematical Problem Solving},
  author={Han, Yerim and Seo, Hyein and Namgoong, Hyuk and Jung, Sangkeun},
  booktitle={Proceedings of the 14th International Joint Conference on Natural Language Processing and the 4th Conference of the Asia-Pacific Chapter of the Association for Computational Linguistics},
  pages={383--395},
  year={2025}
}

@inproceedings{meier2025towards,
  title={Towards human understanding of paraphrase types in large language models},
  author={Meier, Dominik and Wahle, Jan Philip and Ruas, Terry Lima and Gipp, Bela},
  booktitle={Proceedings of the 31st International Conference on Computational Linguistics},
  pages={6298--6316},
  year={2025}
}

@article{mirzadeh2024gsm,
  title={Gsm-symbolic: Understanding the limitations of mathematical reasoning in large language models},
  author={Mirzadeh, Iman and Alizadeh, Keivan and Shahrokhi, Hooman and Tuzel, Oncel and Bengio, Samy and Farajtabar, Mehrdad},
  journal={arXiv preprint arXiv:2410.05229},
  year={2024}
}

@article{nalbandyan2025score,
  title={SCORE: Systematic COnsistency and Robustness Evaluation for Large Language Models},
  author={Nalbandyan, Grigor and Shahbazyan, Rima and Bakhturina, Evelina},
  journal={arXiv preprint arXiv:2503.00137},
  year={2025}
}

@article{raj2025improving,
  title={Improving consistency in large language models through chain of guidance},
  author={Raj, Harsh and Gupta, Vipul and Rosati, Domenic and Majumdar, Subhabrata},
  journal={arXiv preprint arXiv:2502.15924},
  year={2025}
}

@article{ghosh2024logical,
  title={Logical consistency of large language models in fact-checking},
  author={Ghosh, Bishwamittra and Hasan, Sarah and Arafat, Naheed Anjum and Khan, Arijit},
  journal={arXiv preprint arXiv:2412.16100},
  year={2024}
}

@article{burnell2023rethink,
  title={Rethink reporting of evaluation results in AI},
  author={Burnell, Ryan and Schellaert, Wout and Burden, John and Ullman, Tomer D and Martinez-Plumed, Fernando and Tenenbaum, Joshua B and Rutar, Danaja and Cheke, Lucy G and Sohl-Dickstein, Jascha and Mitchell, Melanie and others},
  journal={Science},
  volume={380},
  number={6641},
  pages={136--138},
  year={2023},
  publisher={American Association for the Advancement of Science}
}

%%%%%%%%%%%%%%%%%%%%%%%%%%%%%%%%%%%%%%%%%%%%%%%%%%%%%%%%%%%%

% \appendix

% \section{Technical appendices and supplementary material}
% Technical appendices with additional results, figures, graphs, and proofs may be submitted with the paper submission before the full submission deadline (see above). You can upload a ZIP file for videos or code, but do not upload a separate PDF file for the appendix. There is no page limit for the technical appendices. 

% Note: Think of the appendix as ``optional reading'' for reviewers. The paper must be able to stand alone without the appendix; for example, adding critical experiments that support the main claims to an appendix is inappropriate. 

\newpage
\appendix
\onecolumn % unnecessary?

\section{Limitations}
\label{app:limitations}
Our analysis relies on a finite set of automatically generated and filtered paraphrases, and therefore does not fully capture the diversity of semantic-preserving variations encountered in real-world interactions. Consequently, the reported measurements should be interpreted as \textbf{conservative lower bounds} on paraphrastic instability. \textit{Notably, substantial inconsistencies already emerge even under these constrained and strictly filtered paraphrase settings.}

Our framework also differs from pass@k-style evaluation settings: we study deterministic responses under semantically equivalent inputs rather than stochastic variation under repeated sampling from the same prompt. As a result, our notion of latent capability reflects knowledge recoverable through semantic rephrasings rather than sampling diversity, making the two settings complementary rather than directly comparable. Our goal is therefore to characterize the semantic sensitivity of model behavior under meaning-preserving rephrasings.

Finally, our experiments are limited to a finite set of datasets and models, and broader evaluation across additional domains, languages, modalities, and multimodal settings remains an important direction for future work.

\section{Details of Paraphrase Generation from a Dataset}
\subsection{Constructing Semantically Equivalent Test Sets}
\label{app:paraphrase-generation}

To generate semantically equivalent variants of each original question, we employ a controlled self-paraphrasing pipeline using an instruction-tuned LLM. Rather than relying on unconstrained paraphrasing, we explicitly guide generation using lightweight, non-adversarial cues designed to induce surface-level linguistic variation while preserving the original intent, assumptions, and required information.

We distinguish between two types of generation cues:

\paragraph{Plain (Structural) Tips.}
Plain paraphrases aim to vary wording and syntax without introducing stylistic or tonal changes. For each paraphrase, we randomly sample a single structural tip from a fixed pool of neutral transformations, including:
\begin{itemize}
    \item avoiding verbatim repetition of key nouns where natural,
    \item reordering clauses or changing the sentence opener,
    \item varying syntactic structure,
    \item substituting safe synonyms for content words,
    \item slightly compressing or expanding the sentence while preserving meaning,
    \item introducing harmless connective words,
    \item switching between statement and question form when applicable,
    \item changing between active and passive voice where natural.
\end{itemize}
Each tip is incorporated into the prompt to encourage a specific form of surface variation without altering semantics.

\paragraph{Toned Paraphrases.}
In addition to plain paraphrases, we generate paraphrases conditioned on an explicit stylistic or tonal attribute. The set of tone cues includes: \emph{formal}, \emph{casual}, \emph{polite}, \emph{confident}, \emph{neutral}, \emph{supportive}, and \emph{inquisitive}. These cues request a particular stylistic realization of the same question while explicitly forbidding changes to meaning, intent, or sentiment. Tone cues are designed to reflect common stylistic variation in natural language usage, rather than adversarial or performance-altering instructions.

\begin{figure}[H]
\centering
\begin{tcolorbox}[
  enhanced,
  colframe=black!50,
  colback=white,
  width=\linewidth,
  arc=2mm,
  boxrule=0.5pt,
  left=10pt,
  right=10pt,
  top=8pt,
  bottom=8pt,
  title=\textbf{Prompt for Plain (Structural) Paraphrasing},
  fonttitle=\bfseries
]
You are a careful, concise rewriter. \\
Paraphrase the user's question below without changing its meaning, intent, or sentiment.
\begin{itemize}
    \item Use different wording and structure.
    \item Keep the output natural and fluent.
    \item Output \textbf{only} the final paraphrase as a single line.
\end{itemize}
Diversity tip: [structural tip] \\
Question: [original question]
\end{tcolorbox}
\label{fig:plain_paraphrase_prompt}
\end{figure}

\begin{figure}[H]
\centering
\begin{tcolorbox}[
  enhanced,
  colframe=black!50,
  colback=white,
  width=\linewidth,
  arc=2mm,
  boxrule=0.5pt,
  left=10pt,
  right=10pt,
  top=8pt,
  bottom=8pt,
  title=\textbf{Prompt for Toned Paraphrasing},
  fonttitle=\bfseries
]
You are a careful, concise rewriter. \\
Rewrite the user's question with the requested tone or attribute, but do \emph{not} change its meaning, intent, assumptions, or sentiment.
\begin{itemize}
    \item Use different wording and structure.
    \item Keep the output natural and fluent.
    \item Output \textbf{only} the final paraphrase as a single line.
\end{itemize}
Tone/attribute: [tone] --- [description] \\
Question: [original question]
\end{tcolorbox}
\label{fig:toned_paraphrase_prompt}
\end{figure}

\subsection{LLM Prompt for Semantic Verification}
\label{app:judge-equivalence}
The prompt template to judge the semantic equivalence of two questions goes as follows:

\begin{figure}[H]
\centering
\begin{tcolorbox}[
  enhanced,
  colframe=black!50,
  colback=white,
  coltitle=black,
  colbacktitle=black!20,
  width=\linewidth,
  arc=2mm,
  auto outer arc,
  boxrule=0.5pt,
  left=10pt,
  right=10pt,
  top=8pt,
  bottom=8pt,
  title=\textbf{Prompt for judging the semantic equivalence of questions},
  fonttitle=\bfseries,
  % fontupper=\ttfamily
]
You are a precise evaluator of meaning. \\
Determine if the candidate paraphrase expresses the SAME intent and meaning as the original question, without adding/removing constraints or changing sentiment. \\
In other words, could they be used interchangeably in the same context? and if they are questions, do they ask the same thing? and imply the same thing? and carry the same assumptions? and can they be both answered the same way?\\
\\
Original: [orig] \\
Candidate: [cand] \\
\\
Respond with EXACTLY one token: YES or NO. \\

\end{tcolorbox}
% \caption{One-stage Self-Paraphrasing prompt template}
\label{fig:judge_equivalence_prompt}
\end{figure}

% \section{Validation Details}
\section{Validation of the Paraphrase Equivalence Pipeline}
\label{app:validation}

We provide additional details on the validation of our semantic equivalence pipeline to further clarify our methodology and address potential concerns regarding semantic equivalence, annotation quality, and model bias.

\paragraph{Human Validation.}
We conducted human annotation over $\sim$1250 paraphrase pairs across datasets (MATH-500: 180, GSM8K: 380, SimpleQA: 460, TruthfulQA: 232). Annotators were asked whether two questions “ask the exact same thing” (binary YES/NO), i.e., whether they can be used interchangeably without changing the answer.

Agreement between the automatic pipeline and human judgments is consistently high
across datasets: MATH-500 (94.6\%), GSM8K (92.4\%), SimpleQA (96.9\%), and TruthfulQA (98.0\%).
% :
% \begin{itemize}[noitemsep, topsep=0pt]
%     % \setlength{\itemsep}{2pt}
%     % \setlength{\parskip}{0pt}
%     % \setlength{\topsep}{2pt}
%     \item MATH-500: 94.6\%
%     \item GSM8K: 92.4\%
%     \item SimpleQA: 96.9\%
%     \item TruthfulQA: 98.0\%
% \end{itemize}
Thus, agreement exceeds 92\% in all cases and 95\%+ in 3 out of 4 datasets, confirming that retained paraphrases preserve semantic meaning.

\paragraph{Stricter Semantic Filtering via Multi-Judge Agreement.}
To assess whether our findings depend on the strictness of semantic filtering and to rule out model-specific bias, we evaluate a stricter validation protocol based on AND-style agreement across independent LLM judges (e.g., \texttt{GPT-4.1-mini} and \texttt{GPT-5}). Under this setup, paraphrases are retained only when all judges agree on semantic equivalence.

We compare three settings: (i) the original single-judge pipeline using \texttt{GPT-4.1-mini}, (ii) a stricter pipeline requiring agreement between \texttt{GPT-4.1-mini} and \texttt{GPT-5.4-mini}, and (iii) a stricter pipeline requiring agreement between \texttt{GPT-4.1-mini} and \texttt{GPT-5}.

On GSM8K, this increases agreement from 92.4\% to $\sim$97.7\%, while preserving all qualitative conclusions. Table~\ref{tab:hidden_instability_gsm8k_judges} shows that the key patterns—small changes in aggregate accuracy alongside substantial mismatch rates—remain consistent across all settings, even under stricter filtering.

\begin{figure}[h]
    \centering
    \includegraphics[width=\linewidth]{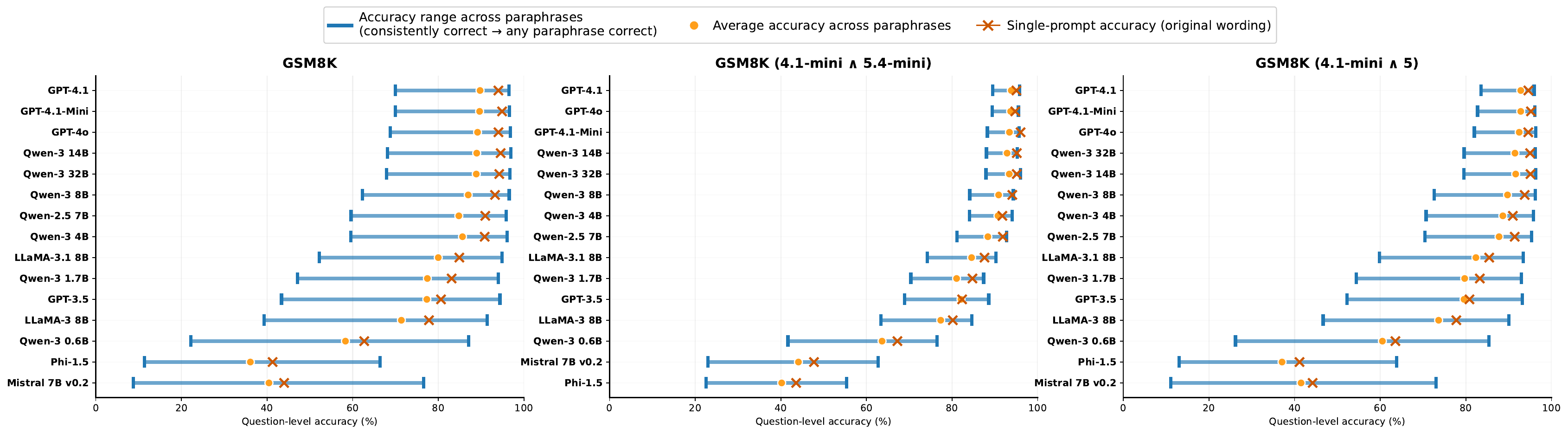}
    % \caption{Performance ranges across paraphrases under different validation settings.}
    \caption{
Performance ranges across paraphrases under progressively stricter semantic filtering. The gap between best-case (latent capability) and worst-case (reliable capability) performance remains substantial across all validation settings, showing that the latent–reliable capability gap is not an artifact of loose paraphrase filtering.
}
    \label{fig:paraphrase_ranges_gsm8k_judges}
\end{figure}

\begin{figure}[H]
    \centering
    \includegraphics[width=\linewidth]{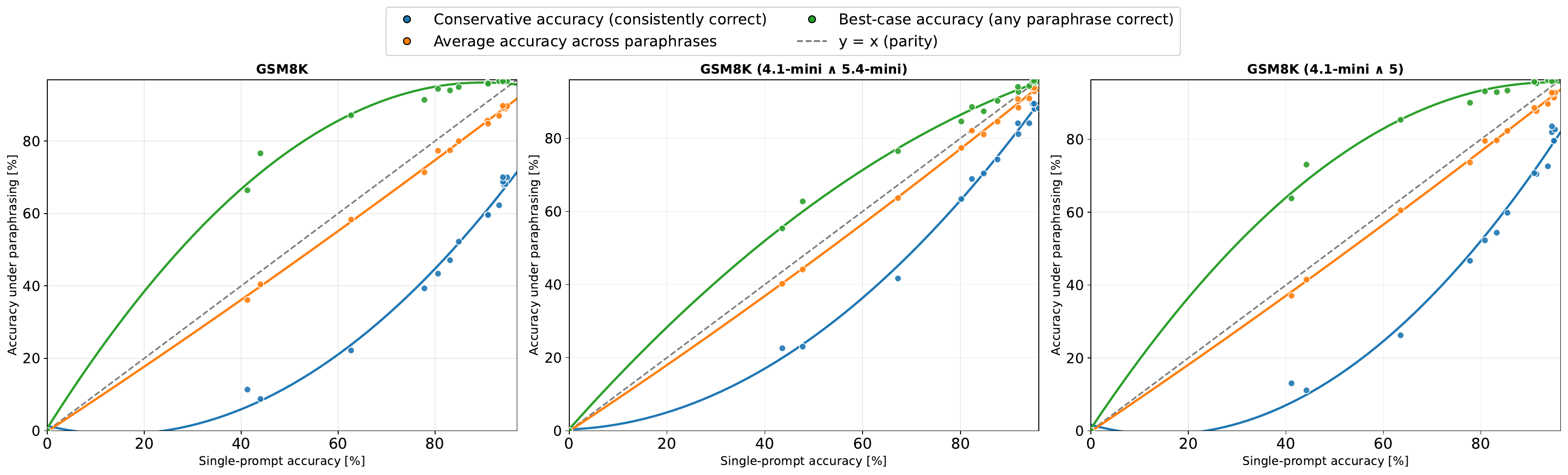}
    % \caption{Relationship between single-prompt accuracy and paraphrase-based metrics under different validation settings.}
    \caption{
Relationship between single-prompt accuracy ($A_{\text{orig}}$) and paraphrase-based metrics under progressively stricter semantic filtering. The qualitative relationships between $A_{\text{orig}}$, $A_{\text{dist}}$, and $A_{\text{any}}$ remain unchanged across validation settings, indicating that observed scaling trends and capability gaps are robust to the validation pipeline.
}
    \label{fig:orig_vs_paraphrased_gsm8k_judges}
\end{figure}

Figure~\ref{fig:paraphrase_ranges_gsm8k_judges} shows that the gap between best-case (latent) and worst-case (reliable) performance remains substantial across all filtering settings. Figure~\ref{fig:orig_vs_paraphrased_gsm8k_judges} shows that the relationships between $A_{\text{orig}}$, $A_{\text{dist}}$, and $A_{\text{any}}$ remain qualitatively unchanged.

These findings indicate that our conclusions are robust to the strictness of the validation pipeline and are not artifacts of loosely constrained paraphrases or model-specific bias.

% <TABLE here??>

\begin{table}[h]
\centering

\caption{
Hidden instability beyond accuracy under progressively stricter semantic filtering. For each setting, we report (i) the change in majority-vote (MV) accuracy relative to the original accuracy ($\Delta A_{\text{MV}}$), and (ii) the mismatch rate—the fraction of instances where the original prediction differs from the MV prediction. Large mismatch with small $\Delta$Acc indicates instability that aggregate accuracy cannot reveal. 
We compare three validation settings: the original single-judge pipeline (\texttt{GPT-4.1-mini}), and stricter AND-style filtering requiring agreement between \texttt{GPT-4.1-mini} and \texttt{GPT-5.4-mini}, or between \texttt{GPT-4.1-mini} and \texttt{GPT-5}. Despite differences in the average number of paraphrases per instance (8.44, 4.37, and 7.30, respectively), the observed instability patterns remain consistent, indicating that our conclusions are robust to the strictness of semantic filtering and are not artifacts of paraphrase count or validation strategy.
}
\label{tab:hidden_instability_gsm8k_judges}

\resizebox{0.95\linewidth}{!}{%
\begin{tabular}{l|cc|cc|cc}
\toprule

\multicolumn{1}{c}{\multirow{2}{*}{\textbf{Model}}}
& \multicolumn{2}{c}{\textbf{GSM8K (4.1-mini only)}}
& \multicolumn{2}{c}{\textbf{GSM8K (4.1-mini $\land$ 5.4-mini)}}
& \multicolumn{2}{c}{\textbf{GSM8K (4.1-mini $\land$ 5)}} \\

\cmidrule(lr){2-3}
\cmidrule(lr){4-5}
\cmidrule(lr){6-7}

\multicolumn{1}{c}{}
& \textbf{$\Delta A_{\text{MV}}$} & \textbf{Mismatch Rate}
& \textbf{$\Delta A_{\text{MV}}$} & \textbf{Mismatch Rate}
& \textbf{$\Delta A_{\text{MV}}$} & \textbf{Mismatch Rate} \\

\midrule
GPT-3.5 & 3.19 & 12.70 & 3.28 & 11.56 & 4.39 & 12.77 \\
GPT-4.1 & -0.10 & 3.09 & 0.24 & 2.67 & 0.10 & 1.98 \\
GPT-4.1-Mini & -1.11 & 2.66 & -1.08 & 2.18 & -0.21 & 1.35 \\
GPT-4o & -0.51 & 2.88 & 0.60 & 2.69 & 0.21 & 2.08 \\
LLaMA-3 8B & -1.55 & 13.68 & 0.92 & 12.62 & -0.34 & 13.10 \\
LLaMA-3.1 8B & -0.21 & 10.64 & 0.51 & 9.36 & 1.62 & 9.51 \\
Qwen-2.5 7B & -1.92 & 6.13 & -2.51 & 6.69 & -0.82 & 5.45 \\
Qwen-3 0.6B & 0.00 & 17.70 & 3.25 & 19.15 & 2.28 & 18.78 \\
Qwen-3 1.7B & -0.30 & 11.40 & -0.96 & 10.50 & 1.13 & 10.77 \\
Qwen-3 4B & 0.81 & 5.67 & 0.84 & 5.34 & 1.03 & 5.14 \\
Qwen-3 8B & -1.01 & 4.78 & -1.79 & 4.12 & -0.62 & 4.17 \\
Qwen-3 14B & -1.22 & 3.29 & -0.84 & 3.17 & -0.72 & 2.82 \\
Qwen-3 32B & -1.82 & 4.12 & -0.36 & 2.69 & -1.13 & 2.91 \\
Mistral 7B v0.2 & -1.64 & 22.34 & 2.22 & 23.14 & 0.45 & 22.53 \\
Phi-1.5 & -3.47 & 18.31 & 1.15 & 19.12 & -1.46 & 17.71 \\
\bottomrule
\end{tabular}
}

\end{table}

\paragraph{Annotation Setup.}
Human annotation was conducted by 5 annotators. Each annotator was presented with pairs of questions and asked to determine whether they ask the exact same thing. 

\paragraph{Summary.}
Overall, our validation pipeline combines human annotation, LLM-based semantic verification, and NLI filtering. High agreement rates and consistent results under stricter validation protocols provide strong evidence that our paraphrase sets preserve semantic equivalence, ruling out the possibility that the observed inconsistencies arise from loosely constrained or invalid paraphrases.

\section{Additional Stability Metrics}
\label{app:stability-metrics}

To further characterize the distribution of model predictions across paraphrases, we report additional diagnostic metrics.

\begin{itemize}

    \item \textbf{Normalized Entropy:} 
    We quantify dispersion in the label distribution via a length-normalized entropy:
    \[
        H_i \;=\; -\frac{1}{\log(k_i)} \sum_{l \in \mathcal{L}} P_i^l \log P_i^l .
    \]
    Higher entropy indicates greater variability in predictions across paraphrases, while values near zero correspond to near-deterministic behavior.

    \item \textbf{IID-Mismatch Rate:} 
    Defined as
    \[
        1 - \sum_{l \in \mathcal{L}} (P_i^l)^2,
    \]
    this metric measures the probability that two independently sampled paraphrases of the same question receive different labels, capturing pairwise inconsistency in model behavior.

    \item \textbf{Average Mode Share:} 
    For each question, we define the mode share as
    \[
        \max_{l \in \mathcal{L}} P_i^l,
    \]
    i.e., the proportion of paraphrases assigned to the most frequent label, reflecting the dominance of the model’s primary prediction.
    We report the average mode share by averaging this quantity across all instances in the dataset.
    % \item \textbf{Non-Determinism (ND) Rate:} The fraction of sets $\mathcal{Q}_i$ where the model produces at least two distinct labels (i.e., absolute disagreement), regardless of correctness.

\end{itemize}

\section{Answer Evaluation Protocol}
\label{app:evaluation}

We describe the procedure used to map model responses to a three-way label space (Correct, Incorrect, Not Attempted) for factual QA datasets.

\subsection{Datasets and Evaluation Setup}

\paragraph{SimpleQA.}
For SimpleQA, we adopt the official evaluation protocol and prompts provided in the dataset repository. Model responses are evaluated against reference answers using the prescribed criteria.

\paragraph{TruthfulQA.}
For TruthfulQA, we follow the dataset guidelines and design a high-quality evaluation prompt to assess correctness. The prompt ensures that responses are judged based on factual accuracy and alignment with ground-truth answers, while accounting for acceptable variations in phrasing.

\subsection{Multi-Judge Evaluation Framework}

To improve evaluation robustness and reduce dependence on any single evaluator, we employ a multi-judge agreement-based evaluation framework.

Specifically, we use four independent judge models:
\begin{itemize}[noitemsep, topsep=0pt]
    \item \texttt{GPT-4.1-mini}
    \item \texttt{GPT-3.5-turbo}
    \item \texttt{Qwen2.5-7B}
    \item \texttt{Qwen3-8B}
\end{itemize}

Each judge independently assigns one of three labels (\textit{Correct}, \textit{Incorrect}, or \textit{Not Attempted}) to a model response.

Final labels are determined using agreement-based validation across judges. Specifically, labels are accepted only when at least three of the four judges assign the same label, including agreement between the stronger closed-source judges (\texttt{GPT-4.1-mini} and \texttt{GPT-3.5-turbo}).

The resulting label is used for downstream aggregation across paraphrases. By requiring consistent agreement across multiple independent evaluators, this protocol reduces evaluator-specific bias and ambiguous labeling decisions, yielding more robust and reliable estimates of model correctness.

\subsection{Consistency Across Paraphrases}

To ensure that differences in labels across paraphrases reflect model behavior rather than evaluation artifacts, we enforce consistent evaluation conditions across paraphrases:
\begin{itemize}[noitemsep, topsep=0pt]
    \item identical prompts are used across all paraphrases of a given question,
    \item normalization and formatting of responses are applied uniformly,
    \item evaluation criteria remain fixed across datasets and runs.
\end{itemize}

\subsection{Remarks on Efficiency}

The multi-judge setup also enables a favorable trade-off between evaluation quality and cost. By combining multiple lower-cost models with stronger closed-source judges, we achieve reliable evaluation while avoiding exclusive reliance on high-cost models.

\section{Why $A_{\text{any}}$ Reflects Latent Knowledge}
\label{app:latent_knowledge_any}

The metric $A_{\text{any}}$ measures whether a model produces a correct answer for at least one paraphrase $Q_i^j \in \mathcal{Q}_i$ of a given question $Q_i$.
Intuitively, this captures whether the model \emph{possesses} the relevant knowledge or reasoning capability, irrespective of whether it applies that knowledge consistently.

In our setting, models generate answers via free-form decoding over a large output space.
For factual or mathematical questions, the probability of producing a correct answer by random chance is extremely small.
For example, when asked ``What is the capital of France?'', even conditioning on the answer being a city name leaves thousands of plausible candidates, making correct guessing unlikely.
Therefore, observing at least one correct response among paraphrases provides strong evidence that the model has internalized the required information or reasoning pattern.

Importantly, $A_{\text{any}}$ does not imply robustness or reliability.
Rather, it serves as a permissive indicator of latent capability, which we contrast with the stricter requirement imposed by $A_{\text{strict}}$.

\section{Prompt for One-stage Self-Paraphrasing}
\label{append:self-paraphrase-prompt}
% \subsection{Self-Paraphrasing for Test-Time Consistency}
% \label{app:self_paraphrasing}

% In this section, we examine whether Large Language Models (LLMs) can improve answer correctness and consistency by generating and conditioning on their own rephrasings of a user query at test time. Given a question from the \textit{SimpleQA} dataset, models are prompted to first generate several paraphrases of the original question that preserve its semantic meaning, and then answer the provided question. These generated paraphrases can be used as additional context to help the model better infer the user’s intent before producing a final answer, effectively acting as a test-time alignment step over the input space.

% \paragraph{Experimental Setup.} We explore two variants of this approach. In the \textbf{two-stage prompting} setting, the model is first prompted to generate paraphrases of the original question. In a second prompt, the model is asked to solve the task while conditioning on the original question together with the generated paraphrases. In the \textbf{one-stage prompting} setting, the model is prompted once to both generate paraphrases and answer the question within a single request. Across all evaluated models, we observe that the two strategies yield very similar results. 

% The prompt template used in the one-stage Self-Paraphrasing setting is provided as follows.

\begin{figure}[H]
\centering
\begin{tcolorbox}[
  enhanced,
  colframe=black!50,
  colback=white,
  coltitle=black,
  colbacktitle=black!20,
  width=\linewidth,
  arc=2mm,
  auto outer arc,
  boxrule=0.5pt,
  left=10pt,
  right=10pt,
  top=8pt,
  bottom=8pt,
  title=\textbf{Prompt for Self-Paraphrasing},
  fonttitle=\bfseries,
  % fontupper=\ttfamily
]
You are a careful, factual assistant. Answer concisely.\\
If the question is ambiguous or unanswerable, say you don't know. \\
If you don't know the answer, say you don't know. \\

You will follow these steps: \\
1. Generate several paraphrases of the question.\\
2. Use the original question and paraphrases as context
   to better understand the user's intent. \\
3. Provide a single final answer. \\

Question:
[original question]

\end{tcolorbox}
% \caption{One-stage Self-Paraphrasing prompt template}
\label{fig:self_paraphrasing_prompt}
\end{figure}

\section{Additional Discussion}
\label{app:discussion}

% Our findings also have implications for how benchmark contamination is interpreted. Strong performance on benchmark questions is often attributed to memorization of training data, particularly when models answer benchmark items with high accuracy. However, the instability we observe under paraphrasing complicates this interpretation. In many cases, models answer the original benchmark wording correctly but fail on semantically equivalent paraphrases, while in other cases the opposite occurs: the original question is answered incorrectly but a paraphrased version is answered correctly. 

% This bidirectional behavior suggests that performance on benchmark items cannot be explained solely by memorization. Instead, sensitivity to surface form and inconsistent knowledge retrieval can produce outcomes that resemble contamination effects. As a result, apparent contamination signals may partly reflect paraphrase sensitivity rather than purely memorized benchmark data.

\subsection{Paraphrase Sensitivity and Benchmark Contamination}

Our findings also affect how benchmark contamination is interpreted. It's often argued that existing popular benchmarks suffer from test set contamination, and that current models have effectively ``seen'' almost all public test samples during training~\cite{golchin2023time, golchin2024datacontaminationquiztool, dekoninck2024constat, singh2024evaluationdatacontaminationllms,cheng-etal-2025-dyepack}. Therefore, models will often only perform well when tested with the given wordings in a benchmark test set, and will suffer a performance drop when the wordings are changed. However, we observed that while models may sometimes answer the original wording correctly while failing on semantically equivalent paraphrases, in many other cases models may succeed on paraphrases after failing on the original question. This \textbf{bidirectional} behavior suggests that performance on benchmark items cannot be explained solely by memorization or contamination. Instead, sensitivity to surface form and inconsistent knowledge retrieval can produce outcomes that resemble contamination effects. As a result, apparent contamination signals may partly reflect paraphrase sensitivity rather than purely memorized benchmark data.

\subsection{Implications for Real-World Deployment}

Paraphrase sensitivity also has important implications for the deployment of language models in real-world settings. In many applications, users expect semantically equivalent questions to produce consistent answers. When model behavior varies across paraphrases, system reliability becomes dependent on how a query is phrased rather than on the underlying information requested. This phenomenon also raises fairness concerns: users differ widely in how they express queries due to language proficiency, cultural background, or dialectal variation. As a result, effective use of language models may implicitly depend on prompt engineering skill, which undermines the goal of making these systems broadly accessible and dependable.

\subsection{Mechanisms Behind Paraphrase Sensitivity}

The instability observed under paraphrasing likely arises from several interacting mechanisms in current language models. Semantically equivalent queries may map to different regions of the model’s representation space, activating different attention patterns and token prediction trajectories. As a result, the same question can trigger different retrieval paths even when the relevant knowledge is encoded in the model, consistent with the gap between latent capability ($A_\text{any}$) and reliable capability ($A_\text{strict}$). In addition, models may rely on surface heuristics or prompt-specific patterns learned during training, causing small variations in wording to activate different associations. Together, these factors suggest that knowledge retrieval in current LLMs remains sensitive to surface form rather than strictly invariant to semantic equivalence.

% \subsection{Paraphrase Sensitivity and Benchmark Contamination}

\section{Compute Resources}

Experiments involved both API-based and locally hosted model evaluations. Local inference experiments were conducted on a shared GPU cluster with heterogeneous hardware, including NVIDIA RTX5000, RTX6000, and L40S GPUs, with some evaluations using multi-GPU configurations. Open-source models were evaluated using the HuggingFace Transformers and vLLM frameworks, while proprietary models were accessed through their respective APIs. Since the work focuses on inference-time evaluation of pretrained models, no model training was performed.

\section{Detailed Result Tables and Figures}
\label{append:more_result_tables}

\begin{figure*}[h]
    \centering
    \includegraphics[width=0.95\linewidth]{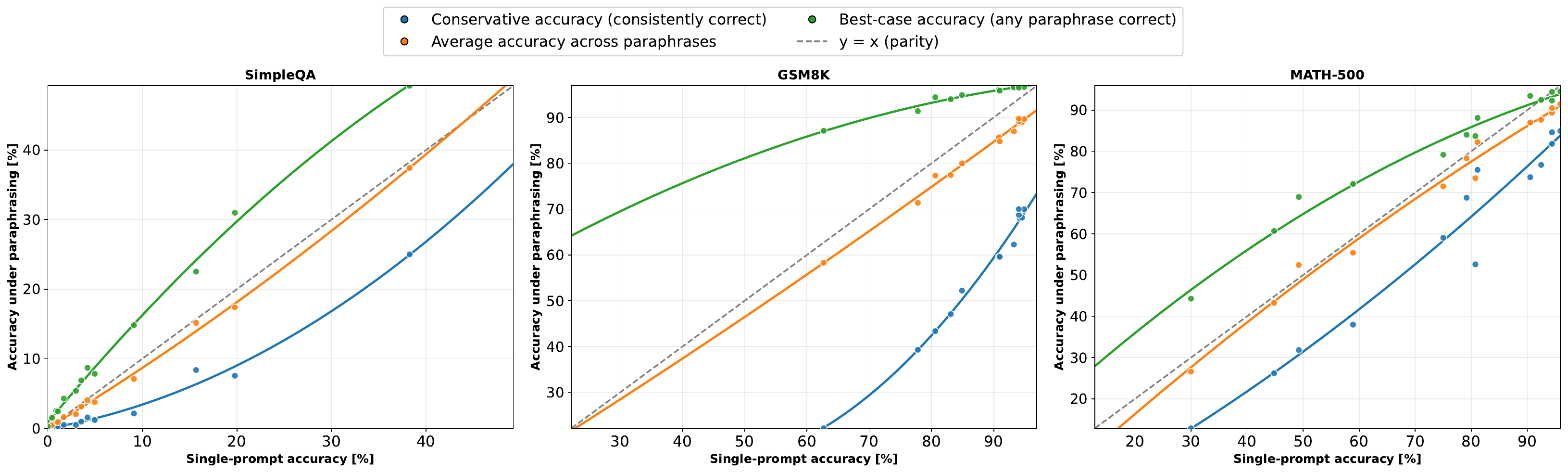}
    \caption{
    Relationship between single-prompt accuracy ($A_{\text{orig}}$) and three capability metrics across all remaining evaluated datasets.
    Each subplot corresponds to a dataset and shows Latent Capability ($A_{\text{any}}$), Average Accuracy across paraphrases ($A_{\text{dist}}$), and Reliable Capability ($A_{\text{strict}}$).
    Consistent with the trends illustrated for \textsc{TruthfulQA} in the main text, single-prompt accuracy lies within a capability band bounded by reliable consistency and latent knowledge.
    }
    \label{fig:metric_correlations}
\end{figure*}

\begin{figure*}[h]
    \centering
    \includegraphics[width=0.95\linewidth]{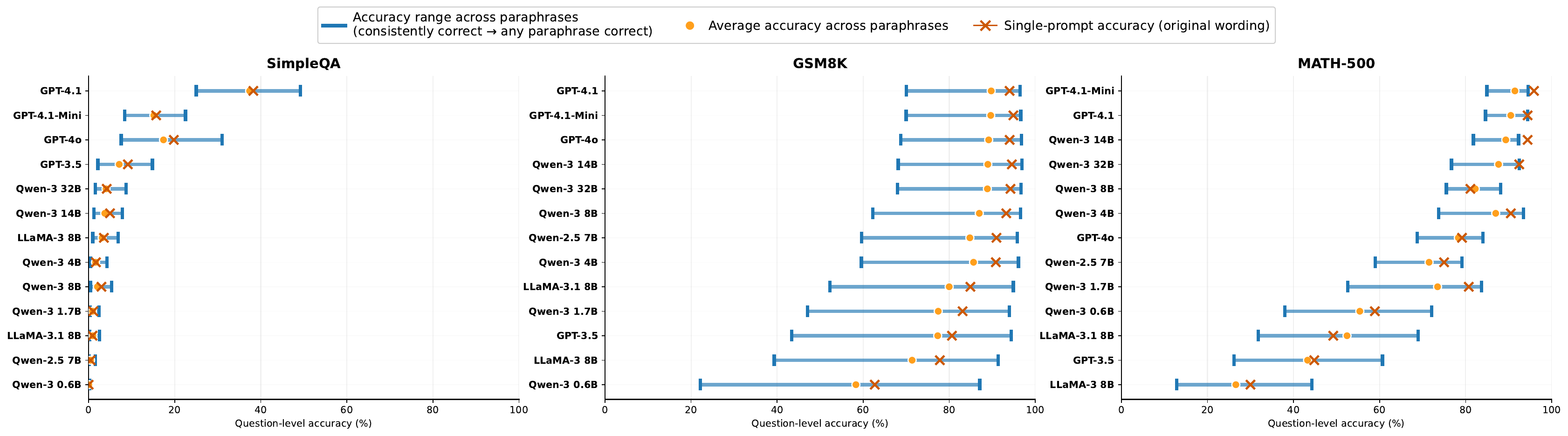}
    \caption{
    Accuracy ranges across meaning-preserving paraphrases for all remaining evaluated datasets.
    Horizontal bars show the range between reliable capability ($A_{\text{strict}}$) and latent capability ($A_{\text{any}}$), with the dot indicating average accuracy across paraphrases and the cross marking single-prompt accuracy.
    Similar performance bands appear across datasets, indicating that the gap between reliable and latent capability is a consistent phenomenon.
    }
    \label{fig:paraphrase_ranges}
\end{figure*}

\begin{figure*}[h]
    \centering
    \includegraphics[width=1\linewidth]{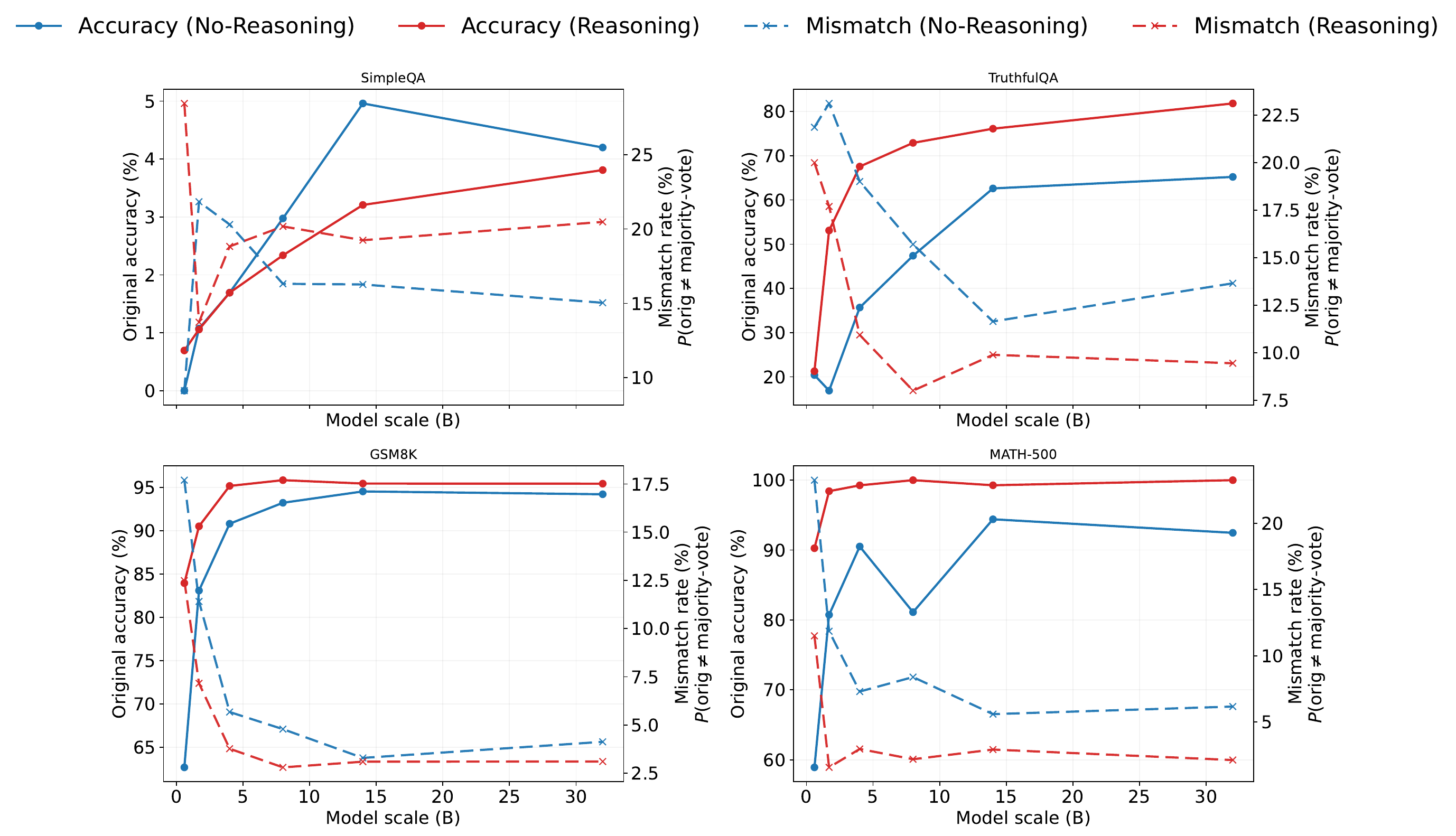}
    \caption{The effect of model scale and reasoning mode on performance. Accuracy (solid lines) generally improves with scale, while the mismatch rate (dashed lines) declines, indicating better consistency. CoT reasoning further enhances stability across all scales.}
    \label{fig:per-model-qwen3}
\end{figure*}

\begin{table*}[h]
\centering
\caption{Aggregate label-distribution shifts under paraphrasing (SimpleQA). We compare the proportions of correct (A), incorrect (B), and abstain (C) predictions on the original questions with the distributions obtained by evaluating all paraphrases individually. $\Delta$A/B/C denote the change in each category.
}
\label{tab:simpleqa-dist-aggregate}

\resizebox{\textwidth}{!}{%
\begin{tabular}{lcccccccccc}
\toprule
model & A (Orig) & B (Orig) & C (Orig) & A (Para) & B (Para) & C (Para) & $\Delta$A & $\Delta$B & $\Delta$C & N \\
\midrule
GPT-3.5 & 9.1 & 36.5 & 54.4 & 7.1 & 27.8 & 65.1 & \textcolor{red}{\textbf{-2.0}} & -8.8 & 10.8 & 1032 \\
GPT-4.1 & 38.3 & 50.0 & 11.7 & 37.4 & 48.7 & 13.9 & \textcolor{red}{-0.9} & -1.3 & 2.2 & 1040 \\
GPT-4.1-Mini & 15.7 & 79.9 & 4.4 & 15.2 & 79.5 & 5.3 & \textcolor{red}{-0.5} & -0.4 & 0.9 & 1039 \\
GPT-4o & 19.8 & 24.4 & 55.8 & 17.4 & 19.1 & 63.6 & \textcolor{red}{\textbf{-2.4}} & -5.3 & 7.7 & 1046 \\
LLaMA-3 8B & 3.5 & 39.0 & 57.5 & 3.1 & 41.9 & 54.9 & \textcolor{red}{-0.4} & 2.9 & -2.5 & 1046 \\
LLaMA-3.1 8B & 0.9 & 3.2 & 95.9 & 1.0 & 3.3 & 95.7 & 0.1 & 0.0 & -0.1 & 1115 \\
Qwen-2.5 7B & 0.4 & 2.5 & 97.1 & 0.5 & 2.5 & 97.0 & 0.0 & 0.0 & -0.0 & 1124 \\
Qwen-3 0.6B & 0.0 & 9.7 & 90.3 & 0.0 & 3.8 & 96.2 & 0.0 & -5.9 & 5.9 & 1106 \\
Qwen-3 0.6B +Thinking & 0.7 & 61.2 & 38.2 & 0.3 & 53.4 & 46.3 & \textcolor{red}{-0.4} & -7.8 & 8.1 & 865 \\
Qwen-3 1.7B & 1.1 & 36.5 & 62.4 & 0.9 & 30.2 & 68.9 & \textcolor{red}{-0.2} & -6.4 & 6.6 & 1026 \\
Qwen-3 1.7B +Thinking & 1.1 & 77.8 & 21.1 & 1.4 & 76.2 & 22.4 & 0.4 & -1.6 & 1.3 & 852 \\
Qwen-3 4B & 1.7 & 49.9 & 48.5 & 1.6 & 42.5 & 55.9 & \textcolor{red}{-0.1} & -7.4 & 7.5 & 1005 \\
Qwen-3 4B +Thinking & 1.7 & 71.7 & 26.6 & 1.6 & 69.3 & 29.2 & \textcolor{red}{-0.1} & -2.5 & 2.5 & 887 \\
Qwen-3 8B & 3.0 & 48.0 & 49.0 & 2.0 & 48.2 & 49.8 & \textcolor{red}{-0.9} & 0.2 & 0.8 & 1042 \\
Qwen-3 8B +Thinking & 2.3 & 39.9 & 57.8 & 2.1 & 42.6 & 55.3 & \textcolor{red}{-0.2} & 2.8 & -2.5 & 813 \\
Qwen-3 14B & 5.0 & 70.2 & 24.8 & 3.7 & 61.3 & 35.0 & \textcolor{red}{\textbf{-1.2}} & -9.0 & 10.2 & 1008 \\
Qwen-3 14B +Thinking & 3.2 & 36.8 & 60.0 & 2.6 & 35.0 & 62.4 & \textcolor{red}{-0.6} & -1.8 & 2.3 & 873 \\
Qwen-3 32B & 4.2 & 67.8 & 28.0 & 4.0 & 61.4 & 34.6 & \textcolor{red}{-0.2} & -6.4 & 6.5 & 1024 \\
Qwen-3 32B +Thinking & 3.8 & 50.1 & 46.1 & 3.3 & 50.6 & 46.1 & \textcolor{red}{-0.6} & 0.5 & 0.1 & 840 \\
\bottomrule
\end{tabular}
}

\end{table*}

\begin{table*}[t]
\centering
\caption{
Class–label shifts under paraphrasing (SimpleQA). 
We compare the distribution of predicted labels for the original questions with 
the majority-vote label obtained from their paraphrases. 
$\Delta$A/B/C denote the change in the proportion of correct, incorrect, and abstain responses.
}
\label{tab:simpleqa-dist-majority}
\resizebox{\textwidth}{!}{%
\begin{tabular}{lcccccccccc}
\toprule
model & A (Orig) & B (Orig) & C (Orig) & A (MV) & B (MV) & C (MV) & $\Delta$A & $\Delta$B & $\Delta$C & N \\
\midrule
GPT-3.5 & 9.1 & 36.5 & 54.4 & 6.8 & 26.6 & 66.6 & \textcolor{red}{\textbf{-2.3}} & -9.9 & 12.2 & 1032 \\
GPT-4.1 & 38.3 & 50.0 & 11.7 & 38.4 & 49.2 & 12.4 & 0.1 & -0.8 & 0.7 & 1040 \\
GPT-4.1-Mini & 15.7 & 79.9 & 4.4 & 15.1 & 81.0 & 3.8 & \textcolor{red}{-0.6} & 1.2 & -0.6 & 1039 \\
GPT-4o & 19.8 & 24.4 & 55.8 & 16.3 & 18.1 & 65.6 & \textcolor{red}{\textbf{-3.4}} & -6.3 & 9.8 & 1046 \\
LLaMA-3 8B & 3.5 & 39.0 & 57.5 & 2.9 & 42.3 & 54.9 & \textcolor{red}{-0.7} & 3.3 & -2.6 & 1046 \\
LLaMA-3.1 8B & 0.9 & 3.2 & 95.9 & 0.9 & 2.7 & 96.4 & 0.0 & -0.5 & 0.5 & 1115 \\
Qwen-2.5 7B & 0.4 & 2.5 & 97.1 & 0.3 & 1.7 & 98.0 & \textcolor{red}{-0.2} & -0.8 & 1.0 & 1124 \\
Qwen-3 0.6B & 0.0 & 9.7 & 90.3 & 0.0 & 1.6 & 98.4 & 0.0 & -8.0 & 8.0 & 1106 \\
Qwen-3 0.6B +Thinking & 0.7 & 61.2 & 38.2 & 0.1 & 54.7 & 45.2 & \textcolor{red}{-0.6} & -6.5 & 7.1 & 865 \\
Qwen-3 1.7B & 1.1 & 36.5 & 62.4 & 0.6 & 27.0 & 72.4 & \textcolor{red}{-0.5} & -9.6 & 10.0 & 1026 \\
Qwen-3 1.7B +Thinking & 1.1 & 77.8 & 21.1 & 0.6 & 79.7 & 19.7 & \textcolor{red}{-0.5} & 1.9 & -1.4 & 852 \\
Qwen-3 4B & 1.7 & 49.9 & 48.5 & 1.3 & 41.8 & 56.9 & \textcolor{red}{-0.4} & -8.1 & 8.5 & 1005 \\
Qwen-3 4B +Thinking & 1.7 & 71.7 & 26.6 & 1.1 & 72.6 & 26.3 & \textcolor{red}{-0.6} & 0.9 & -0.3 & 887 \\
Qwen-3 8B & 3.0 & 48.0 & 49.0 & 1.5 & 48.1 & 50.4 & \textcolor{red}{\textbf{-1.4}} & 0.1 & 1.3 & 1042 \\
Qwen-3 8B +Thinking & 2.3 & 39.9 & 57.8 & 2.0 & 41.3 & 56.7 & \textcolor{red}{-0.4} & 1.5 & -1.1 & 813 \\
Qwen-3 14B & 5.0 & 70.2 & 24.8 & 3.6 & 64.0 & 32.4 & \textcolor{red}{\textbf{-1.4}} & -6.2 & 7.6 & 1008 \\
Qwen-3 14B +Thinking & 3.2 & 36.8 & 60.0 & 2.1 & 33.1 & 64.8 & \textcolor{red}{\textbf{-1.1}} & -3.7 & 4.8 & 873 \\
Qwen-3 32B & 4.2 & 67.8 & 28.0 & 3.5 & 63.9 & 32.6 & \textcolor{red}{-0.7} & -3.9 & 4.6 & 1024 \\
Qwen-3 32B +Thinking & 3.8 & 50.1 & 46.1 & 2.9 & 50.5 & 46.7 & \textcolor{red}{-1.0} & 0.4 & 0.6 & 840 \\
\bottomrule
\end{tabular}
}

\end{table*}

\begin{table*}[t]
\centering
\caption{Aggregate label-distribution shifts under paraphrasing (TruthfulQA). We compare the proportions of correct (A), incorrect (B), and abstain (C) predictions on the original questions with the distributions obtained by evaluating all paraphrases individually. $\Delta$A/B/C denote the change in each category.
}
\label{tab:truthfulqa-dist-aggregate}
\resizebox{\textwidth}{!}{%
\begin{tabular}{lcccccccccc}
\toprule
model & A (Orig) & B (Orig) & C (Orig) & A (Para) & B (Para) & C (Para) & $\Delta$A & $\Delta$B & $\Delta$C & N \\
\midrule
GPT-3.5 & 52.3 & 25.5 & 22.2 & 51.0 & 26.1 & 22.9 & \textcolor{red}{\textbf{-1.3}} & 0.5 & 0.7 & 419 \\
GPT-4.1 & 76.3 & 14.1 & 9.6 & 73.7 & 14.9 & 11.4 & \textcolor{red}{\textbf{-2.6}} & 0.9 & 1.7 & 405 \\
GPT-4.1-Mini & 70.6 & 24.8 & 4.6 & 68.2 & 25.1 & 6.7 & \textcolor{red}{\textbf{-2.4}} & 0.3 & 2.2 & 395 \\
GPT-4o & 56.6 & 12.4 & 31.1 & 55.8 & 13.8 & 30.4 & \textcolor{red}{-0.8} & 1.4 & -0.7 & 396 \\
LLaMA-3 8B & 28.8 & 29.0 & 42.2 & 29.6 & 31.3 & 39.0 & 0.9 & 2.3 & -3.1 & 396 \\
LLaMA-3.1 8B & 30.9 & 26.2 & 42.9 & 28.8 & 27.5 & 43.7 & \textcolor{red}{\textbf{-2.1}} & 1.3 & 0.8 & 385 \\
Qwen-2.5 7B & 44.8 & 17.0 & 38.2 & 43.3 & 17.0 & 39.7 & \textcolor{red}{\textbf{-1.5}} & -0.0 & 1.5 & 382 \\
Qwen-3 0.6B & 20.4 & 29.4 & 50.1 & 21.3 & 20.7 & 58.0 & 0.9 & -8.7 & 7.9 & 343 \\
Qwen-3 0.6B +Thinking & 21.3 & 49.8 & 28.9 & 20.5 & 50.4 & 29.1 & \textcolor{red}{-0.7} & 0.6 & 0.1 & 235 \\
Qwen-3 1.7B & 16.9 & 32.7 & 50.4 & 18.3 & 30.7 & 51.0 & 1.4 & -2.0 & 0.6 & 385 \\
Qwen-3 1.7B +Thinking & 53.1 & 37.2 & 9.7 & 52.0 & 35.1 & 12.9 & \textcolor{red}{\textbf{-1.1}} & -2.1 & 3.2 & 226 \\
Qwen-3 4B & 35.7 & 37.8 & 26.6 & 38.6 & 34.1 & 27.3 & \textbf{2.9} & -3.6 & 0.8 & 384 \\
Qwen-3 4B +Thinking & 67.5 & 21.1 & 11.3 & 64.3 & 23.4 & 12.3 & \textcolor{red}{\textbf{-3.2}} & 2.3 & 1.0 & 265 \\
Qwen-3 8B & 47.4 & 34.0 & 18.6 & 49.9 & 33.2 & 16.9 & 2.5 & -0.8 & -1.7 & 382 \\
Qwen-3 8B +Thinking & 72.9 & 13.7 & 13.4 & 72.3 & 14.0 & 13.7 & \textcolor{red}{-0.6} & 0.3 & 0.3 & 262 \\
Qwen-3 14B & 62.6 & 31.7 & 5.7 & 60.8 & 29.9 & 9.2 & \textcolor{red}{\textbf{-1.8}} & -1.8 & 3.5 & 369 \\
Qwen-3 14B +Thinking & 76.1 & 13.0 & 10.9 & 74.3 & 12.6 & 13.1 & \textcolor{red}{\textbf{-1.8}} & -0.4 & 2.2 & 293 \\
Qwen-3 32B & 65.2 & 24.0 & 10.8 & 61.3 & 26.7 & 12.0 & \textcolor{red}{\textbf{-3.9}} & 2.7 & 1.1 & 388 \\
Qwen-3 32B +Thinking & 81.8 & 11.3 & 6.9 & 80.3 & 14.0 & 5.7 & \textcolor{red}{\textbf{-1.6}} & 2.8 & -1.2 & 275 \\
\bottomrule
\end{tabular}
}

\end{table*}

\begin{table*}[t]
\centering
\caption{
Class–label shifts under paraphrasing (TruthfulQA). 
We compare the distribution of predicted labels for the original questions with 
the majority-vote label obtained from their paraphrases. 
$\Delta$A/B/C denote the change in the proportion of correct, incorrect, and abstain responses.
}
\label{tab:truthfulqa-dist-majority}
\resizebox{\textwidth}{!}{%
\begin{tabular}{lcccccccccc}
\toprule
model & A (Orig) & B (Orig) & C (Orig) & A (MV) & B (MV) & C (MV) & $\Delta$A & $\Delta$B & $\Delta$C & N \\
\midrule
GPT-3.5 & 52.3 & 25.5 & 22.2 & 52.7 & 24.3 & 22.9 & 0.5 & -1.2 & 0.7 & 419 \\
GPT-4.1 & 76.3 & 14.1 & 9.6 & 75.1 & 14.3 & 10.6 & \textcolor{red}{\textbf{-1.2}} & 0.2 & 1.0 & 405 \\
GPT-4.1-Mini & 70.6 & 24.8 & 4.6 & 69.9 & 24.6 & 5.6 & \textcolor{red}{-0.8} & -0.3 & 1.0 & 395 \\
GPT-4o & 56.6 & 12.4 & 31.1 & 56.3 & 13.1 & 30.6 & \textcolor{red}{-0.3} & 0.8 & -0.5 & 396 \\
LLaMA-3 8B & 28.8 & 29.0 & 42.2 & 28.8 & 31.3 & 39.9 & 0.0 & 2.3 & -2.3 & 396 \\
LLaMA-3.1 8B & 30.9 & 26.2 & 42.9 & 26.8 & 27.0 & 46.2 & \textcolor{red}{\textbf{-4.2}} & 0.8 & 3.4 & 385 \\
Qwen-2.5 7B & 44.8 & 17.0 & 38.2 & 43.5 & 16.0 & 40.6 & \textcolor{red}{\textbf{-1.3}} & -1.0 & 2.4 & 382 \\
Qwen-3 0.6B & 20.4 & 29.4 & 50.1 & 19.0 & 19.2 & 61.8 & \textcolor{red}{\textbf{-1.5}} & -10.2 & 11.7 & 343 \\
Qwen-3 0.6B +Thinking & 21.3 & 49.8 & 28.9 & 20.4 & 50.2 & 29.4 & \textcolor{red}{-0.9} & 0.4 & 0.4 & 235 \\
Qwen-3 1.7B & 16.9 & 32.7 & 50.4 & 16.9 & 28.8 & 54.3 & 0.0 & -3.9 & 3.9 & 385 \\
Qwen-3 1.7B +Thinking & 53.1 & 37.2 & 9.7 & 53.5 & 34.5 & 11.9 & 0.4 & -2.7 & 2.2 & 226 \\
Qwen-3 4B & 35.7 & 37.8 & 26.6 & 37.8 & 34.9 & 27.3 & 2.1 & -2.9 & 0.8 & 384 \\
Qwen-3 4B +Thinking & 67.5 & 21.1 & 11.3 & 66.4 & 21.9 & 11.7 & \textcolor{red}{\textbf{-1.1}} & 0.8 & 0.4 & 265 \\
Qwen-3 8B & 47.4 & 34.0 & 18.6 & 50.8 & 33.0 & 16.2 & \textbf{3.4} & -1.0 & -2.4 & 382 \\
Qwen-3 8B +Thinking & 72.9 & 13.7 & 13.4 & 73.3 & 13.7 & 13.0 & 0.4 & 0.0 & -0.4 & 262 \\
Qwen-3 14B & 62.6 & 31.7 & 5.7 & 63.1 & 28.7 & 8.1 & 0.5 & -3.0 & 2.4 & 369 \\
Qwen-3 14B +Thinking & 76.1 & 13.0 & 10.9 & 76.5 & 11.9 & 11.6 & 0.3 & -1.0 & 0.7 & 293 \\
Qwen-3 32B & 65.2 & 24.0 & 10.8 & 62.4 & 26.0 & 11.6 & \textcolor{red}{\textbf{-2.8}} & 2.1 & 0.8 & 388 \\
Qwen-3 32B +Thinking & 81.8 & 11.3 & 6.9 & 81.8 & 13.1 & 5.1 & 0.0 & 1.8 & -1.8 & 275 \\
\bottomrule
\end{tabular}
}

\end{table*}

\begin{table*}[t]
\centering
\caption{Directional flip rates and nondeterminism under paraphrasing (SimpleQA).
% The “original” answer refers to the model's response to the dataset's original wording of each question.
% We report how often paraphrasing causes a correct original answer to become incorrect 
% ($P(\neg A_{\text{para}} \mid A_{\text{orig}})$), or an originally incorrect answer to become correct 
% ($P(A_{\text{para}} \mid \neg A_{\text{orig}})$). 
% We also measure nondeterminism, defined as cases where the paraphrased versions do not agree on a single label.
% These quantities characterize the fragility and recoverability of model behavior under changes in wording.
$A_{\text{orig}}$ denotes standard accuracy on the original question wording, and 
$A_{\text{strict}}$ denotes reliable accuracy, where the model is correct across all paraphrases.
We report directional flip rates showing how often paraphrasing causes a correct original answer to become incorrect 
($P(\neg A_{\text{para}} \mid A_{\text{orig}})$) or an originally incorrect answer to become correct 
($P(A_{\text{para}} \mid \neg A_{\text{orig}})$).
We also report non-determinism (ND) rate, defined as the fraction of sets $\mathcal{Q}_i$ where the model produces at least two distinct labels (i.e., absolute disagreement), regardless of correctness.
}
\label{tab:simpleqa-flip-nd}
\resizebox{\textwidth}{!}{%
\begin{tabular}{lccccccc}
% \toprule
% model & $P(A_{\text{orig}})$ [Orig Acc] & $P(A_{\text{strict}})$ & $P(\neg A_{\text{para}} \mid A_{\text{orig}})$ & $P(A_{\text{para}} \mid \neg A_{\text{orig}})$ & $P(\text{ND} \mid A_{\text{orig}})$ & $P(\text{ND} \mid \neg A_{\text{orig}})$ & $P(\text{ND})$ \\
% \midrule
% GPT-3.5 & 9.1 & 2.1 & 41.5 & 1.6 & 78.7 & 45.8 & 48.8 \\
% GPT-4.1 & 38.3 & 25.0 & 11.3 & 7.2 & 33.9 & 42.1 & 38.9 \\
% GPT-4.1-Mini & 15.7 & 8.4 & 16.6 & 2.4 & 44.2 & 22.6 & 26.0 \\
% GPT-4o & 19.8 & 7.6 & 31.9 & 3.6 & 60.4 & 40.8 & 44.6 \\
% LLaMA-3 8B & 3.5 & 1.0 & 32.4 & 0.5 & 73.0 & 42.2 & 43.3 \\
% LLaMA-3.1 8B & 0.9 & 0.3 & 30.0 & 0.3 & 60.0 & 6.5 & 7.0 \\
% Qwen-2.5 7B & 0.4 & 0.1 & 40.0 & 0.0 & 80.0 & 6.7 & 7.0 \\
% Qwen-3 0.6B & 0.0 & 0.0 & 0.0 & 0.0 & 0.0 & 15.6 & 15.6 \\
% Qwen-3 0.6B +Thinking & 0.7 & 0.0 & 83.3 & 0.0 & 83.3 & 62.9 & 63.0 \\
% Qwen-3 1.7B & 1.1 & 0.3 & 72.7 & 0.3 & 54.5 & 61.1 & 61.0 \\
% Qwen-3 1.7B +Thinking & 1.1 & 0.4 & 55.6 & 0.1 & 44.4 & 42.9 & 43.0 \\
% Qwen-3 4B & 1.7 & 0.5 & 47.1 & 0.4 & 70.6 & 55.2 & 55.4 \\
% Qwen-3 4B +Thinking & 1.7 & 0.6 & 66.7 & 0.6 & 60.0 & 48.3 & 48.5 \\
% Qwen-3 8B & 3.0 & 0.5 & 61.3 & 0.4 & 71.0 & 44.6 & 45.4 \\
% Qwen-3 8B +Thinking & 2.3 & 0.6 & 36.8 & 0.5 & 57.9 & 46.3 & 46.6 \\
% Qwen-3 14B & 5.0 & 1.2 & 48.0 & 1.0 & 72.0 & 48.3 & 49.5 \\
% Qwen-3 14B +Thinking & 3.2 & 1.0 & 42.9 & 0.2 & 60.7 & 47.6 & 48.0 \\
% Qwen-3 32B & 4.2 & 1.6 & 32.6 & 0.7 & 62.8 & 48.3 & 48.9 \\
% Qwen-3 32B +Thinking & 3.8 & 0.8 & 40.6 & 0.6 & 68.8 & 51.2 & 51.9 \\
% \bottomrule

\toprule
model & $P(A_{\text{orig}})$ [Orig Acc] & $P(A_{\text{strict}})$ & $P(\neg A_{\text{para}} \mid A_{\text{orig}})$ & $P(A_{\text{para}} \mid \neg A_{\text{orig}})$ & $P(\text{ND} \mid A_{\text{orig}})$ & $P(\text{ND} \mid \neg A_{\text{orig}})$ \\
\midrule
GPT-3.5 & 9.1 & 2.1 & 41.5 & 1.6 & 78.7 & 45.8 \\
GPT-4.1 & 38.3 & 25.0 & 11.3 & 7.2 & 33.9 & 42.1 \\
GPT-4.1-Mini & 15.7 & 8.4 & 16.6 & 2.4 & 44.2 & 22.6 \\
GPT-4o & 19.8 & 7.6 & 31.9 & 3.6 & 60.4 & 40.8 \\
LLaMA-3 8B & 3.5 & 1.0 & 32.4 & 0.5 & 73.0 & 42.2 \\
LLaMA-3.1 8B & 0.9 & 0.3 & 30.0 & 0.3 & 60.0 & 6.5 \\
Qwen-2.5 7B & 0.4 & 0.1 & 40.0 & 0.0 & 80.0 & 6.7 \\
Qwen-3 0.6B & 0.0 & 0.0 & 0.0 & 0.0 & 0.0 & 15.6 \\
Qwen-3 0.6B +Thinking & 0.7 & 0.0 & 83.3 & 0.0 & 83.3 & 62.9 \\
Qwen-3 1.7B & 1.1 & 0.3 & 72.7 & 0.3 & 54.5 & 61.1 \\
Qwen-3 1.7B +Thinking & 1.1 & 0.4 & 55.6 & 0.1 & 44.4 & 42.9 \\
Qwen-3 4B & 1.7 & 0.5 & 47.1 & 0.4 & 70.6 & 55.2 \\
Qwen-3 4B +Thinking & 1.7 & 0.6 & 66.7 & 0.6 & 60.0 & 48.3 \\
Qwen-3 8B & 3.0 & 0.5 & 61.3 & 0.4 & 71.0 & 44.6 \\
Qwen-3 8B +Thinking & 2.3 & 0.6 & 36.8 & 0.5 & 57.9 & 46.3 \\
Qwen-3 14B & 5.0 & 1.2 & 48.0 & 1.0 & 72.0 & 48.3 \\
Qwen-3 14B +Thinking & 3.2 & 1.0 & 42.9 & 0.2 & 60.7 & 47.6 \\
Qwen-3 32B & 4.2 & 1.6 & 32.6 & 0.7 & 62.8 & 48.3 \\
Qwen-3 32B +Thinking & 3.8 & 0.8 & 40.6 & 0.6 & 68.8 & 51.2 \\
\bottomrule

% \midrule
% GPT-3.5 & 9.4 & 2.0 & 37.9 & 2.5 & 80.6 & 48.8 \\
% GPT-4.1 & 39.0 & 23.9 & 11.8 & 8.0 & 38.0 & 43.8 \\
% GPT-4.1-Mini & 16.0 & 8.1 & 16.2 & 4.3 & 47.4 & 25.1 \\
% GPT-4o & 20.9 & 7.2 & 30.0 & 4.7 & 64.3 & 42.7 \\
% LLaMA-3 8B & 3.9 & 0.9 & 30.2 & 1.1 & 76.7 & 44.7 \\
% LLaMA-3.1 8B & 1.1 & 0.3 & 25.0 & 0.4 & 66.7 & 6.9 \\
% Qwen-2.5 7B & 0.4 & 0.1 & 40.0 & 0.0 & 80.0 & 6.9 \\
% Qwen-3 0.6B & 0.0 & 0.0 & 0.0 & 0.0 & 0.0 & 16.2 \\
% Qwen-3 0.6B +Thinking & 0.8 & 0.0 & 75.0 & 0.1 & 87.5 & 66.6 \\
% Qwen-3 1.7B & 1.0 & 0.3 & 72.7 & 0.4 & 54.5 & 63.5 \\
% Qwen-3 1.7B +Thinking & 1.3 & 0.3 & 41.7 & 0.4 & 58.3 & 46.7 \\
% Qwen-3 4B & 2.0 & 0.5 & 42.9 & 0.9 & 76.2 & 57.5 \\
% Qwen-3 4B +Thinking & 2.0 & 0.5 & 57.9 & 0.8 & 68.4 & 52.2 \\
% Qwen-3 8B & 3.0 & 0.5 & 59.4 & 0.7 & 71.9 & 46.7 \\
% Qwen-3 8B +Thinking & 2.5 & 0.6 & 36.4 & 0.7 & 63.6 & 49.9 \\
% Qwen-3 14B & 5.2 & 1.1 & 45.5 & 1.5 & 74.5 & 50.9 \\
% Qwen-3 14B +Thinking & 3.2 & 1.0 & 40.0 & 0.6 & 63.3 & 50.7 \\
% Qwen-3 32B & 4.4 & 1.5 & 31.9 & 1.3 & 66.0 & 50.7 \\
% Qwen-3 32B +Thinking & 4.0 & 0.8 & 41.7 & 1.1 & 72.2 & 55.0 \\
% \bottomrule
\end{tabular}
}

\end{table*}

\begin{table*}[t]
\centering
\caption{Directional flip rates and nondeterminism under paraphrasing (TruthfulQA).
$A_{\text{orig}}$ denotes standard accuracy on the original question wording, and 
$A_{\text{strict}}$ denotes reliable accuracy, where the model is correct across all paraphrases.
We report directional flip rates showing how often paraphrasing causes a correct original answer to become incorrect 
($P(\neg A_{\text{para}} \mid A_{\text{orig}})$) or an originally incorrect answer to become correct 
($P(A_{\text{para}} \mid \neg A_{\text{orig}})$).
We also report non-determinism (ND) rate, defined as the fraction of sets $\mathcal{Q}_i$ where the model produces at least two distinct labels (i.e., absolute disagreement), regardless of correctness.
}
\label{tab:truthfulqa-flip-nd}
\resizebox{\textwidth}{!}{%
\begin{tabular}{lccccccc}
% \toprule
% model & $P(A_{\text{orig}})$ [Orig Acc] & $P(A_{\text{strict}})$ & $P(\neg A_{\text{para}} \mid A_{\text{orig}})$ & $P(A_{\text{para}} \mid \neg A_{\text{orig}})$ & $P(\text{ND} \mid A_{\text{orig}})$ & $P(\text{ND} \mid \neg A_{\text{orig}})$ & $P(\text{ND})$ \\
% \midrule
% GPT-3.5 & 52.3 & 29.8 & 13.2 & 15.5 & 44.3 & 47.5 & 45.8 \\
% GPT-4.1 & 76.3 & 58.5 & 5.8 & 13.5 & 22.7 & 54.2 & 30.1 \\
% GPT-4.1-Mini & 70.6 & 48.9 & 7.5 & 15.5 & 30.1 & 48.3 & 35.4 \\
% GPT-4o & 56.6 & 40.2 & 11.2 & 14.0 & 28.6 & 51.2 & 38.4 \\
% LLaMA-3 8B & 28.8 & 13.1 & 21.1 & 8.5 & 55.3 & 48.6 & 50.5 \\
% LLaMA-3.1 8B & 30.9 & 15.3 & 26.9 & 6.0 & 50.4 & 42.9 & 45.2 \\
% Qwen-2.5 7B & 44.8 & 25.7 & 16.4 & 10.9 & 41.5 & 45.5 & 43.7 \\
% Qwen-3 0.6B & 20.4 & 8.5 & 27.1 & 5.1 & 58.6 & 53.1 & 54.2 \\
% Qwen-3 0.6B +Thinking & 21.3 & 8.9 & 28.0 & 6.5 & 56.0 & 43.8 & 46.4 \\
% Qwen-3 1.7B & 16.9 & 7.0 & 33.8 & 6.9 & 52.3 & 49.7 & 50.1 \\
% Qwen-3 1.7B +Thinking & 53.1 & 34.1 & 11.7 & 14.2 & 36.7 & 38.7 & 37.6 \\
% Qwen-3 4B & 35.7 & 16.9 & 14.6 & 11.3 & 53.3 & 63.2 & 59.6 \\
% Qwen-3 4B +Thinking & 67.5 & 48.3 & 8.4 & 14.0 & 29.6 & 31.4 & 30.2 \\
% Qwen-3 8B & 47.4 & 31.9 & 9.4 & 14.9 & 36.5 & 50.2 & 43.7 \\
% Qwen-3 8B +Thinking & 72.9 & 61.5 & 3.7 & 11.3 & 16.8 & 39.4 & 22.9 \\
% Qwen-3 14B & 62.6 & 41.2 & 7.8 & 14.5 & 34.6 & 42.0 & 37.4 \\
% Qwen-3 14B +Thinking & 76.1 & 64.8 & 4.9 & 17.1 & 14.3 & 27.1 & 17.4 \\
% Qwen-3 32B & 65.2 & 44.1 & 10.7 & 11.9 & 31.6 & 40.7 & 34.8 \\
% Qwen-3 32B +Thinking & 81.8 & 69.8 & 4.4 & 20.0 & 14.2 & 50.0 & 20.7 \\
% \bottomrule

\toprule
model & $P(A_{\text{orig}})$ [Orig Acc] & $P(A_{\text{strict}})$ & $P(\neg A_{\text{para}} \mid A_{\text{orig}})$ & $P(A_{\text{para}} \mid \neg A_{\text{orig}})$ & $P(\text{ND} \mid A_{\text{orig}})$ & $P(\text{ND} \mid \neg A_{\text{orig}})$  \\
\midrule
GPT-3.5 & 52.3 & 29.8 & 13.2 & 15.5 & 44.3 & 47.5 \\
GPT-4.1 & 76.3 & 58.5 & 5.8 & 13.5 & 22.7 & 54.2 \\
GPT-4.1-Mini & 70.6 & 48.9 & 7.5 & 15.5 & 30.1 & 48.3 \\
GPT-4o & 56.6 & 40.2 & 11.2 & 14.0 & 28.6 & 51.2 \\
LLaMA-3 8B & 28.8 & 13.1 & 21.1 & 8.5 & 55.3 & 48.6 \\
LLaMA-3.1 8B & 30.9 & 15.3 & 26.9 & 6.0 & 50.4 & 42.9 \\
Qwen-2.5 7B & 44.8 & 25.7 & 16.4 & 10.9 & 41.5 & 45.5 \\
Qwen-3 0.6B & 20.4 & 8.5 & 27.1 & 5.1 & 58.6 & 53.1 \\
Qwen-3 0.6B +Thinking & 21.3 & 8.9 & 28.0 & 6.5 & 56.0 & 43.8 \\
Qwen-3 1.7B & 16.9 & 7.0 & 33.8 & 6.9 & 52.3 & 49.7 \\
Qwen-3 1.7B +Thinking & 53.1 & 34.1 & 11.7 & 14.2 & 36.7 & 38.7 \\
Qwen-3 4B & 35.7 & 16.9 & 14.6 & 11.3 & 53.3 & 63.2 \\
Qwen-3 4B +Thinking & 67.5 & 48.3 & 8.4 & 14.0 & 29.6 & 31.4 \\
Qwen-3 8B & 47.4 & 31.9 & 9.4 & 14.9 & 36.5 & 50.2 \\
Qwen-3 8B +Thinking & 72.9 & 61.5 & 3.7 & 11.3 & 16.8 & 39.4 \\
Qwen-3 14B & 62.6 & 41.2 & 7.8 & 14.5 & 34.6 & 42.0 \\
Qwen-3 14B +Thinking & 76.1 & 64.8 & 4.9 & 17.1 & 14.3 & 27.1 \\
Qwen-3 32B & 65.2 & 44.1 & 10.7 & 11.9 & 31.6 & 40.7 \\
Qwen-3 32B +Thinking & 81.8 & 69.8 & 4.4 & 20.0 & 14.2 & 50.0 \\
\bottomrule
\end{tabular}
}

\end{table*}

\begin{table*}[t]
\centering
\caption{Directional flip rates and nondeterminism under paraphrasing (GSM8K).
$A_{\text{orig}}$ denotes standard accuracy on the original question wording, and 
$A_{\text{strict}}$ denotes reliable accuracy, where the model is correct across all paraphrases.
We report directional flip rates showing how often paraphrasing causes a correct original answer to become incorrect 
($P(\neg A_{\text{para}} \mid A_{\text{orig}})$) or an originally incorrect answer to become correct 
($P(A_{\text{para}} \mid \neg A_{\text{orig}})$).
We also report non-determinism (ND) rate, defined as the fraction of sets $\mathcal{Q}_i$ where the model produces at least two distinct labels (i.e., absolute disagreement), regardless of correctness.
}
\label{tab:gsm8k-flip-nd}

\resizebox{\textwidth}{!}{%
\begin{tabular}{lccccccc}

\toprule
model & $P(A_{\text{orig}})$ [Orig Acc] & $P(A_{\text{strict}})$ & $P(\neg A_{\text{para}} \mid A_{\text{orig}})$ & $P(A_{\text{para}} \mid \neg A_{\text{orig}})$ & $P(\text{ND} \mid A_{\text{orig}})$ & $P(\text{ND} \mid \neg A_{\text{orig}})$ \\
\midrule
GPT-OSS 20B & 94.0 & 66.9 & 2.3 & 24.1 & 28.4 & 50.0 \\
GPT-3.5 & 80.6 & 43.4 & 6.9 & 36.7 & 46.9 & 68.3 \\
GPT-4.1 & 94.0 & 70.0 & 1.9 & 22.4 & 25.9 & 36.2 \\
GPT-4.1-Mini & 94.9 & 70.0 & 2.2 & 12.0 & 26.2 & 36.0 \\
GPT-4o & 94.0 & 68.8 & 2.0 & 17.2 & 27.0 & 44.8 \\
LLaMA-3 8B & 77.8 & 39.3 & 10.9 & 23.3 & 49.9 & 59.6 \\
LLaMA-3.1 8B & 84.9 & 52.2 & 6.9 & 31.7 & 38.7 & 64.7 \\
Qwen-2.5 7B & 91.0 & 59.6 & 4.6 & 21.8 & 34.4 & 55.2 \\
Qwen3-0.6B & 62.7 & 22.2 & 16.0 & 20.6 & 64.1 & 66.3 \\
Qwen3-0.6B +Thinking & 84.0 & 45.9 & 7.6 & 37.9 & 45.4 & 58.6 \\
Qwen3-1.7B & 83.1 & 47.1 & 7.9 & 28.7 & 43.6 & 63.1 \\
Qwen3-1.7B +Thinking & 90.5 & 58.2 & 4.3 & 34.4 & 35.9 & 62.2 \\
Qwen3-4B & 90.8 & 59.6 & 3.2 & 30.3 & 34.7 & 53.9 \\
Qwen3-4B +Thinking & 95.2 & 72.6 & 2.5 & 28.3 & 23.4 & 58.7 \\
Qwen-3 8B & 93.2 & 62.3 & 3.5 & 23.1 & 33.2 & 49.2 \\
Qwen3-8B +Thinking & 95.9 & 72.8 & 2.2 & 17.5 & 23.9 & 47.5 \\
Qwen3-14B & 94.6 & 68.1 & 2.5 & 17.0 & 28.0 & 41.5 \\
Qwen3-14B +Thinking & 95.5 & 74.2 & 2.5 & 15.9 & 22.2 & 45.5 \\
Qwen3-32B & 94.2 & 67.9 & 3.2 & 19.6 & 28.0 & 41.1 \\
Qwen3-32B +Thinking & 95.4 & 71.1 & 2.5 & 15.9 & 25.1 & 54.5 \\
\bottomrule
\end{tabular}
}

\end{table*}

\begin{table*}[t]
\centering
\caption{Directional flip rates and nondeterminism under paraphrasing (MATH-500).
$A_{\text{orig}}$ denotes standard accuracy on the original question wording, and 
$A_{\text{strict}}$ denotes reliable accuracy, where the model is correct across all paraphrases.
We report directional flip rates showing how often paraphrasing causes a correct original answer to become incorrect 
($P(\neg A_{\text{para}} \mid A_{\text{orig}})$) or an originally incorrect answer to become correct 
($P(A_{\text{para}} \mid \neg A_{\text{orig}})$).
We also report non-determinism (ND) rate, defined as the fraction of sets $\mathcal{Q}_i$ where the model produces at least two distinct labels (i.e., absolute disagreement), regardless of correctness.
}
\label{tab:math-500-flip-nd}
\resizebox{\textwidth}{!}{%
\begin{tabular}{lccccccc}
% \toprule
% model & $P(A_{\text{orig}})$ [Orig Acc] & $P(A_{\text{strict}})$ & $P(\neg A_{\text{para}} \mid A_{\text{orig}})$ & $P(A_{\text{para}} \mid \neg A_{\text{orig}})$ & $P(\text{ND} \mid A_{\text{orig}})$ & $P(\text{ND} \mid \neg A_{\text{orig}})$ & $P(\text{ND})$ \\
% \midrule
% GPT-OSS 20B & 96.3 & 87.4 & 3.8 & 20.0 & 6.9 & 20.0 & 7.4 \\
% GPT-3.5 & 44.8 & 26.2 & 21.5 & 13.8 & 40.0 & 30.0 & 34.5 \\
% GPT-4.1 & 94.4 & 84.6 & 4.4 & 12.5 & 8.1 & 37.5 & 9.8 \\
% GPT-4.1-Mini & 95.9 & 84.9 & 3.6 & 16.7 & 10.0 & 0.0 & 9.6 \\
% GPT-4o & 79.2 & 68.8 & 6.1 & 23.3 & 14.0 & 20.0 & 15.3 \\
% LLaMA-3 8B & 30.0 & 12.9 & 26.2 & 6.1 & 57.1 & 20.4 & 31.4 \\
% LLaMA-3.1 8B & 49.2 & 31.8 & 18.5 & 25.4 & 35.4 & 38.8 & 37.1 \\
% Qwen-2.5 7B & 75.0 & 59.0 & 5.6 & 13.9 & 19.4 & 22.2 & 20.1 \\
% Qwen-3 0.6B & 58.9 & 38.0 & 23.7 & 22.6 & 32.9 & 35.8 & 34.1 \\
% Qwen-3 0.6B +Thinking & 90.3 & 75.2 & 7.8 & 45.5 & 14.7 & 54.5 & 18.6 \\
% Qwen-3 1.7B & 80.7 & 52.6 & 10.1 & 19.2 & 32.1 & 26.9 & 31.1 \\
% Qwen-3 1.7B +Thinking & 98.4 & 91.3 & 1.6 & 0.0 & 5.6 & 50.0 & 6.3 \\
% Qwen-3 4B & 90.5 & 73.7 & 4.8 & 30.8 & 19.4 & 23.1 & 19.7 \\
% Qwen-3 4B +Thinking & 99.3 & 93.3 & 2.2 & 100.0 & 4.5 & 0.0 & 4.4 \\
% Qwen-3 8B & 81.1 & 75.5 & 4.3 & 25.9 & 9.5 & 25.9 & 12.6 \\
% Qwen-3 8B +Thinking & 100.0 & 94.2 & 2.2 & - & 3.6 & - & 3.6 \\
% Qwen-3 14B & 94.4 & 81.8 & 5.2 & 12.5 & 11.1 & 0.0 & 10.5 \\
% Qwen-3 14B +Thinking & 99.3 & 93.4 & 2.2 & 100.0 & 5.1 & 0.0 & 5.1 \\
% Qwen-3 32B & 92.5 & 76.7 & 4.4 & 27.3 & 16.3 & 9.1 & 15.8 \\
% Qwen-3 32B +Thinking & 100.0 & 93.6 & 2.1 & - & 4.3 & - & 4.3 \\
% \bottomrule

\toprule
model & $P(A_{\text{orig}})$ [Orig Acc] & $P(A_{\text{strict}})$ & $P(\neg A_{\text{para}} \mid A_{\text{orig}})$ & $P(A_{\text{para}} \mid \neg A_{\text{orig}})$ & $P(\text{ND} \mid A_{\text{orig}})$ & $P(\text{ND} \mid \neg A_{\text{orig}})$ \\
\midrule
GPT-OSS 20B & 96.3 & 87.4 & 3.8 & 20.0 & 6.9 & 20.0 \\
GPT-3.5 & 44.8 & 26.2 & 21.5 & 13.8 & 40.0 & 30.0 \\
GPT-4.1 & 94.4 & 84.6 & 4.4 & 12.5 & 8.1 & 37.5 \\
GPT-4.1-Mini & 95.9 & 84.9 & 3.6 & 16.7 & 10.0 & 0.0 \\
GPT-4o & 79.2 & 68.8 & 6.1 & 23.3 & 14.0 & 20.0 \\
LLaMA-3 8B & 30.0 & 12.9 & 26.2 & 6.1 & 57.1 & 20.4 \\
LLaMA-3.1 8B & 49.2 & 31.8 & 18.5 & 25.4 & 35.4 & 38.8 \\
Qwen-2.5 7B & 75.0 & 59.0 & 5.6 & 13.9 & 19.4 & 22.2 \\
Qwen3-0.6B & 58.9 & 38.0 & 23.7 & 22.6 & 32.9 & 35.8 \\
Qwen3-0.6B +Thinking & 90.3 & 75.2 & 7.8 & 45.5 & 14.7 & 54.5 \\
Qwen3-1.7B & 80.7 & 52.6 & 10.1 & 19.2 & 32.1 & 26.9 \\
Qwen3-1.7B +Thinking & 98.4 & 91.3 & 1.6 & 0.0 & 5.6 & 50.0 \\
Qwen3-4B & 90.5 & 73.7 & 4.8 & 30.8 & 19.4 & 23.1 \\
Qwen3-4B +Thinking & 99.3 & 93.3 & 2.2 & 100.0 & 4.5 & 0.0 \\
Qwen-3 8B & 81.1 & 75.5 & 4.3 & 25.9 & 9.5 & 25.9 \\
Qwen3-8B +Thinking & 100.0 & 94.2 & 2.2 & - & 3.6 & - \\
Qwen3-14B & 94.4 & 81.8 & 5.2 & 12.5 & 11.1 & 0.0 \\
Qwen3-14B +Thinking & 99.3 & 93.4 & 2.2 & 100.0 & 5.1 & 0.0 \\
Qwen3-32B & 92.5 & 76.7 & 4.4 & 27.3 & 16.3 & 9.1 \\
Qwen3-32B +Thinking & 100.0 & 93.6 & 2.1 & - & 4.3 & - \\
\bottomrule
\end{tabular}
}

\end{table*}

\begin{table*}[t]
\centering
\caption{Accuracy degradation, mismatch rates, and entropy-based instability under paraphrasing (SimpleQA).}
\label{tab:simpleqa-acc-mismatch-combined}
\resizebox{\textwidth}{!}{%
\begin{tabular}{lcccccccc}
\toprule
model & Orig Acc & Para Acc & $\Delta$Acc & Mismatch (\%) & IID-Mismatch & Entropy & Mode Share (\%) & N \\
\midrule
GPT-3.5 & 9.1 & 7.1 & \textbf{-2.0} & \textbf{21.8} & \textbf{0.19} & \textbf{0.15} & 86.1 & 1032 \\
GPT-4.1 & 38.3 & 37.4 & \textbf{-0.9} & \textbf{13.9} & \textbf{0.14} & \textbf{0.12} & 89.8 & 1040 \\
GPT-4.1-Mini & 15.7 & 15.2 & \textbf{-0.5} & 6.9 & 0.09 & 0.07 & 93.9 & 1039 \\
GPT-4o & 19.8 & 17.4 & \textbf{-2.4} & \textbf{19.1} & \textbf{0.16} & \textbf{0.13} & 88.3 & 1046 \\
LLaMA-3 8B & 3.5 & 3.1 & \textbf{-0.4} & \textbf{15.7} & \textbf{0.16} & \textbf{0.13} & 88.9 & 1046 \\
LLaMA-3.1 8B & 0.9 & 1.0 & 0.1 & 2.6 & 0.02 & 0.02 & 98.4 & 1115 \\
Qwen-2.5 7B & 0.4 & 0.5 & 0.0 & 1.8 & 0.03 & 0.02 & 98.2 & 1124 \\
Qwen-3 0.6B & 0.0 & 0.0 & 0.0 & 9.1 & 0.05 & 0.04 & 97.0 & 1106 \\
Qwen-3 0.6B +Thinking & 0.7 & 0.3 & \textbf{-0.4} & \textbf{28.4} & \textbf{0.23} & \textbf{0.2} & 82.9 & 865 \\
Qwen-3 1.7B & 1.1 & 0.9 & \textbf{-0.2} & \textbf{21.8} & \textbf{0.22} & \textbf{0.18} & 84.1 & 1026 \\
Qwen-3 1.7B +Thinking & 1.1 & 1.4 & 0.4 & \textbf{13.7} & \textbf{0.15} & \textbf{0.14} & 89.1 & 852 \\
Qwen-3 4B & 1.7 & 1.6 & \textbf{-0.1} & \textbf{20.3} & \textbf{0.2} & \textbf{0.17} & 85.9 & 1005 \\
Qwen-3 4B +Thinking & 1.7 & 1.6 & \textbf{-0.1} & \textbf{18.8} & \textbf{0.18} & \textbf{0.15} & 87.4 & 887 \\
Qwen-3 8B & 3.0 & 2.0 & \textbf{-0.9} & \textbf{16.3} & \textbf{0.17} & \textbf{0.14} & 88.0 & 1042 \\
Qwen-3 8B +Thinking & 2.3 & 2.1 & \textbf{-0.2} & \textbf{20.2} & \textbf{0.17} & \textbf{0.15} & 87.4 & 813 \\
Qwen-3 14B & 5.0 & 3.7 & \textbf{-1.2} & \textbf{16.3} & \textbf{0.18} & \textbf{0.15} & 87.1 & 1008 \\
Qwen-3 14B +Thinking & 3.2 & 2.6 & \textbf{-0.6} & \textbf{19.2} & \textbf{0.17} & \textbf{0.15} & 87.8 & 873 \\
Qwen-3 32B & 4.2 & 4.0 & \textbf{-0.2} & \textbf{15.0} & \textbf{0.17} & \textbf{0.14} & 87.9 & 1024 \\
Qwen-3 32B +Thinking & 3.8 & 3.3 & \textbf{-0.6} & \textbf{20.5} & \textbf{0.19} & \textbf{0.17} & 86.1 & 840 \\
\bottomrule
\end{tabular}
}

\end{table*}

\begin{table*}[t]
\centering
\caption{Accuracy degradation, mismatch rates, and entropy-based instability under paraphrasing (TruthfulQA).}
\label{tab:truthfulqa-acc-mismatch-combined}
\resizebox{\textwidth}{!}{%
\begin{tabular}{lcccccccc}
\toprule
model & Orig Acc & Para Acc & $\Delta$Acc & Mismatch (\%) & IID-Mismatch & Entropy & Mode Share (\%) & N \\
\midrule
GPT-3.5 & 52.3 & 51.0 & \textbf{-1.3} & \textbf{16.5} & \textbf{0.16} & \textbf{0.13} & 88.6 & 419 \\
GPT-4.1 & 76.3 & 73.7 & \textbf{-2.6} & 9.1 & \textbf{0.1} & 0.09 & 92.8 & 405 \\
GPT-4.1-Mini & 70.6 & 68.2 & \textbf{-2.4} & \textbf{11.1} & \textbf{0.13} & \textbf{0.11} & 90.9 & 395 \\
GPT-4o & 56.6 & 55.8 & \textbf{-0.8} & \textbf{16.4} & \textbf{0.14} & \textbf{0.11} & 89.6 & 396 \\
LLaMA-3 8B & 28.8 & 29.6 & 0.9 & \textbf{17.2} & \textbf{0.18} & \textbf{0.15} & 87.1 & 396 \\
LLaMA-3.1 8B & 30.9 & 28.8 & \textbf{-2.1} & \textbf{19.0} & \textbf{0.17} & \textbf{0.14} & 88.2 & 385 \\
Qwen-2.5 7B & 44.8 & 43.3 & \textbf{-1.5} & \textbf{17.5} & \textbf{0.16} & \textbf{0.13} & 88.6 & 382 \\
Qwen-3 0.6B & 20.4 & 21.3 & 0.9 & \textbf{21.9} & \textbf{0.21} & \textbf{0.18} & 84.8 & 343 \\
Qwen-3 0.6B +Thinking & 21.3 & 20.5 & \textbf{-0.7} & \textbf{20.0} & \textbf{0.18} & \textbf{0.17} & 87.3 & 235 \\
Qwen-3 1.7B & 16.9 & 18.3 & 1.4 & \textbf{23.1} & \textbf{0.19} & \textbf{0.16} & 86.1 & 385 \\
Qwen-3 1.7B +Thinking & 53.1 & 52.0 & \textbf{-1.1} & \textbf{17.7} & \textbf{0.13} & \textbf{0.12} & 90.8 & 226 \\
Qwen-3 4B & 35.7 & 38.6 & 2.9 & \textbf{19.0} & \textbf{0.23} & \textbf{0.2} & 83.1 & 384 \\
Qwen-3 4B +Thinking & 67.5 & 64.3 & \textbf{-3.2} & \textbf{10.9} & \textbf{0.11} & \textbf{0.1} & 92.5 & 265 \\
Qwen-3 8B & 47.4 & 49.9 & 2.5 & \textbf{15.7} & \textbf{0.15} & \textbf{0.12} & 89.4 & 382 \\
Qwen-3 8B +Thinking & 72.9 & 72.3 & \textbf{-0.6} & 8.0 & 0.08 & 0.07 & 94.5 & 262 \\
Qwen-3 14B & 62.6 & 60.8 & \textbf{-1.8} & \textbf{11.7} & \textbf{0.13} & \textbf{0.11} & 90.9 & 369 \\
Qwen-3 14B +Thinking & 76.1 & 74.3 & \textbf{-1.8} & 9.9 & 0.07 & 0.06 & 95.3 & 293 \\
Qwen-3 32B & 65.2 & 61.3 & \textbf{-3.9} & \textbf{13.7} & \textbf{0.13} & \textbf{0.1} & 91.2 & 388 \\
Qwen-3 32B +Thinking & 81.8 & 80.3 & \textbf{-1.6} & 9.5 & 0.08 & 0.08 & 94.4 & 275 \\
\bottomrule
\end{tabular}
}

\end{table*}

\begin{table*}[t]
\centering
\caption{Accuracy degradation, mismatch rates, and entropy-based stability under paraphrasing (GSM8K). }
\label{tab:gsm8k-acc-mismatch}
\resizebox{\textwidth}{!}{%
\begin{tabular}{lcccccccc}
\toprule
model & Orig Acc & Para Acc & $\Delta$Acc & Mismatch (\%) & IID-Mismatch & Entropy & Mode Share (\%) & N \\
\midrule
GPT-OSS 20B & 94.0 & 89.1 & \textcolor{red}{\textbf{-4.9}} & 3.6 & 0.09 & 0.07 & 94.1 & 965 \\
GPT-3.5 & 80.6 & 77.3 & \textcolor{red}{\textbf{-3.3}} & \textbf{12.7} & \textbf{0.16} & \textbf{0.12} & 88.6 & 929 \\
GPT-4.1 & 94.0 & 89.7 & \textcolor{red}{\textbf{-4.3}} & 3.1 & 0.08 & 0.06 & 94.7 & 970 \\
GPT-4.1-Mini & 94.9 & 89.6 & \textcolor{red}{\textbf{-5.3}} & 2.7 & 0.08 & 0.06 & 94.4 & 979 \\
GPT-4o & 94.0 & 89.2 & \textcolor{red}{\textbf{-4.9}} & 2.9 & 0.09 & 0.06 & 94.1 & 973 \\
LLaMA-3 8B & 77.8 & 71.4 & \textcolor{red}{\textbf{-6.4}} & \textbf{13.7} & \textbf{0.17} & \textbf{0.13} & 87.6 & 870 \\
LLaMA-3.1 8B & 84.9 & 80.0 & \textcolor{red}{\textbf{-4.9}} & \textbf{10.6} & \textbf{0.14} & \textbf{0.11} & 89.8 & 921 \\
Qwen-2.5 7B & 91.0 & 84.8 & \textcolor{red}{\textbf{-6.2}} & 6.1 & \textbf{0.11} & 0.08 & 92.1 & 963 \\
Qwen-3 0.6B & 62.7 & 58.3 & \textcolor{red}{\textbf{-4.4}} & \textbf{17.7} & \textbf{0.23} & \textbf{0.16} & 84.0 & 938 \\
Qwen-3 0.6B +Thinking & 84.0 & 78.4 & \textcolor{red}{\textbf{-5.6}} & \textbf{12.5} & \textbf{0.16} & \textbf{0.12} & 88.9 & 873 \\
Qwen-3 1.7B & 83.1 & 77.4 & \textcolor{red}{\textbf{-5.7}} & \textbf{11.4} & \textbf{0.16} & \textbf{0.11} & 88.8 & 947 \\
Qwen-3 1.7B +Thinking & 90.5 & 85.4 & \textcolor{red}{\textbf{-5.1}} & 7.2 & \textbf{0.12} & 0.09 & 91.7 & 950 \\
Qwen-3 4B & 90.8 & 85.6 & \textcolor{red}{\textbf{-5.2}} & 5.7 & \textbf{0.12} & 0.09 & 92.0 & 970 \\
Qwen-3 4B +Thinking & 95.2 & 90.9 & \textcolor{red}{\textbf{-4.3}} & 3.8 & 0.08 & 0.06 & 94.8 & 956 \\
Qwen-3 8B & 93.2 & 87.0 & \textcolor{red}{\textbf{-6.3}} & 4.8 & \textbf{0.11} & 0.08 & 92.7 & 962 \\
Qwen-3 8B +Thinking & 95.9 & 91.0 & \textcolor{red}{\textbf{-4.8}} & 2.8 & 0.08 & 0.06 & 94.9 & 965 \\
Qwen-3 14B & 94.6 & 89.0 & \textcolor{red}{\textbf{-5.6}} & 3.3 & 0.09 & 0.07 & 93.7 & 973 \\
Qwen-3 14B +Thinking & 95.5 & 91.1 & \textcolor{red}{\textbf{-4.4}} & 3.1 & 0.07 & 0.05 & 95.0 & 970 \\
Qwen-3 32B & 94.2 & 88.9 & \textcolor{red}{\textbf{-5.4}} & 4.1 & 0.09 & 0.06 & 93.9 & 970 \\
Qwen-3 32B +Thinking & 95.4 & 90.5 & \textcolor{red}{\textbf{-4.9}} & 3.1 & 0.08 & 0.06 & 94.5 & 967 \\
\bottomrule
\end{tabular}
}

\end{table*}

\begin{table*}[t]
\centering
\caption{Accuracy degradation, mismatch rates, and entropy-based stability under paraphrasing (MATH-500). }
\label{tab:math-500-acc-mismatch}

\resizebox{\textwidth}{!}{%
\begin{tabular}{lcccccccc}
\toprule
model & Orig Acc & Para Acc & $\Delta$Acc & Mismatch (\%) & IID-Mismatch & Entropy & Mode Share (\%) & N \\
\midrule
GPT-OSS 20B & 96.3 & 92.6 & \textcolor{red}{\textbf{-3.7}} & 4.4 & 0.03 & 0.03 & 98.3 & 135 \\
GPT-3.5 & 44.8 & 43.3 & \textcolor{red}{\textbf{-1.6}} & \textbf{17.2} & \textbf{0.13} & \textbf{0.11} & 91.1 & 145 \\
GPT-4.1 & 94.4 & 90.5 & \textcolor{red}{\textbf{-3.9}} & 4.9 & 0.03 & 0.03 & 97.6 & 143 \\
GPT-4.1-Mini & 95.9 & 91.5 & \textcolor{red}{\textbf{-4.4}} & 4.1 & 0.03 & 0.03 & 97.6 & 146 \\
GPT-4o & 79.2 & 78.3 & \textcolor{red}{\textbf{-0.9}} & 9.7 & 0.05 & 0.04 & 96.2 & 144 \\
LLaMA-3 8B & 30.0 & 26.6 & \textcolor{red}{\textbf{-3.4}} & \textbf{12.1} & \textbf{0.11} & \textbf{0.1} & 91.7 & 140 \\
LLaMA-3.1 8B & 49.2 & 52.4 & 3.2 & \textbf{22.0} & \textbf{0.15} & \textbf{0.14} & 89.4 & 132 \\
Qwen-2.5 7B & 75.0 & 71.5 & \textcolor{red}{\textbf{-3.5}} & 7.6 & 0.08 & 0.06 & 94.7 & 144 \\
Qwen-3 0.6B & 58.9 & 55.4 & \textcolor{red}{\textbf{-3.5}} & \textbf{23.3} & \textbf{0.13} & \textbf{0.12} & 90.6 & 129 \\
Qwen-3 0.6B +Thinking & 90.3 & 86.7 & \textcolor{red}{\textbf{-3.6}} & \textbf{11.5} & 0.06 & 0.06 & 95.7 & 113 \\
Qwen-3 1.7B & 80.7 & 73.5 & \textcolor{red}{\textbf{-7.3}} & \textbf{11.9} & \textbf{0.11} & 0.09 & 92.6 & 135 \\
Qwen-3 1.7B +Thinking & 98.4 & 95.9 & \textcolor{red}{\textbf{-2.6}} & 1.6 & 0.02 & 0.02 & 98.5 & 127 \\
Qwen-3 4B & 90.5 & 87.0 & \textcolor{red}{\textbf{-3.5}} & 7.3 & 0.07 & 0.06 & 95.1 & 137 \\
Qwen-3 4B +Thinking & 99.3 & 97.0 & \textcolor{red}{\textbf{-2.2}} & 3.0 & 0.01 & 0.01 & 99.3 & 135 \\
Qwen-3 8B & 81.1 & 82.3 & 1.1 & 8.4 & 0.04 & 0.03 & 97.4 & 143 \\
Qwen-3 8B +Thinking & 100.0 & 97.2 & \textcolor{red}{\textbf{-2.8}} & 2.2 & 0.01 & 0.01 & 99.4 & 137 \\
Qwen-3 14B & 94.4 & 89.3 & \textcolor{red}{\textbf{-5.1}} & 5.6 & 0.03 & 0.03 & 97.7 & 143 \\
Qwen-3 14B +Thinking & 99.3 & 97.4 & \textcolor{red}{\textbf{-1.9}} & 2.9 & 0.01 & 0.01 & 99.1 & 137 \\
Qwen-3 32B & 92.5 & 87.7 & \textcolor{red}{\textbf{-4.8}} & 6.2 & 0.05 & 0.05 & 96.5 & 146 \\
Qwen-3 32B +Thinking & 100.0 & 97.1 & \textcolor{red}{\textbf{-2.9}} & 2.1 & 0.01 & 0.01 & 99.2 & 141 \\
\bottomrule
\end{tabular}
}

\end{table*}

\begin{table*}[t]
\centering
\caption{Value-level variability under paraphrasing (GSM8K). 
We report how often paraphrased variants produce \emph{different numerical values} even when their correctness labels agree. 
These metrics quantify stochastic variation that is invisible to binary correctness measures, revealing instability in the model’s underlying reasoning rather than just its accuracy.}
\label{tab:gsm8k-anslevel-variability}
\resizebox{\textwidth}{!}{%
\begin{tabular}{lccccccc}
\toprule
model & Orig Acc & Para Acc & $\Delta$Acc & IID-Mismatch (Value) & Entropy (Value) & Mode Share (Value) & N \\
\midrule
GPT-OSS 20B & 94.0 & 89.1 & \textcolor{red}{\textbf{-4.9}} & \textbf{0.11} & 0.09 & 92.9 & 965 \\
GPT-3.5 & 80.6 & 77.3 & \textcolor{red}{\textbf{-3.3}} & \textbf{0.24} & \textbf{0.21} & 82.2 & 929 \\
GPT-4.1 & 94.0 & 89.7 & \textcolor{red}{\textbf{-4.3}} & \textbf{0.1} & 0.08 & 93.6 & 970 \\
GPT-4.1-Mini & 94.9 & 89.6 & \textcolor{red}{\textbf{-5.3}} & \textbf{0.1} & 0.08 & 93.4 & 979 \\
GPT-4o & 94.0 & 89.2 & \textcolor{red}{\textbf{-4.9}} & \textbf{0.11} & 0.09 & 92.9 & 973 \\
LLaMA-3 8B & 77.8 & 71.4 & \textcolor{red}{\textbf{-6.4}} & \textbf{0.29} & \textbf{0.26} & 78.3 & 870 \\
LLaMA-3.1 8B & 84.9 & 80.0 & \textcolor{red}{\textbf{-4.9}} & \textbf{0.2} & \textbf{0.19} & 85.1 & 921 \\
Qwen-2.5 7B & 91.0 & 84.8 & \textcolor{red}{\textbf{-6.2}} & \textbf{0.15} & \textbf{0.13} & 89.1 & 963 \\
Qwen-3 0.6B & 62.7 & 58.3 & \textcolor{red}{\textbf{-4.4}} & \textbf{0.42} & \textbf{0.39} & 66.7 & 938 \\
Qwen-3 0.6B +Thinking & 84.0 & 78.4 & \textcolor{red}{\textbf{-5.6}} & \textbf{0.22} & \textbf{0.2} & 84.0 & 873 \\
Qwen-3 1.7B & 83.1 & 77.4 & \textcolor{red}{\textbf{-5.7}} & \textbf{0.23} & \textbf{0.2} & 83.0 & 947 \\
Qwen-3 1.7B +Thinking & 90.5 & 85.4 & \textcolor{red}{\textbf{-5.1}} & \textbf{0.15} & \textbf{0.13} & 89.5 & 950 \\
Qwen-3 4B & 90.8 & 85.6 & \textcolor{red}{\textbf{-5.2}} & \textbf{0.15} & \textbf{0.12} & 89.7 & 970 \\
Qwen-3 4B +Thinking & 95.2 & 90.9 & \textcolor{red}{\textbf{-4.3}} & 0.09 & 0.07 & 93.9 & 956 \\
Qwen-3 8B & 93.2 & 87.0 & \textcolor{red}{\textbf{-6.3}} & \textbf{0.13} & \textbf{0.11} & 90.9 & 962 \\
Qwen-3 8B +Thinking & 95.9 & 91.0 & \textcolor{red}{\textbf{-4.8}} & 0.09 & 0.07 & 94.2 & 965 \\
Qwen-3 14B & 94.6 & 89.0 & \textcolor{red}{\textbf{-5.6}} & \textbf{0.11} & 0.09 & 92.4 & 973 \\
Qwen-3 14B +Thinking & 95.5 & 91.1 & \textcolor{red}{\textbf{-4.4}} & 0.08 & 0.07 & 94.4 & 970 \\
Qwen-3 32B & 94.2 & 88.9 & \textcolor{red}{\textbf{-5.4}} & \textbf{0.11} & 0.09 & 92.6 & 970 \\
Qwen-3 32B +Thinking & 95.4 & 90.5 & \textcolor{red}{\textbf{-4.9}} & 0.09 & 0.08 & 93.8 & 967 \\
\bottomrule
\end{tabular}
}

\end{table*}

\begin{table*}[t]
\centering
\caption{Value-level variability under paraphrasing (MATH-500). 
We report how often paraphrased variants produce \emph{different numerical values} even when their correctness labels agree. 
These metrics quantify stochastic variation that is invisible to binary correctness measures, revealing instability in the model’s underlying reasoning rather than just its accuracy.}
\label{tab:math-500-anslevel-variability}
\resizebox{\textwidth}{!}{%
\begin{tabular}{lccccccc}
\toprule
model & Orig Acc & Para Acc & $\Delta$Acc & IID-Mismatch (Value) & Entropy (Value) & Mode Share (Value) & N \\
\midrule
GPT-OSS 20B & 96.3 & 92.6 & \textcolor{red}{\textbf{-3.7}} & 0.04 & 0.05 & 97.0 & 135 \\
GPT-3.5 & 44.8 & 43.3 & \textcolor{red}{\textbf{-1.6}} & \textbf{0.35} & \textbf{0.42} & 69.3 & 145 \\
GPT-4.1 & 94.4 & 90.5 & \textcolor{red}{\textbf{-3.9}} & 0.05 & 0.05 & 96.2 & 143 \\
GPT-4.1-Mini & 95.9 & 91.5 & \textcolor{red}{\textbf{-4.4}} & 0.06 & 0.06 & 95.2 & 146 \\
GPT-4o & 79.2 & 78.3 & \textcolor{red}{\textbf{-0.9}} & \textbf{0.12} & \textbf{0.13} & 90.5 & 144 \\
LLaMA-3 8B & 30.0 & 26.6 & \textcolor{red}{\textbf{-3.4}} & \textbf{0.45} & \textbf{0.56} & 60.0 & 140 \\
LLaMA-3.1 8B & 49.2 & 52.4 & 3.2 & \textbf{0.3} & \textbf{0.38} & 74.0 & 132 \\
Qwen-2.5 7B & 75.0 & 71.5 & \textcolor{red}{\textbf{-3.5}} & \textbf{0.17} & \textbf{0.19} & 86.8 & 144 \\
Qwen-3 0.6B & 58.9 & 55.4 & \textcolor{red}{\textbf{-3.5}} & \textbf{0.31} & \textbf{0.36} & 73.8 & 129 \\
Qwen-3 0.6B +Thinking & 90.3 & 86.7 & \textcolor{red}{\textbf{-3.6}} & \textbf{0.1} & \textbf{0.11} & 92.3 & 113 \\
Qwen-3 1.7B & 80.7 & 73.5 & \textcolor{red}{\textbf{-7.3}} & \textbf{0.19} & \textbf{0.21} & 84.8 & 135 \\
Qwen-3 1.7B +Thinking & 98.4 & 95.9 & \textcolor{red}{\textbf{-2.6}} & 0.03 & 0.02 & 98.0 & 127 \\
Qwen-3 4B & 90.5 & 87.0 & \textcolor{red}{\textbf{-3.5}} & \textbf{0.11} & \textbf{0.11} & 91.8 & 137 \\
Qwen-3 4B +Thinking & 99.3 & 97.0 & \textcolor{red}{\textbf{-2.2}} & 0.02 & 0.02 & 98.5 & 135 \\
Qwen-3 8B & 81.1 & 82.3 & 1.1 & \textbf{0.11} & \textbf{0.12} & 91.5 & 143 \\
Qwen-3 8B +Thinking & 100.0 & 97.2 & \textcolor{red}{\textbf{-2.8}} & 0.02 & 0.03 & 98.4 & 137 \\
Qwen-3 14B & 94.4 & 89.3 & \textcolor{red}{\textbf{-5.1}} & 0.06 & 0.07 & 95.3 & 143 \\
Qwen-3 14B +Thinking & 99.3 & 97.4 & \textcolor{red}{\textbf{-1.9}} & 0.03 & 0.03 & 98.1 & 137 \\
Qwen-3 32B & 92.5 & 87.7 & \textcolor{red}{\textbf{-4.8}} & 0.08 & 0.09 & 93.7 & 146 \\
Qwen-3 32B +Thinking & 100.0 & 97.1 & \textcolor{red}{\textbf{-2.9}} & 0.02 & 0.03 & 98.2 & 141 \\
\bottomrule
\end{tabular}
}

\end{table*}

\begin{table*}[t]
\centering
\caption{
Partial-correctness (“any-correct”) statistics under paraphrasing (SimpleQA).
$A_{\text{orig}}$ denotes accuracy on the original question wording, 
$A_{\text{para}}$ the average accuracy across paraphrases,
$A_{\text{any}}$ the best-case accuracy where at least one paraphrase is correct, 
and $A_{\text{strict}}$ the reliable accuracy where all paraphrases are correct.
$P(A_{\text{any}} \mid \neg A_{\text{orig}})$ measures cases where knowledge is revealed only under paraphrasing.
% Here $P(A_{\text{any}})$ reflects cases where the model produces a correct answer for at least one 
% paraphrase, revealing knowledge that exists internally but is not reliably accessed across different 
% wordings.
% \textbf{Columns.}
% $P(A_{\text{orig}})$: accuracy on original wording.
% $P(A_{\text{any}})$: how often at least one paraphrase is correct.
% $P(A_{\text{any}} \mid \neg A_{\text{orig}})$: knowledge only revealed under paraphrasing.
% $P(A_{\text{any}} \mid \neg A_{\text{para}})$: correct answer appears even when the paraphrase majority is wrong.
}
\label{tab:simpleqa-anycorrect}
\begin{tabular}{lccccc}
\toprule
model & $P(A_{\text{orig}})$ & $P(A_{\text{para}})$ & $P(A_{\text{any}})$ & $P(A_{\text{strict}})$ & $P(A_{\text{any}} \mid \neg A_{\text{orig}})$ \\
\midrule
GPT-3.5 & 9.1 & 7.1 & 14.8 & 2.1 & 7.5 \\
GPT-4.1 & 38.3 & 37.4 & 49.2 & 25.0 & 20.4 \\
GPT-4.1-Mini & 15.7 & 15.2 & 22.5 & 8.4 & 8.8 \\
GPT-4o & 19.8 & 17.4 & 31.0 & 7.6 & 16.3 \\
LLaMA-3 8B & 3.5 & 3.1 & 6.9 & 1.0 & 3.8 \\
LLaMA-3.1 8B & 0.9 & 1.0 & 2.5 & 0.3 & 1.7 \\
Qwen-2.5 7B & 0.4 & 0.5 & 1.5 & 0.1 & 1.1 \\
Qwen-3 0.6B & 0.0 & 0.0 & 0.3 & 0.0 & 0.3 \\
Qwen-3 0.6B +Thinking & 0.7 & 0.3 & 1.2 & 0.0 & 0.9 \\
Qwen-3 1.7B & 1.1 & 0.9 & 2.4 & 0.3 & 1.9 \\
Qwen-3 1.7B +Thinking & 1.1 & 1.4 & 4.7 & 0.4 & 4.0 \\
Qwen-3 4B & 1.7 & 1.6 & 4.3 & 0.5 & 3.2 \\
Qwen-3 4B +Thinking & 1.7 & 1.6 & 4.2 & 0.6 & 3.1 \\
Qwen-3 8B & 3.0 & 2.0 & 5.4 & 0.5 & 3.4 \\
Qwen-3 8B +Thinking & 2.3 & 2.1 & 4.7 & 0.6 & 2.9 \\
Qwen-3 14B & 5.0 & 3.7 & 7.8 & 1.2 & 3.5 \\
Qwen-3 14B +Thinking & 3.2 & 2.6 & 5.8 & 1.0 & 3.3 \\
Qwen-3 32B & 4.2 & 4.0 & 8.7 & 1.6 & 5.0 \\
Qwen-3 32B +Thinking & 3.8 & 3.3 & 7.1 & 0.8 & 4.6 \\
\bottomrule
\end{tabular}

\end{table*}

\begin{table*}[t]
\centering
\caption{
Partial-correctness (“any-correct”) statistics under paraphrasing (TruthfulQA). 
$A_{\text{orig}}$ denotes accuracy on the original question wording, 
$A_{\text{para}}$ the average accuracy across paraphrases,
$A_{\text{any}}$ the best-case accuracy where at least one paraphrase is correct, 
and $A_{\text{strict}}$ the reliable accuracy where all paraphrases are correct.
$P(A_{\text{any}} \mid \neg A_{\text{orig}})$ measures cases where knowledge is revealed only under paraphrasing.
}
\label{tab:truthfulqa-anycorrect}
\begin{tabular}{lccccc}
\toprule
model & $P(A_{\text{orig}})$ & $P(A_{\text{para}})$ & $P(A_{\text{any}})$ & $P(A_{\text{strict}})$ & $P(A_{\text{any}} \mid \neg A_{\text{orig}})$ \\
\midrule
GPT-3.5 & 52.3 & 51.0 & 69.0 & 29.8 & 40.0 \\
GPT-4.1 & 76.3 & 73.7 & 85.7 & 58.5 & 44.8 \\
GPT-4.1-Mini & 70.6 & 68.2 & 82.0 & 48.9 & 43.1 \\
GPT-4o & 56.6 & 55.8 & 72.2 & 40.2 & 39.5 \\
LLaMA-3 8B & 28.8 & 29.6 & 51.0 & 13.1 & 33.0 \\
LLaMA-3.1 8B & 30.9 & 28.8 & 47.8 & 15.3 & 26.7 \\
Qwen-2.5 7B & 44.8 & 43.3 & 60.5 & 25.7 & 31.8 \\
Qwen-3 0.6B & 20.4 & 21.3 & 43.1 & 8.5 & 30.4 \\
Qwen-3 0.6B +Thinking & 21.3 & 20.5 & 35.3 & 8.9 & 21.6 \\
Qwen-3 1.7B & 16.9 & 18.3 & 37.4 & 7.0 & 27.2 \\
Qwen-3 1.7B +Thinking & 53.1 & 52.0 & 65.0 & 34.1 & 32.1 \\
Qwen-3 4B & 35.7 & 38.6 & 63.0 & 16.9 & 44.1 \\
Qwen-3 4B +Thinking & 67.5 & 64.3 & 74.7 & 48.3 & 26.7 \\
Qwen-3 8B & 47.4 & 49.9 & 68.3 & 31.9 & 41.3 \\
Qwen-3 8B +Thinking & 72.9 & 72.3 & 79.8 & 61.5 & 29.6 \\
Qwen-3 14B & 62.6 & 60.8 & 74.3 & 41.2 & 37.0 \\
Qwen-3 14B +Thinking & 76.1 & 74.3 & 78.8 & 64.8 & 22.9 \\
Qwen-3 32B & 65.2 & 61.3 & 75.5 & 44.1 & 36.3 \\
Qwen-3 32B +Thinking & 81.8 & 80.3 & 87.6 & 69.8 & 38.0 \\
\bottomrule
\end{tabular}

\end{table*}

\begin{table*}[t]
\centering

\caption{
Partial-correctness (“any-correct”) statistics under paraphrasing (GSM8K).
$A_{\text{orig}}$ denotes accuracy on the original question wording, 
$A_{\text{para}}$ the average accuracy across paraphrases,
$A_{\text{any}}$ the best-case accuracy where at least one paraphrase is correct, 
and $A_{\text{strict}}$ the reliable accuracy where all paraphrases are correct.
$P(A_{\text{any}} \mid \neg A_{\text{orig}})$ measures cases where knowledge is revealed only under paraphrasing.
}
\label{tab:gsm8k-anycorrect}
\begin{tabular}{lccccc}
\toprule
model & $P(A_{\text{orig}})$ & $P(A_{\text{para}})$ & $P(A_{\text{any}})$ & $P(A_{\text{strict}})$ & $P(A_{\text{any}} \mid \neg A_{\text{orig}})$ \\
\midrule
GPT-OSS 20B & 94.0 & 89.1 & 96.7 & 86.7 & 53.4 \\
GPT-3.5 & 80.6 & 77.3 & 94.4 & 70.6 & 74.4 \\
GPT-4.1 & 94.0 & 89.7 & 96.5 & 87.2 & 48.3 \\
GPT-4.1-Mini & 94.9 & 89.6 & 96.6 & 86.7 & 38.0 \\
GPT-4o & 94.0 & 89.2 & 96.8 & 86.0 & 50.0 \\
LLaMA-3 8B & 77.8 & 71.4 & 91.4 & 62.6 & 64.2 \\
LLaMA-3.1 8B & 84.9 & 80.0 & 94.9 & 73.2 & 71.2 \\
Qwen-2.5 7B & 91.0 & 84.8 & 95.8 & 80.3 & 59.8 \\
Qwen-3 0.6B & 62.7 & 58.3 & 87.1 & 46.1 & 69.4 \\
Qwen-3 0.6B +Thinking & 84.0 & 78.4 & 93.5 & 71.1 & 67.9 \\
Qwen-3 1.7B & 83.1 & 77.4 & 94.0 & 69.1 & 69.4 \\
Qwen-3 1.7B +Thinking & 90.5 & 85.4 & 96.6 & 80.9 & 70.0 \\
Qwen-3 4B & 90.8 & 85.6 & 96.1 & 81.8 & 59.6 \\
Qwen-3 4B +Thinking & 95.2 & 90.9 & 97.7 & 89.1 & 60.9 \\
Qwen-3 8B & 93.2 & 87.0 & 96.6 & 83.5 & 55.4 \\
Qwen-3 8B +Thinking & 95.9 & 91.0 & 97.7 & 89.1 & 52.5 \\
Qwen-3 14B & 94.6 & 89.0 & 96.9 & 85.5 & 47.2 \\
Qwen-3 14B +Thinking & 95.5 & 91.1 & 97.5 & 88.9 & 50.0 \\
Qwen-3 32B & 94.2 & 88.9 & 96.7 & 86.0 & 46.4 \\
Qwen-3 32B +Thinking & 95.4 & 90.5 & 97.6 & 88.5 & 59.1 \\
\bottomrule
\end{tabular}

\end{table*}

\begin{table*}[t]
\centering
\caption{
Partial-correctness (“any-correct”) statistics under paraphrasing (MATH-500).
$A_{\text{orig}}$ denotes accuracy on the original question wording, 
$A_{\text{para}}$ the average accuracy across paraphrases,
$A_{\text{any}}$ the best-case accuracy where at least one paraphrase is correct, 
and $A_{\text{strict}}$ the reliable accuracy where all paraphrases are correct.
$P(A_{\text{any}} \mid \neg A_{\text{orig}})$ measures cases where knowledge is revealed only under paraphrasing.
}
\label{tab:math-500-anycorrect}
\begin{tabular}{lccccc}
\toprule
model & $P(A_{\text{orig}})$ & $P(A_{\text{para}})$ & $P(A_{\text{any}})$ & $P(A_{\text{strict}})$ & $P(A_{\text{any}} \mid \neg A_{\text{orig}})$ \\
\midrule
GPT-OSS 20B & 96.3 & 92.6 & 97.8 & 91.9 & 40.0 \\
GPT-3.5 & 44.8 & 43.3 & 64.1 & 35.9 & 35.0 \\
GPT-4.1 & 94.4 & 90.5 & 96.5 & 88.1 & 37.5 \\
GPT-4.1-Mini & 95.9 & 91.5 & 96.6 & 89.7 & 16.7 \\
GPT-4o & 79.2 & 78.3 & 86.1 & 75.7 & 33.3 \\
LLaMA-3 8B & 30.0 & 26.6 & 45.7 & 17.9 & 22.4 \\
LLaMA-3.1 8B & 49.2 & 52.4 & 72.0 & 42.4 & 44.8 \\
Qwen-2.5 7B & 75.0 & 71.5 & 81.2 & 68.1 & 25.0 \\
Qwen-3 0.6B & 58.9 & 55.4 & 76.7 & 47.3 & 43.4 \\
Qwen-3 0.6B +Thinking & 90.3 & 86.7 & 98.2 & 85.0 & 81.8 \\
Qwen-3 1.7B & 80.7 & 73.5 & 88.1 & 69.6 & 38.5 \\
Qwen-3 1.7B +Thinking & 98.4 & 95.9 & 99.2 & 95.3 & 50.0 \\
Qwen-3 4B & 90.5 & 87.0 & 94.9 & 83.2 & 46.2 \\
Qwen-3 4B +Thinking & 99.3 & 97.0 & 100.0 & 97.0 & 100.0 \\
Qwen-3 8B & 81.1 & 82.3 & 89.5 & 81.1 & 44.4 \\
Qwen-3 8B +Thinking & 100.0 & 97.2 & 100.0 & 97.1 & - \\
Qwen-3 14B & 94.4 & 89.3 & 95.1 & 88.8 & 12.5 \\
Qwen-3 14B +Thinking & 99.3 & 97.4 & 100.0 & 97.8 & 100.0 \\
Qwen-3 32B & 92.5 & 87.7 & 94.5 & 86.3 & 27.3 \\
Qwen-3 32B +Thinking & 100.0 & 97.1 & 100.0 & 97.2 & - \\
\bottomrule
\end{tabular}

\end{table*}

%%%%%%%%%%%%%%%%%%%%%%%%%%%%%%%%%%%%%%%%%%%%%%%%%%%%%%%%%%%%

% \newpage
% \clearpage
% \input{checklist.tex}

\end{document}